\documentclass[pmlr]{jmlr} 

\usepackage{longtable}

\usepackage{booktabs}
\usepackage[load-configurations=version-1]{siunitx} 
\usepackage{ulem}
\usepackage{longtable}

\theorembodyfont{\upshape}
\theoremheaderfont{\scshape}
\theorempostheader{:}
\theoremsep{\newline}

\jmlrproceedings{PMLR}{Pre-print under review, submitted to Proceedings of Machine Learning Research}%
\jmlrvolume{}
\jmlryear{}
\jmlrsubmitted{}
\jmlrpublished{}
\jmlrpages{}
\jmlrworkshop{NeurIPS 2021 - Traffic4cast} 

\def\eg{\textit{e.\,g.}}
\def\ie{\textit{i.\,e.}}

\def\t4c{\textit{Traffic4cast}\relax}

\usepackage{graphicx}
\usepackage{color}
\usepackage{xcolor}
\usepackage{amssymb}
\PassOptionsToPackage{hyphens}{url}
\usepackage{hyperref}
\usepackage{csquotes}
\usepackage{mdframed}
\usepackage{verbatim}   
\usepackage{ifthen}
\usepackage{lipsum}
\usepackage{footmisc}
\usepackage{tablefootnote}
\usepackage{microtype}

\let\cite\citep

\def\eg{\textit{e.\,g.}}
\def\ie{\textit{i.\,e.}}

\def\t4c{\textit{Traffic4cast}\relax}

\title[\t4c at NeurIPS 2021]{\t4c at NeurIPS 2021 -- Temporal and Spatial Few-Shot Transfer Learning in Gridded Geo-Spatial Processes \\ {\small \url{http://traffic4cast.ai} -- \url{https://github.com/iarai/NeurIPS2021-traffic4cast}}}
\def\eg{\textit{e.\,g.}}
\def\ie{\textit{i.\,e.}}

\def\t4c{\textit{Traffic4cast}\relax}

\author{%
\Name{Christian Eichenberger}\thanks{Institute of Advanced Research in Artificial Intelligence (IARAI), Vienna, Austria} \Email{christian.eichenberger@iarai.ac.at}
\AND
\Name{Moritz Neun}\footnotemark[1] \Email{moritz.neun@iarai.ac.at}
\AND
\Name{Henry Martin}\footnotemark[1]\thanks{Institute of Cartography and Geoinformation, ETH Zurich, Switzerland} \Email{henry.martin@iarai.ac.at}
\AND
\Name{Pedro Herruzo}\footnotemark[1] \Email{pedro.herruzo@iarai.ac.at}
\AND
\Name{Markus Spanring}\footnotemark[1] \Email{markus.spanring@iarai.ac.at}
\AND
\Name{Yichao Lu}\thanks{Layer 6 AI, Toronto, Canada} \Email{yichao@layer6.ai}
\AND
\Name{Sungbin Choi} \Email{sungbin.choi.1@gmail.com}
\AND
\Name{Vsevolod Konyakhin}\thanks{ITMO University, Saint Petersburg, Russia},
\Name{Nina Lukashina}\thanks{JetBrains Research, Saint Petersburg, Russia},\\\mbox{  }\Name{Aleksei Shpilman}\thanks{HSE University, Saint Petersburg, Russia}\footnotemark[4]
\Email{sevakonyakhin@gmail.com}
\AND
\Name{Nina Wiedemann}\footnotemark[2], \Name{Martin Raubal}\footnotemark[2]
\Email{nwiedemann@ethz.ch}
\AND
\Name{Bo Wang}\thanks{Institute of Transport Studies, Monash University, Clayton Victoria, Australia}, \Name{Hai L. Vu}\footnotemark[7],
\Name{Reza Mohajerpoor}\thanks{CSIRO's Data61, Eveleigh, Australia},
\Name{Chen Cai}\footnotemark[8],\\ \mbox{  }
\Name{Inhi Kim}\thanks{Institute Civil and Environmental Engineering Department, Kongju National University, South Korea}
\Email{bo.wang1@monash.edu}
\AND
\Name{Luca Hermes}\thanks{Machine Learning \& Neuroinformatics Group, Bielefeld University, Germany}, \Name{Andrew Melnik}\footnotemark[10], \Name{Riza Velioglu}\footnotemark[10],\\ \mbox{  }
\Name{Markus Vieth}\footnotemark[10], \Name{Malte Schilling}\footnotemark[10]
\Email{lucahermes24@gmail.com}
\AND
\Name{Alabi Bojesomo} \thanks{Electrical Engineering and Computer Science Department, Khalifa University, Abu Dhabi, UAE}, \Name{Hasan Al Marzouqi}\footnotemark[11], \\ \mbox{  }\Name{Panos Liatsis}\footnotemark[11]
\Email{alabi.bojesomo@ku.ac.ae}
\AND
\Name{Jay Santokhi}\thanks{Alchera Data Technologies Ltd, Cambridge, UK},
\Name{Dylan Hillier}\footnotemark[12], \Name{Yiming Yang}\footnotemark[12], \Name{Joned Sarwar}\footnotemark[12],\\ \mbox{  }
\Name{Anna Jordan}\footnotemark[12], \Name{Emil Hewage}\footnotemark[12]
\Email{jay@alcheratechnologies.com}
\AND
\Name{David Jonietz} \thanks{HERE Technologies, Zurich, Switzerland} \Email{david.jonietz@here.com}
\AND
\Name{Fei Tang} \footnotemark[13] \Email{fei.tang@here.com}
\AND
\Name{Aleksandra Gruca}\thanks{Silesian University of Technology, Gliwice, Poland} \Email{aleksandra.gruca@polsl.pl}
\AND
\Name{Michael Kopp}\footnotemark[1] \Email{michael.kopp@iarai.ac.at}
\AND
\Name{David Kreil}\footnotemark[1] \Email{david.kreil@iarai.ac.at}
\AND
\Name{Sepp Hochreiter}\thanks{Machine Learning Institute, Johannes Kepler University Linz, Austria}\footnotemark[1] \Email{sepp.hochreiter@iarai.ac.at}
\AND
}

\date{\today}
\editor{tbd.}

\begin{document}

\maketitle

\newpage
\begin{abstract}
The IARAI \t4c competitions at NeurIPS 2019 and 2020 showed that neural networks can successfully predict future traffic conditions 1 hour into the future on simply aggregated GPS probe data in time and space bins. We thus reinterpreted the challenge of forecasting traffic conditions as a movie completion task.
U-Nets proved to be the winning architecture, demonstrating an ability to extract relevant features in this complex real-world geo-spatial process.
Building on the previous competitions, \t4c 2021 now focuses on the question of model robustness and generalizability across time and space. Moving from one city to an entirely different city, or moving from pre-COVID times to times after COVID hit the world thus introduces a clear domain shift. We thus, for the first time, release data featuring such domain shifts. The competition now covers ten cities over 2 years, providing data compiled from over $10^{12}$ GPS probe data.
Winning solutions captured traffic dynamics sufficiently well to even cope with these complex domain shifts.
Surprisingly, this seemed to require only the previous 1h traffic dynamic history and static road graph as input.
\end{abstract}

\section{Introduction}\label{sec:introduction}
The global trends of urbanization and increased personal mobility force us to rethink the way we use urban space. The \t4c competitions \cite{Kreil_Traffic4cast_2019,Kopp_Traffic4cast_2020} tackle this problem in a data driven way, encouraging the application of the latest methods in machine learning to modeling complex spatial systems over time.

This year, we provide a unique data set derived from industrial scale trajectories of over $10^{12}$ raw GPS position fixes, with latitude, longitude, time stamp, as well as vehicle speed and driving direction recorded at that time. The data are made available by HERE Technologies and originate from a large fleet of vehicles. For the new temporal and spatial transfer learning challenges introduced in \t4c 2021, we provide data for 10 culturally and socially diverse metropolitan areas around the world, covering a time span of 2 years.

\begin{figure}[htb]
  \centering
  \includegraphics[width=0.95\textwidth]{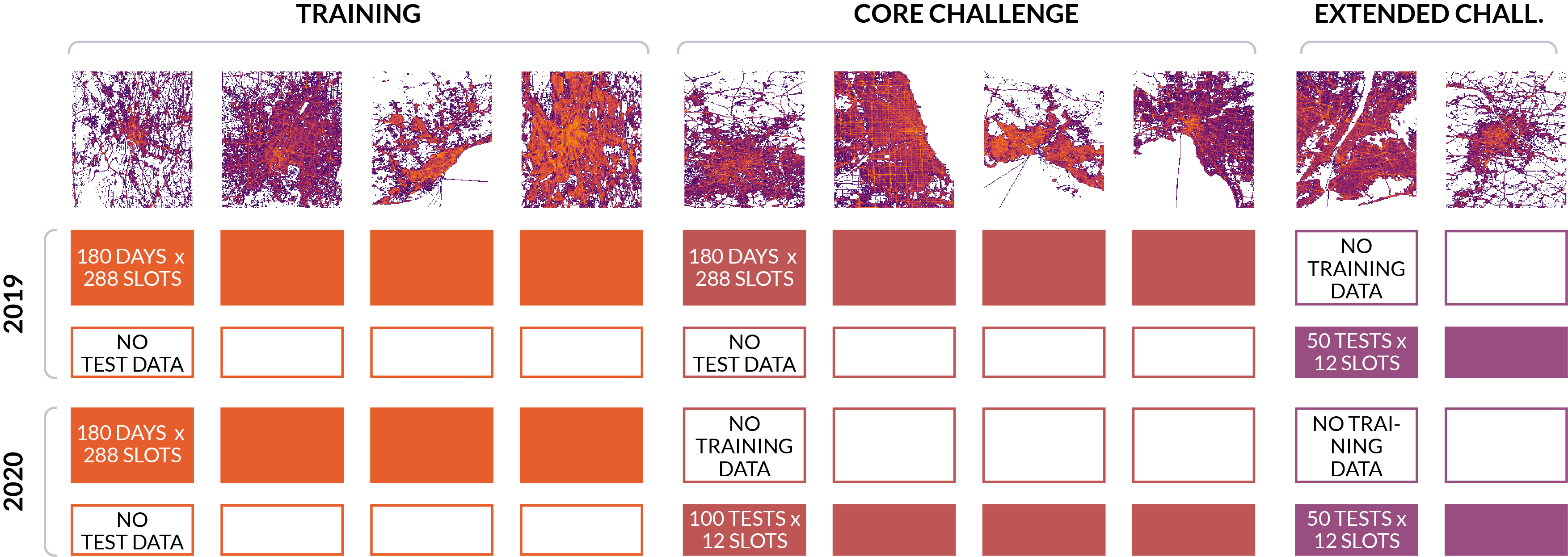}
  \vspace*{-4mm}
  \caption{Data overview. There is data from 10 cities from 2019 and 2020. 4 cities are used for training (180 full days 2019 and 2020; 4 cities are used for the core challenge (180 full days 2019 pre-Covid and 100 test slots 2020 in-Covid); 2 cities are used for the extended challenge (50 test slots 2019 pre-Covid and 50 test slots 2020 in-Covid).}
  \label{fig:data}
  \vspace*{-2mm}
\end{figure}

The competition task was to predict from 1 hour traffic the next 5, 10, 15, 30, 45 and 60\,min into the future. The training data was provided in the same format as last year \cite{Kopp_Traffic4cast_2020}. An overview of the dynamic training data provided and the 1h test time slots to predict is shown in Figure~\ref{fig:data}.
Along with each 1h test time slot, we provided the time of day and the day of the week 
but not the exact date.
For each city, we provided a static graph derived from a road map in the same spatial resolution as the dynamic data, which could be used as mask or as a graph 
\cite{neun2021t4ccompetitiondesign, traffic4cast2021-github}.

For all cities considered, COVID-19 has lead to a visible shift in daily mobility traffic patterns, as can be seen by comparing traffic volumes from the pre-Covid (April/May 2019) and in-Covid data (April/May 2020), see also \citet{neun2021t4ccompetitiondesign} and \citet{traffic4cast2021-github}.
Our {\bf core challenge} requires participants to transfer learn across this domain shift~\cite{bendavid2010, kouw2018, kouw2019,webb2018,gama2014,widmer1996} in traffic caused by the COVID-19 pandemic.
This challenge is thus a {\bf few-shot learning task} \cite{fei2006one, lu20,guo19}, which requires to {\bf transfer learn  traffic dynamics across a temporal domain shift}.

Our {\bf extended challenge} encourages participants to use all the data provided so far (data from $8$ different cities for the pre-COVID era and $4$ different cities during the in-COVID era)
on two hitherto unseen cities for which no further training data is provided.
For each city, $100$ one-hour test time slots are randomly chosen, $50$ from the pre-COVID era and $50$ from the in-COVID era. The underlying machine learning challenge is thus a {\bf few-shot transfer of traffic dynamics across a spatial and temporal domain shift}. It is noted that solutions to the extended challenge can be used as a solution for the core challenge as well, although participants will then make no use of the additional training data which might contain local, spatially relevant information. Apart from the static data and the test time slots, no further data from the target cities are exposed.

Informed by the previous \t4c competitions \cite{Kreil_Traffic4cast_2019,Kopp_Traffic4cast_2020}, we chose two non-trivial baselines.
For the core competition, we used a vanilla U-Net \cite{ronneberger2015u} and trained models for each city separately for 4 epochs on the city's 2019 training data only. 
For the extended competition, a Graph ResNet was used following \citet{martin_graph-resnets_2020}.
More details about the baselines can be found in \citet{neun2021t4ccompetitiondesign} and \citet{traffic4cast2021-github}.

The competition brings together a range of highly active fields in machine learning -- few-shot learning, transfer learning, meta-learning more generally, as well as video frame prediction or graph based modelling. Compared to last year's competition an order of magnitude more data was provided, covering ten cities across 2019 and 2020. 
This wealth of data was the basis for being able to tackle how far data and machine learning driven approaches alone can be used to decipher the implicit, largely unknown rules governing the phenomena of traffic by applying them across complex domain shifts.

The data encoding as Traffic Map Movies \cite{Kreil_Traffic4cast_2019}, featuring multiple channels, provides a natural way to fuse information from multiple sources and show-cases the power of Machine Learning to excel at tasks that previously had to depend on domain knowledge, special data structures working on graphs \cite{NEURIPS2019_ee389847}, and manual feature engineering. 
Leading the way, this approach also proved itself in other domains such as rainfall prediction \cite{agrawal2019machine, Gruca_cdceo_21, herruzo_high_resolution_2021}, featured at CIKM and IEEE Big Data 2021 (\url{https://weather4cast.ai/}).

\section{Standout Solutions}\label{sec:leaderboard_standoutsolutions}
In the following, we give a short summary of outstanding solutions to our Traffic4cast 2021 competition.
For each submission, we give a high-level diagram which we call inference diagram and which, in contrast to architecture diagrams typically used in the ML literature, summarizes the approach from an information-flow perspective, highlighting the trained models, the data used to train these models and to the ensembling of these trained models at inference time (if any). This sheds light on how the large and diverse amount of input data was used for the different tasks. The notation is the following: rectangles refer to data (test input, and test output); rounded rectangles represent functions in the inferences, arrows represent flow of information. We use square brackets to denote parameterized data or functions, bound to the parameters in the test input. For each model, after the colon, we also show the data the model was trained on, using the notation in Table~\ref{tab:synopsis}, see $^d$ in caption.

\subsection{oahciy: U-Net + Multi-Task Learning}
\citet{lu2021learning} presents an amazingly simple multi-task learning framework by randomly sampling data from all available cities (4 training and 4 core) for all models in the ensemble. In the core competition 9 models were trained for 5 epochs while in the extended competition 7 models were trained for 50'000 steps only (8\% of an epoch). The models are all U-Net with varying architecture and seeds. For more details see Table~\ref{tab:synopsis} and Appendix~\ref{appendix:standoutsolutions:lu2021learning}.

\citet{lu2021learning} argues that this multi-task learning can be regarded as an implicit data augmentation and regularization technique when trained on one city only and forcing to learn city-agnostic representations thereby improving data efficiency. The implicit domain-adaptation through the addition of 2019 and 2020 data for at least one city is reported as crucial encouraging the model to learn to adapt to temporal domain shifts during training.

\begin{figure}[htb]
  \centering
  \includegraphics[scale=0.65]{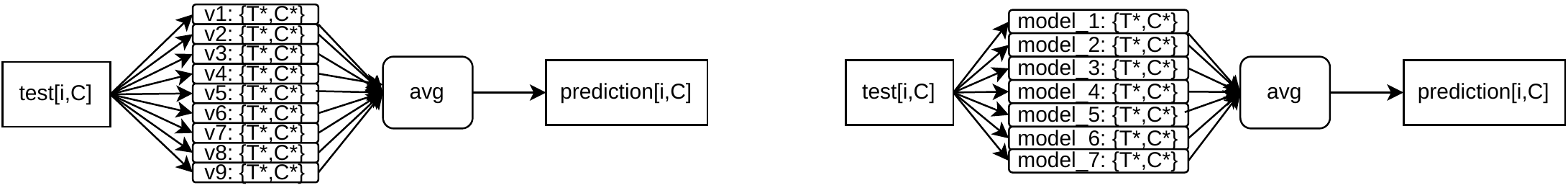}
  \vspace*{-6mm}
  \caption{Inference oahciy \citep{lu2021learning} (left: core competition, right: extended competition).}
  \label{fig:inference_oahciy}
  \vspace*{-4mm}
\end{figure}

\subsection{sungbin: U-Net Ensemble}
The approach of \citet{choi2021applying} is very similar to \cite{lu2021learning}, also averaging ensembles of different U-Net architectures (4 city-independent models). In contrast to \citet{lu2021learning}, \citet{choi2021applying} trained on target city training data, too (3 models in core). For more details see Table~\ref{tab:synopsis} and Appendix~\ref{appendix:standoutsolutions:choi2021applying}.

\begin{figure}[htb]
  \centering
  \includegraphics[scale=0.7]{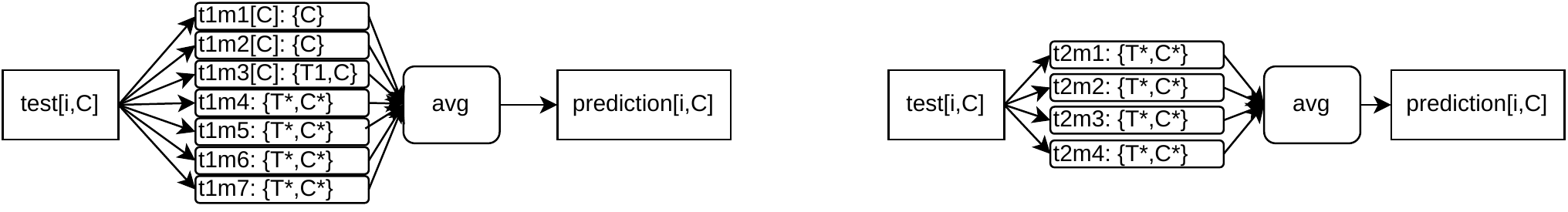}
  \vspace*{-6mm}
  \caption{Inference sungbin \citep{choi2021applying} (left: core competition, right: extended competition).} 
  \label{fig:inference_sungbin}
  \vspace*{-4mm}
\end{figure}

\subsection{sevakon: U-Net with Temporal Domain Adaptation}
\citet{konyakhin2021solving} also base their approach on the success of U-Nets in previous competitions. However, in contrast to \cite{lu2021learning} and \cite{choi2021applying}, they train their models on each target city in the core competition only (they did not participate in the extended competition). They use three different architectures (vanilla U-Net, DenseNet, and EfficientNet \cite{tan2019efficientnet} pre-trained on Imagenet \cite{Deng2009ImageNet}), a static mask derived from dynamic data and a per-pixel and per-channel temporal domain-adaptation (TDA) factor. Their final prediction is derived from these 3 models; each model is used with and without TDA, resulting in 6 predictions to which the static mask is applied and which are then averaged.

\begin{figure}[htb]
  \centering
  \includegraphics[scale=0.65]{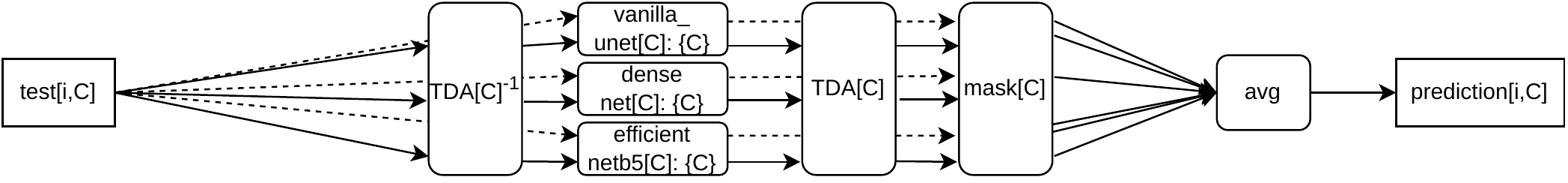}
  \vspace*{-4mm}
  \caption{Inference sevakon \citep{konyakhin2021solving} (core competition). The city mask is derived from the training and test data.} 
  \label{fig:inference_sevakon}
  \vspace*{-4mm}
\end{figure}

\subsection{nina: U-Net++ on Patches}
\citet{wiedemann2021traffic} also use a U-Net variant, but in a patch-based manner, as it was shown to be beneficial in other segmentation tasks~\cite{zhang2006image,ghimire2020patch} (also \cite{misra2020patch} in classification). No static road information was used. This method allowed them to use a parameter-heavier UNet++ with more skip connections \cite{zhou2019unetplusplus,zhou2018unet++}, which they suggest was helpful in light of the sparsity of the data.
Patches are sampled from all available labelled data. After different experiments, choosing the available patches to be $100\times100$ crops with stride 10 was found to perform best. The patch-wise prediction exhibits an ensemble-like behavior.

\begin{figure}[htb]
  \centering
  \includegraphics[scale=0.7]{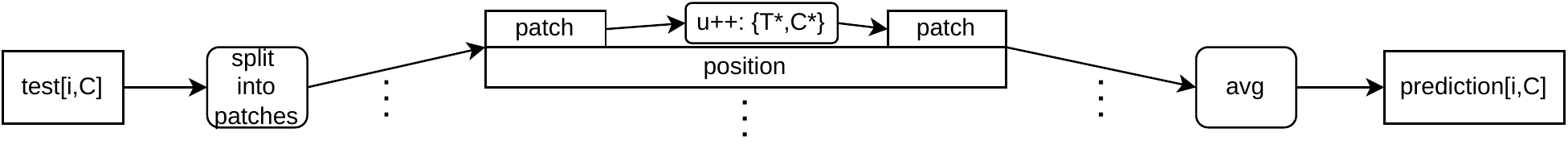}
  \vspace*{-5mm}
  \caption{Inference nina \citep{wiedemann2021traffic} (core competition and extended competition).} 
  \label{fig:inference_nina}
  \vspace*{-2mm}
\end{figure}

\subsection{ai4ex: SWIN-Transformer}
\citet{bojesomo2021hierarchical} uses a Swin-UNet structure where all convolution blocks are replaced by shifted window self attention; downsampling in the encoder is achieved by trainable patch merging layers and upsampling by patch expanding layers in the decoder branch; skip connections are implemented by a combination of addition and concatenation. 

\begin{figure}[htb]
  \centering
  \includegraphics[scale=0.7]{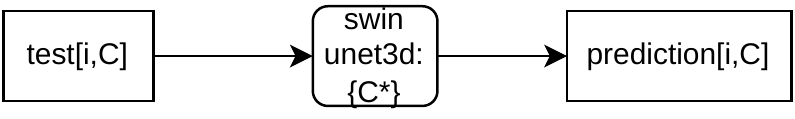}
  \vspace*{-5mm}
  \caption{Inference ai4ex \citep{bojesomo2021hierarchical} (core and extended competition).} 
  \label{fig:inference_ai4ex}
  \vspace*{-2mm}
\end{figure}

\subsection{dninja: Graph-Based U-Net}
\citet{hermes2021Graphbased} present a graph based approach aiming at better generalization and transfer by leveraging knowledge of the underlying road network whilst ignoring areas without any traffic information.
In order for their solution to make full use of the 2d topological information contained in the competition data they use 4 subgraphs corresponding to the provided 4 heading channels of said data.

\subsection{resuly: 3DResNet and Sparse-UNet}
\citet{wang2021traffic4cast} use a 3DResnet \cite{RN13} with 3D convolutions in 4 residual blocks and an output block. For the core competition, this output block consists of sequential CNN layers to restrain the temporal relationship. For the extended competition, the output block consists of a sparse U-Net \cite{graham2014spa,choy20194d} with data in Coordinate Format (COO) for the extended competition. 
\begin{figure}[htb]
  \centering
  \includegraphics[scale=0.7]{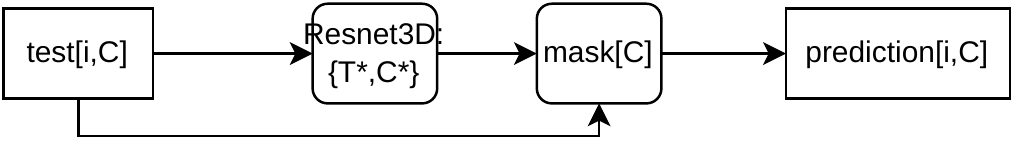}
  \vspace*{-6mm}
  \caption{Inference resuly \citep{wang2021traffic4cast} (left: core competition, right: extended competition). The mask is derived from the test data.} 
  \label{fig:inference_resuly}
  \vspace*{-2mm}
\end{figure}

\subsection{jaysantokhi: Dual-Encoding U-Net}
\citet{santokhi2021dual} use a dual encoding U-Net architecture aiming at a lightweight approach for real world deployments containing significantly fewer parameters (see also Table~\ref{tab:synopsis}) and shorter training times. 
The architecture consists of two encoders one of which has skip connections to the decoder; encoder and decoder consist of Convolutional LSTM layers. The skip connections are not vanilla, but designed to carry the hidden and cell states of the encoder LSTM to the decoder LSTM, which is crucial for the approach.
In both competitions, 4 models are pre-trained on the training cities and fine-tuned on the core competition cities. In the core competition, the city-specific fine-tuned model is used, whereas in the extended competition an architecture with fewer parameters is used and predictions are built by averaging over the outputs of all 4 models.

\begin{figure}[htb]
  \centering
  \includegraphics[scale=0.7]{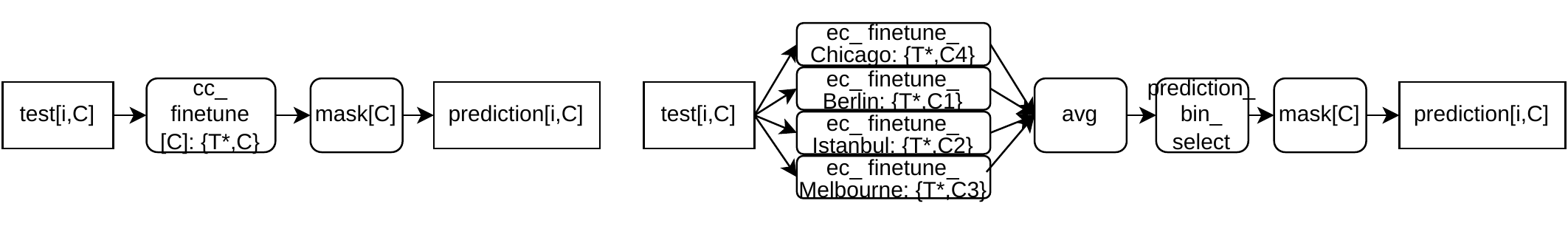}
  \vspace*{-6mm}
  \caption{Inference jaysantokhi \citep{santokhi2021dual} (left: core competition, right: extended competition). The city mask is derived from test data.} 
  \label{fig:inference_jaysantokhi}
  \vspace*{-2mm}
\end{figure}

\section{Synopsis and Discussion}

Looking at the different solutions above we see a large variety of mostly U-Net based approaches. Based on the experiences of the previous year \citep{Kopp_Traffic4cast_2020}, this is not a complete surprise. However, it is interesting to see that for instance the simple averaging in ensembles in the winning solutions was also able to handle the domain shifts, even slightly outperforming some more tailored domain adaptation techniques. Table ~\ref{tab:synopsis} highlights the key aspects and differences of the chosen architectures and informs the discussion below.

\begin{table}[htb]
    \scriptsize\centering
\begin{tabular}{|p{3.6cm}|p{1.7cm}|p{1.0cm}|p{1.2cm}|p{2.8cm}|p{1.6cm}|p{0.7cm}|}
\hline
Team, rank (core/ext), approach & road~graph, time-of-day, day-of-week$^a$ & models trained p. city$^b$ & \#{}models trained$^c$ & Training datasets$^d$ & $\sum\#{}$params core / ext $^e$ & mask$^f$\\ \hline \hline
\textbf{oahciy (1/2)}\newline  U-Net + multi-task learning \citep{lu2021learning} & road graph (concat) & no& 9 / 7& (9/7)$\times$\{T*,C*\} & 1710.2M / 17.1M& --\\ \hline
\textbf{sungbin (2/1)}\newline  U-Net Ensemble \citep{choi2021applying}& road graph (concat) & \textls[10]{in two U-Nets} & 16/4& $2\times$\{C[1-4]\}; \{T1,C[1-4]\}; $4\times$ \{T*,C*\} / $4\times$ \{T*,C*\}& 123.6M / 33.9M& --\\ \hline
\textbf{sevakon (3/--)}\newline \textls[10]{U-Net with Temporal Domain Adaptation \citep{konyakhin2021solving}} & no & yes& 11/--& $3\times$\{C1\}; $2\times$\{C2\}; $3\times$\{C3\}; $3\times$\{C4\} & 342.0M / -- & \textls[10]{city (train/ test data)}\\ \hline
\textbf{nina (6/3)}\newline \textls[2]{U-Net++ on patches \citep{wiedemann2021traffic}} & no& no& 1=1& \{T*,C*\}& 36.7M / 36.7M& --\\ \hline
\textbf{ai4ex (4/6)}\newline SWIN-Transformer \citep{bojesomo2021hierarchical}& no& no& 1=1& \{C*\}& 141.9M / 141.9M& --\\ \hline
\textbf{dninja (7/4)}\newline Graph-based U-Net \citep{hermes2021Graphbased}& road graph, time-of-day, day-of-week& no& 1=1& \{T*,C*\}& 5.8M / 5.8M& by GNN\\ \hline
\textbf{resuly (5/--)}\newline \textls[10]{3DResNet, Sparse-UNet \citep{wang2021traffic4cast}} & road graph & no& 1/1& \{T*, C*\} & 17.3M / 43k& test (test data)\\ \hline
\textbf{jay\-santokhi (8/5)}\newline Dual-Encoding U-Net \citep{santokhi2021dual}& no & \textls[10]{after pre-training} & 4/4& \{T*,C[1-4]\} & 1.0M / 0.3M& city (test data)\\ \hline
\end{tabular}
    \vspace*{-4mm}
    \caption{Synopsis.
      $^a$~what supplemental information is used;
      $^b$~whether some of the trained models used in the inference are specifically trained on the target city;
      $^c$~9/7 means 9 models in the core and 7 different models in the extended competition, whereas 1=1 means the same trained model was used in both competition;
      $^d$~T1=Moscow, T2=Barcelona, T3=Antwerp, T4=Bangkok, C1=Berlin, C2=Istanbul, C3=Melbourne, C4=Chicago, E1=Vienna, E2=New York, T*=all training cities, C*=all core cities, E*=all extended cities. E.g. (9/7)$\times$\{T*,C*\} means 9 models trained on all training and core competition cities for the core competition and 7 from the same cities for the extended competition, and \{T*,C[1-4]\} expands to a model for each city trained on all training cities plus one of the core cities;
      $^e$ Sum of trainable weights of all the model checkpoints used in the inference as extracted from the participants' checkpoint using code in \citet{traffic4cast2021-github};
      $^f$~Kind of mask used for post-processing.}
    \label{tab:synopsis}
    \vspace*{-2mm}
\end{table}

\subsection{Why were the same strategies successful in both competitions?}

\citet{lu2021learning} won the core competition with an ensemble of models all trained with data from all training and core competition cities. This indicates that his solution already is able to capture spatial transfer, although at the price of a high amount of parameters (see Table~\ref{tab:synopsis}). Hence, the combination of a diverse enough set of training cities with a large amount of data together with the static road information seems to be enough to solve both transfer learning challenges of our competition. 
Although the second placed approach of \citep{choi2021applying} is very similar, apart from minor architectural choices, the main difference is that \citet{lu2021learning} only has city-independent models.
In contrast, the models by \citet{konyakhin2021solving} trained only per city are competitive with regards to the temporal transfer. This again gives an indication that the explicit temporal domain-adaptation was necessary in this approach. In contrast, the first two ensembles \cite{lu20,choi2021applying} successfully did an implicit temporal adaptation through the city-independent models trained on data from multiple cities from before and after the temporal shift. Naively, we would have expected to see more domain-adaptation approaches in the core competition like the temporal domain-adaptations by \cite{konyakhin2021solving} or data augmentation techniques to use the test slot inputs for few-shot learning.

It is also remarkable to see that the patch-based approach of \citet{wiedemann2021traffic} is competitive especially for the spatial transfer in the extended challenge. The patches introduced an additional implicit level of ensemble learning within a city. This seems to have had a similar positive impact. The work of \citet{bojesomo2021hierarchical}, \citet{hermes2021Graphbased} and \citet{santokhi2021dual} show that transformers on patches, graph-based approaches and light-weight UNets are also able to handle the transfer tasks in both competitions with, in some cases, significantly smaller models (see Table~\ref{tab:synopsis}).

In addition, we see clear traces of temporal averaging in the qualitative analysis of the outlier special prize (see Section~\ref{sec:results_outliers} and Appendix~\ref{appendix:SpecialPrize}).
Hence, it seems that non-distributional predictions evaluated by MSE seem to encourage trading off spatial, temporal, and channel-wise features jointly. A similar temporal and spatial ``blurring'' effect has been reported in weather forecasting \cite{ravuri_skilful_2021, sonderby_metnet_2020, witt_rainbench_2021}.

Finally, we can see that most approaches did not use any additional dynamic features such as time-of-day or day-of-week. Hence, 1h dynamic input data together with static data seems to capture the city-specific dynamics already, which models can and did then exploit in both our domain shift transfer learning challenges. Of course, one cannot exclude that completely different approaches than those considered and submitted by our competition participants could benefit from such additional features or that they could be beneficial in a different task setting (see Section~\ref{sec:desiderata_metric}).

\subsection{Where do the submission performances differ?}\label{sec:results_performance_difference}
We analyse the MSE loss for each solution above per directed cell (viewing all 4 headings of a pixel as virtual detector for volume and speed separately) -- we bin these directional pixels by their standard deviation in the ground truth speed data (details and full analysis can be found in Appendix~\ref{appendix:mse_variance}).
Thereby, we see where predictions are hard and where the competition was decided. We see that all models struggle in the same std ranges. Of course, MSE does not optimize each directed location independently, so the interpretation here has to be taken \textit{cum grano salis}. 
\begin{figure}[htbp]
  \centering
  \includegraphics[width=0.92\linewidth]{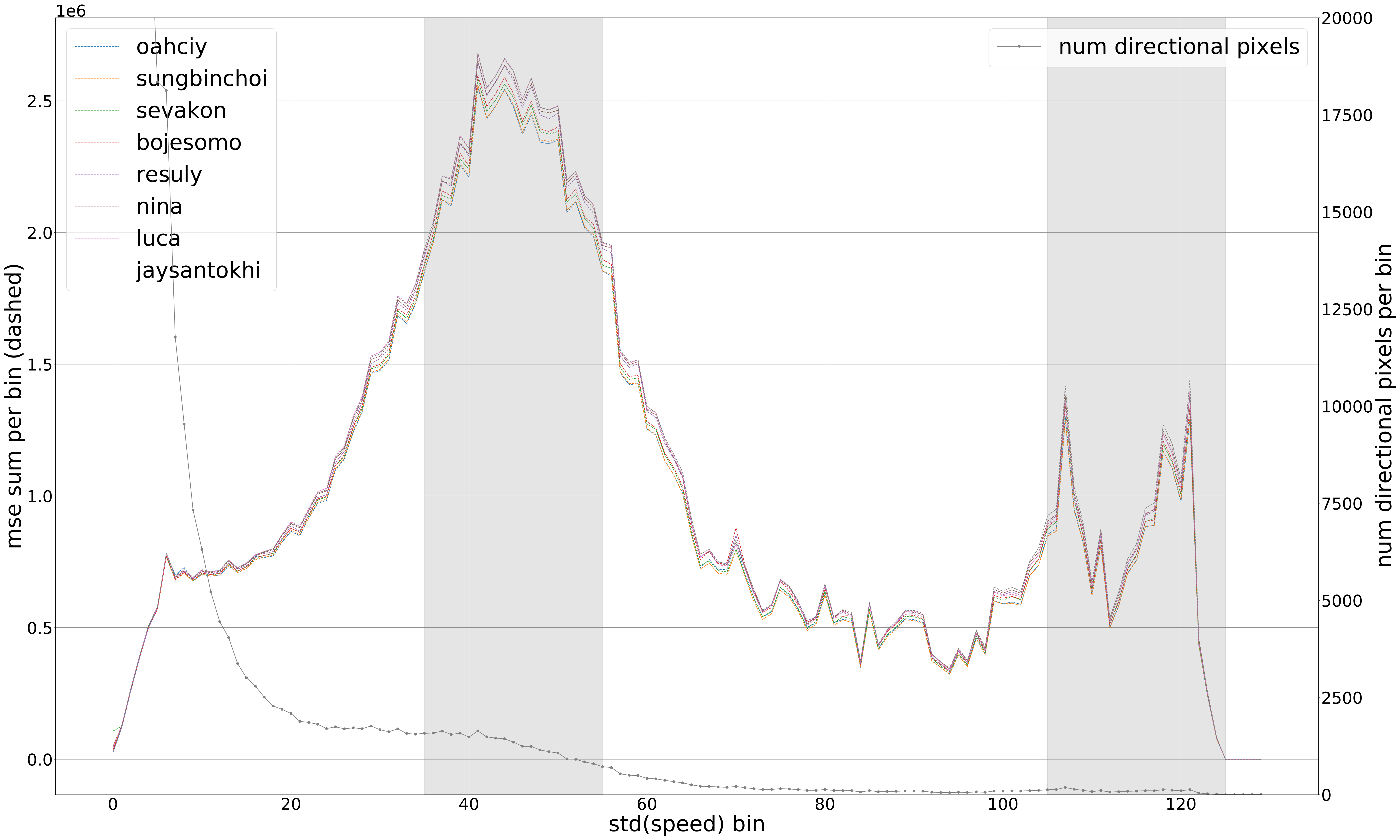}
  \vspace*{-6mm}
  \caption{Relating MSE to std for speeds in BERLIN core: distribution of std among oriented pixels and summed MSE. The shaded gray areas highlights the two critical speed std ranges.} 
  \label{fig:mse_std_speeds_compact}
  \vspace*{-2mm}
\end{figure}
Referring to Figure~\ref{fig:mse_std_speeds_compact}, most of speed MSE losses for all solutions under consideration come from the ranges 35--55 and 105--125 in speed std.
If no data is collected, speed is set to $0$, hence, the higher range implies an oscillation between very high speeds in the upper part of the value range $[0,255]$ and no-data and all solutions seem to have struggled in providing accurate predictions. We localized these two ranges on the city map, see Appendix~\ref{appendix:mse_variance}, and observe: 35--55 speed std range tends to cover the main arteria and some long-range country roads with commuting traffic starting early in the morning; the 105--125 speed std range tends to cover areas with usually high speeds.
This would hint at all solutions struggling where it is hard to find a good strategy of predicting speed in areas of very high variance
and in the large area of medium standard deviation where the amount of such locations makes the total penalty so high.

\subsection{What is the contribution of static road information?}

Regarding static road information, \citet{lu2021learning} and \citet{choi2021applying} did not explicitly compare training with and without static road information as input. The patch-based approach of \citet{wiedemann2021traffic} does not use the static information; this might explain why this approach ranks lower in the core competition.  We suppose that for many test inputs, the static road information is already in the dynamic data; however, in particular in areas of data sparsity, the static road information may be essential. So it remains unclear to what degree static information is required and, in particular, whether it helps in the critical ranges above.

\subsection{What did models learn in outlier situations?}\label{sec:results_outliers}
In order to tackle this questions, we asked participants to re-run their models on a new test set provided for our Outlier Special Prize. The test slots were from two cities of the core competition and the slots were quantitatively evaluated with a masked MSE. For the Special Prize, that mask only contained one pixel and two channels (hence $200 \cdot 6 \cdot 2$ integer values were evaluated against ground truth). These pixels were selected to contain a traffic jam situation reflected by a drastic drop in speed with a simultaneous increase in volume. In the examples analysed in Appendix~\ref{appendix:SpecialPrize}, the winning models of our core and extended competition predict a smoothed version of a jam resolution, underestimating speed and overestimating density. As MSE encourages model predictions to tend towards the mean in the data, they are never predicting a rare scenario such as a jam resolving more or less quickly than in expectation. In the selected outliers MSE for volume and speed is at the same level. More details on the heuristics and the design of our Outlier Special Prize can be found in Appendix~\ref{appendix:SpecialPrize}.

\subsection{What did we learn about the metric?}\label{sec:desiderata_metric}

Both challenges used the pixel-wise mean squared error (MSE) as a loss metric as it is also used in movie prediction tasks~\cite{srivastava2015unsupervised, lee2018stochastic, kwon2019predicting, walker2016uncertain, xue2016visual, han2019video, oprea2020review}. From the analysis in previous subsections pertaining to this year's competition challenges as well as from similar observations from our previous competitions \citep{Kreil_Traffic4cast_2019,Kopp_Traffic4cast_2020},
we are able to identify important, desirable properties a loss metric should have without currently being able to construct one. These properties are as follows.

{\it Desired Property 1: The metric should not be distorted by missing values}.
Zero speed values in the case of no-data lead to a high variance in the speeds to predict in any temporally gridded bin.
In contrast, zero volume for missing values does not affect the variance. A possible solution, suggested by \citet{wiedemann2021traffic}, is to evaluate speed only in case of non-zero volume.
With that models might be less defensive predicting speed.

{\it Desired Property 2: The metric should be flexible in handling spatial and temporal shifts.}
Predicting just slightly at the wrong place leads to a double punishment, encouraging solutions that can be visually observed to show an ‘averaging phenomenon’ even in high scoring solutions as outlined in~\citet{qi2020traffic4cast}.
Shifts in traffic phenomena need to be handled depending on the application needs.

{\it Desired Property 3: The metric should cope with different scales}.
Volume values can be heavily biased due to the fact that GPS probes are only collected by a fraction of the full traffic. Speed values are in most cases less biased (apart from idiosyncrasies such as (un)loading) but often cluster (e.g. around signaled speed limits under normal traffic conditions). In the competition no normalization was used on purpose, giving more weight to speed predictions.
Furthermore, outlier situations are rare, but critical, and therefore tend to not be given enough weight compared to their utility by MSE.
Hence, the metric should be able to incorporate multiple aggregation levels.

{\it Desired Property 4: The metric should be distribution-aware.}
Single-value predictions tend to produce averaging scenarios as we see in the analysis of outliers (see Appendix~\ref{appendix:SpecialPrize}).
Similar to \citet{espeholt2021skillful} and \citet{ravuri2021skillful} predicting a possible distribution instead of a single scalar could help estimating the uncertainty and evaluating at the distributional level. Such an approach would probably make most sense in combination with {\it Desired Property 2}.

\subsection{Has transfer learning been achieved? In what sense?}

If we think of the winning U-Net architecture \citep{lu2021learning}, then models are able to combine global and local information to create a good average forecast. However, the pixel-wise MSE metric leads to models that do not yet cover sudden changes over time that can occur in real-world traffic dynamics (see Sections~\ref{sec:results_outliers} and \ref{sec:desiderata_metric}).

\section{Summary and Outlook}

It is encouraging to see that so many different machine learning approaches lead to competitive results in our map movie completion task.
Winning solutions
captured traffic dynamics sufficiently well to even cope with complex domain shifts.
Surprisingly, this seemed to require only the previous 1h traffic dynamic history and static road graph as input.
In addition, our competition results point to interesting future research directions and questions.

The pixel-wise MSE loss metric encouraged solutions that performed poorly on real-world relevant
outlier situations with their prediction often being averaged scenarios.
Improving on this metric 
may necessitate a range of relevant real-world tasks, each with their own external task-specific metric and reference data, such as predicting estimated times of arrival, ETA \cite{derrowpinion2021ETA}. In another domain, the climate extremes indices provide a standard catalogue of metrics \cite{climdex}. The need to consider complementary domain specific tasks was recently also demonstrated in precipitation forecasting \cite{ravuri2021skillful,espeholt2021skillful}. For traffic, unfortunately, there seems to be no such standard catalogue of application tasks, metrics, and data.

The \t4c 2021 competition has proven again that representing traffic in a temporal and spatial grid is a powerful framework
that allows for complex questions in traffic research to be addressed.
The prediction of effects of road closures, changed speed limits, or an addition of lanes could thus be equally  formulated in such a representation as simple completion tasks. 
Fundamental questions of classical traffic research can thus be approached in a purely data-centric, non-expert dependent way with the help of machine learning.
Finally, it is of key interest to explore whether this framework can be adapted or extended to incorporate new data sources such as satellite data, loop counter data, or crowd-sourced traffic data. The necessary sensor fusion and cross-platform translation tasks are interesting applications of machine learning.

\appendix
\acks{We would like to thank HERE technologies for making our competition data available.}
\addcontentsline{toc}{part}{References}
\bibliography{proceedings/references.bib, proceedings/submissions.bib}
\newpage
\part*{Appendices}
\addcontentsline{toc}{part}{Appendix}

\section{Standout Solution Details}\label{appendix:standoutsolutions}

This appendix gives more details on the leaderboard contributions. In the section header, we give the paper title along with their achieved ranks in the core/extended competition in parentheses.
For each submission, we also give an inference diagram which in our opinion is supposed to summarize the approach from an information-flow perspective by linking to the trained models, the data used in training rather than to the architecture details and to the ensembling.

\subsection{oahciy:  U-Net + Multi-Task Learning}
\label{appendix:standoutsolutions:lu2021learning}
\cite{lu2021learning} presents an amazingly simple multi-task learning framework by randomly sampling data from all available cities and training the models to jointly predict future traffic for different cities. 
\cite{lu2021learning} uses ensembles by averaging 9/7 different U-Net models with varying architecture and seeds, all trained on all data for 4 training cities and 4 core cities. 
The models in the core competition are trained for 5 epochs, the models for the extended competition for 50'000 steps only.

\cite{lu2021learning} argues that the multi-task learning can be regarded as an implicit data augmentation and regularization technique preventing overfitting when trained on one city only and forcing to learn city-agnostic representations and improving data efficiency. \cite{lu2021learning} reports the results of a comparison with a series of domain-adaptation techniques and find that their approach is superior in both competitions and report to have tried different ensembling techniques. For temporal domain adapation, he conducted experiments by using 3 of the training cities as training data and 1 city (Bangkok) for evaluation; his findings suggest that training on the target city is crucial in comparison to more data from the 3 training cities and that adding 2019 and 2020 data for at least one city is crucial encouraging the model to learn to adapt to temporal domain shifts during training. With respect to the core competition, they achieved good performance in the extended competition by reducing the number of parameters in the U-Net model and adopting an early stopping strategy. \cite{lu2021learning} suggests performance could be improved by fine-tuning models in a city-dependent manner and using manually designed features like time of day.

\begin{figure}[htb]
  \centering
  \includegraphics[scale=0.65]{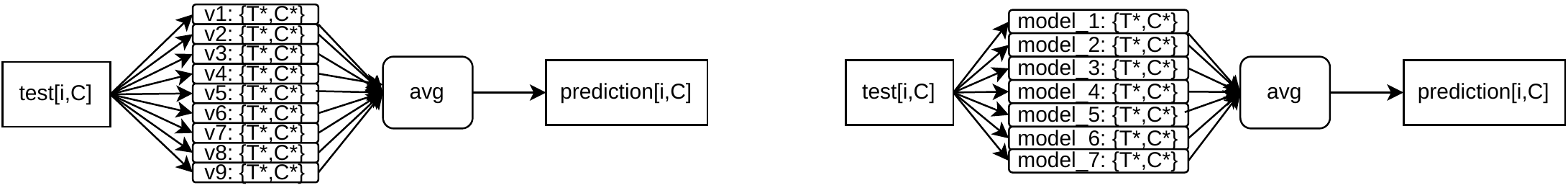}
  \vspace*{-4mm}
  \caption{Inference oahciy \citep{lu2021learning} (left: core competition, right: extended competition).}
  \label{fig:inference_oahciy_appendix}
  \vspace*{-2mm}
\end{figure}

\subsection{sungbin:  U-Net Ensemble}
\label{appendix:standoutsolutions:choi2021applying}
The approach of \cite{choi2021applying} is very similar to \cite{lu2021learning}, also using different U-Net architectures and averaging ensembles. In contrast to \cite{lu2021learning}, \cite{choi2021applying} trains on target city-dependent training data, too, resulting in ensembles of 16/4 models in the two competitions; each prediction comes from an ensemble of 7/4 models, of which 3/- are city-dependent. 
\cite{choi2021applying} reports that in previous  participations \cite{choi2019traffic,choi2020utilizing} having training data from different cities hurt performance; \cite{choi2021applying} suggests that the additional cities for training only make the difference; however, no ablation studies are presented for this claim. As Berlin and Istanbul are in the core competition target cities as in the previous competition, we miss the exploration of the performance of last year's approach and of the contribution of static road data.

\begin{figure}[htb]
  \centering
  \includegraphics[scale=0.7]{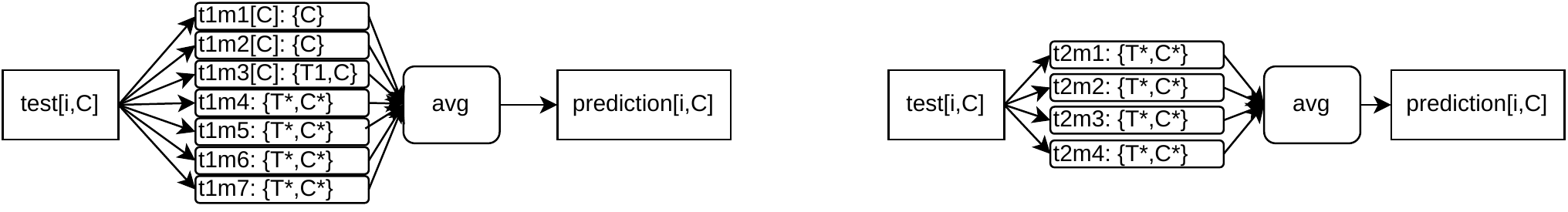}
  \vspace*{-4mm}
  \caption{Inference sungbin \citep{choi2021applying} (left: core competition, right: extended competition).} 
  \label{fig:inference_sungbin_appendix}
  \vspace*{-2mm}
\end{figure}

\subsection{sevakon:  U-Net with Temporal Domain Adaptation}
\label{appendix:standoutsolutions:konyakhin2021solving}
\cite{konyakhin2021solving} also base their approach on the success of U-Nets in previous competitions. However, in contrast to \cite{lu2021learning} and \cite{choi2021applying}, they train their models on the target city in the core competition only (they did not participate in the extended competition). They use three different architectures (vanilla U-Net, DenseNet, and EfficientNet pre-trained on Imagenet \cite{Deng2009ImageNet}), a static mask derived from dynamic data and a per-pixel and per-channel temporal domain-adaptation factor. Their final prediction is derived from the 3 models; each model is used with and without TDA, resulting in 6 predictions to which the static mask is applied and which are then averaged. They find slightly better results are achieved by using a static pixel-wise binary mask generated from the training and test data for non-zero values among all channels. In addition to their TDA heuristics, \cite{konyakhin2021solving} also investigated pseudo-labelling without improvement; pseudo-labels are generated by applying the model on the (unlabelled) samples from the target domain and use these in training; pseudo-labelling had been used as an entropy regularization for probabilistic multi-class classification tasks \cite{jaiswal2019semisupervised,Lee2013pseudolabel}.

\begin{figure}[htb]
  \centering
  \includegraphics[scale=0.7]{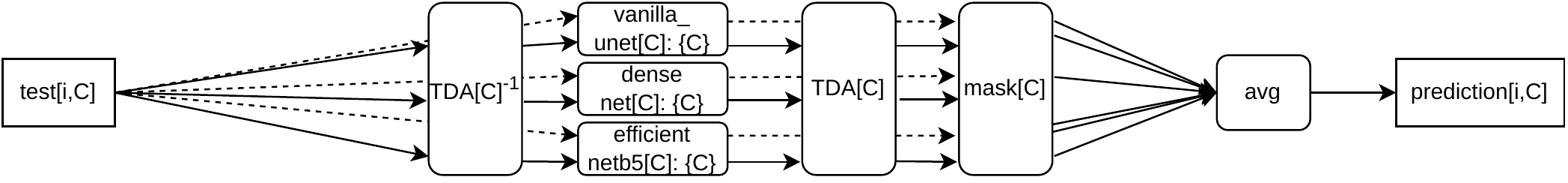}
  \vspace*{-4mm}
  \caption{Inference sevakon \citep{konyakhin2021solving} (core competition). The city mask is derived from the training and test data.} 
  \label{fig:inference_sevakon_appendix}
  \vspace*{-2mm}
\end{figure}

\subsection{nina:  U-Net++ on Patches}
\label{appendix:standoutsolutions:wiedemann2021traffic}
\cite{wiedemann2021traffic} also use a U-Net variant, but in patch-based manner, as it was shown beneficial in other segmentation tasks~\cite{zhang2006image,ghimire2020patch} (also \cite{misra2020patch} in classification), without using the static road information. Using patches allowed them to  use parameter-heavier UNet++ with more skip connections \cite{zhou2019unetplusplus,zhou2018unet++}, which they suggest to have been helpful in light of the sparsity of the data. They subsample from the available labelled data, processing  1000 patches for two epochs and then re-sampling (10 patches from 100 files). At inference time, the predictions are built by averaging the per-cell predictions from all patches that contain the cell.

They compare different model variants and different patch sizes and strides; their best performance was achieved in both competitions with quadratic $100\times100$ patches and stride 10. They show that the ensemble-like behaviour of patch-wise prediction accounts for small but significant increase in performance with smaller strides; they argue that the main advantage of the method lies in the simplification of the problem by splitting the data into smaller parts.
Furthermore, \cite{wiedemann2021traffic} did an error analysis: the overwhelming part of the error is due to the predictions in the speed channels and volumes are zero most of the time. They further looked into those cells with non-zero volume data and derive speed MSE on those cells, which seems to be ``hardly better than chance'', at least for a small validation set from the labelled data. 
They even investigated into training with a loss function that masked out speed on zero ground truth volumes. At test time, the speed of zero-predicted-volumes is imputed with zero.
However, the current competition metric does not reflect the joint volume-speed distribution directly and under MSE different speed predictions might be less penalized.
In discussion, the authors also pointed out that MSE masked this way might incentivize models to learn to predict ``free flow speed'' (\ie{} speeds taken without the presence of other vehicles, see \eg{} \url{https://en.wikipedia.org/wiki/Fundamental_diagram_of_traffic_flow}) in case of zero volume or a speed appropriate to the current traffic situation and not needing to gamble for the data collection gaps (as the GPS probes come from vehicle fleets representing only a part of total traffic).

\begin{figure}[htb]
  \centering
  \includegraphics[scale=0.7]{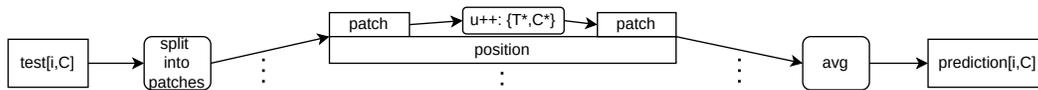}
  \vspace*{-5mm}
  \caption{Inference nina \citep{wiedemann2021traffic} (core competition and extended competition).} 
  \label{fig:inference_nina_appendix}
  \vspace*{-2mm}
\end{figure}

\subsection{ai4ex:  SWIN-Transformer}
\label{appendix:standoutsolutions:bojesomo2021hierarchical}
\cite{bojesomo2021hierarchical} uses a Swin-UNet structure where all convolution blocks are replaced by shifted window self attention; downsampling in the encoder is achieved by trainable patch merging layers and upsampling by patch expanding layers in the decoder branch; skip connections are implemented by a combination of addition and concatenation. 
The paper compares 3 different configurations with different embedding dimensions, feature mixing to find the best model in the leaderboard.

\begin{figure}[htb]
  \centering
  \includegraphics[scale=0.7]{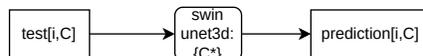}
  \vspace*{-4mm}
  \caption{Inference ai4ex \citep{bojesomo2021hierarchical} (core and extended competition).} 
  \label{fig:inference_ai4ex_appendix}
  \vspace*{-2mm}
\end{figure}

\subsection{dninja:  Graph-Based U-Net}
\label{appendix:standoutsolutions:hermes2021Graphbased}
\cite{hermes2021Graphbased} are the only graph-based contribution in the competition; they aim to leverage prior knowledge on the underlying structure of the street network, ignoring areas without any traffic information, and thereby to achieve better generalization and transfer  by using a graph-based approach.
Vanilla graph layers do not capture 2-dimensional topology like CNNs do -- in order to capture this information, they enhance existing graph layers by using 4 subgraphs corresponding to the 4 headings of the challenge data.
They show that these subgraphs are consistently beneficial for all cities; edge features contain a CNN-based embedding of the road graph as image; a global state vector contains the summed-up and scaled node features and encoding of time of day and day of week. \cite{hermes2021Graphbased} use these Graph layers in a UNet architecture, where the 2D node positions are use for up and downsampling on graphs: downsampling works by taking the feature-wise max of node and edge features, while upsampling works by introducing edges from the input graph to the target graph and using graph propagation. This up- and downsampling approach effectively expands the receptive field of the graph approach without requiring too deep GNNs, which are thought to lead performance drop at a certain depth. Although ablation studies and extensive hyperparameter tuning are missing in their work, the work seems promising.

\subsection{resuly:  3DResNet and Sparse-UNet}
\label{appendix:standoutsolutions:wang2021traffic4cast}
\cite{wang2021traffic4cast} use 3DResnet \cite{RN13} with 3D convolutions in 4 residual block and an output block of sequential CNN layers to restrain the temporal relationship in the core competition and Sparse U-Net \cite{graham2014spa,choy20194d} with data in Coordinate Format (COO) for the extended competition. 
They enhance data loading by two-level shuffling over day files and then over indices, randomly picking up a number of files as training samples in each epoch. To ensure spatial and temporal diversity, these day files need to include all available cities in 2019 and 2020 and cover all seven days of the week.
They show large training speed-ups for the Sparse UNets. Comparing the two approaches by training on each city separately, the find no consistently better method, with considerable differences in the convergence behaviour and the final loss achieved; above all, they conjecture that sparse UNets generalize better on the extended challenge but perform worse on the core challenge, however without giving systematic comparison of the same method applied in both settings.

\begin{figure}[htb]
  \centering
  \includegraphics[scale=0.7]{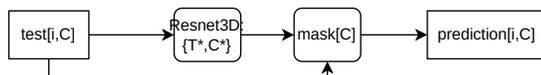}
  \vspace*{-5mm}
  \caption{Inference resuly \citep{wang2021traffic4cast} (left: core competition, right: extended competition). The mask is derived from the test data.} 
  \label{fig:inference_resuly_appendix}
  \vspace*{-2mm}
\end{figure}

\subsection{jaysantokhi:  Dual-Encoding U-Net}
\label{appendix:standoutsolutions:santokhi2021dual}
\cite{santokhi2021dual} use a dual encoding U-Net architecture aiming at a lightweight approach for real world deployments containing significantly fewer parameters and shorter training times. 
The architecture consists of two encoders one of which has skip connections to the decoder; encoder and decoder consist of Convolutional LSTM layers. The skip connections are not vanilla, but designed to carry the hidden and cell states of the encoder LSTM to the decoder LSTM, which is crucial for the approach.
In both competitions, 4 models are pre-trained on the training cities and fine-tuned on the core competition cities. In the core competition, the city-specific fine-tuned model is used, whereas in the extended competition an architecture with fewer parameters is used and predictions are built by averaging over the outputs of all 4 models;
in addition, in the core competition, the model outputs directly 6 frames, whereas the model in the extended competition outputs 12 frames. In both competition, a mask derived from the test data was found to be more performant than the static mask or mask based on the training data.

\begin{figure}[htb]
  \centering
  \includegraphics[scale=0.7]{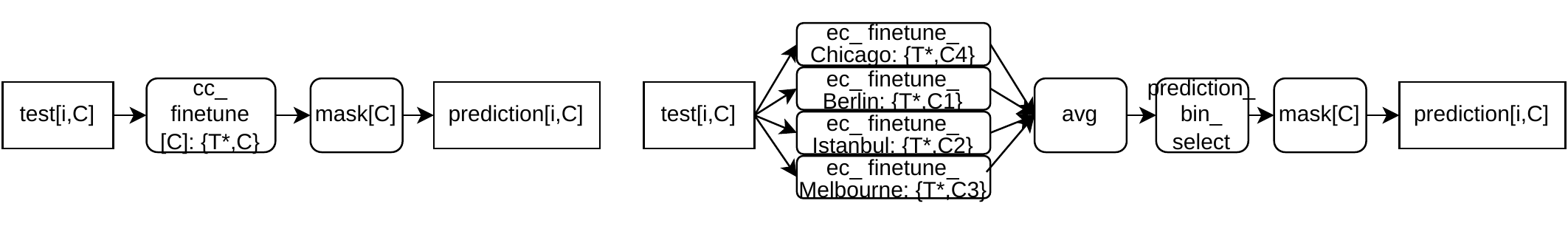}
  \vspace*{-4mm}
  \caption{Inference jaysantokhi \citep{santokhi2021dual} (left: core competition, right: extended competition). The city mask is derived from test data.} 
  \label{fig:inference_jaysantokhi_appendix}
  \vspace*{-2mm}
\end{figure}

\clearpage
\section{Relating MSE to the Variance in the Data: Where do Models Struggle?}\label{appendix:mse_variance}
Here, we detail on Section~\ref{sec:results_performance_difference}.

For the analysis, we took standard deviation on the ground truth data of the 100 tests of the core competition (\ie{} 18 time bins each, the 1 hour input and the 6 prediction slots from the 1 hour prediction horizon, see \cite{neun2021t4ccompetitiondesign} for the test slot sampling details). We compute MSE  for each directional pixel ($863280=495\cdot 436 \cdot 4$ in total), separately for volume and speed. Finally, we plot MSE per std bin to find ``hot spots''.

\subsection{Relating speed variance to speed MSE}

\begin{figure}[htbp]
    \centering
    \includegraphics[width=0.95\textwidth]{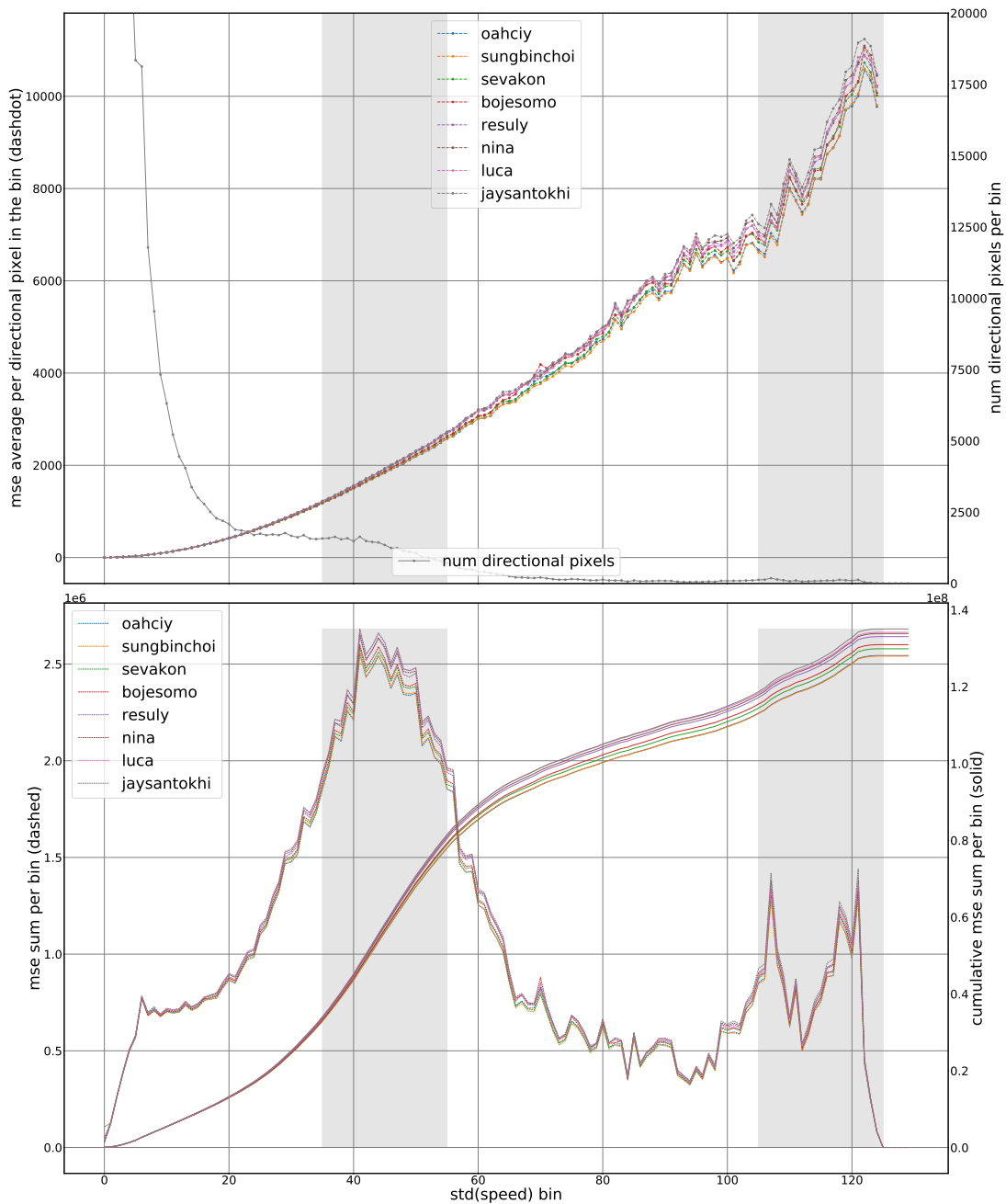}
    \caption{Relating MSE to std for speeds in BERLIN core: distribution of std among directional pixels (axis capped) and average MSE (top); summed MSE and cumulated summed MSE (bottom).}
    \label{fig:mse_std_speeds}
\end{figure}

We have a strong asymmetry in the MSE levels in both competitions (see Figure~\ref{fig:leaderboard_speeds_volumes}), \eg{} for \citet{lu2021learning}, the winner of our core competition, we have
speed\_mse: 148.427,
vol\_mse: 10.440,
all\_mse: 79.434
for MSE on the city of Berlin only (plots and data can be found in \citet{traffic4cast2021-github}).
Therefore, we put the focus of our analysis on relating speed variance to speed MSE. 
If we bin the test data by speed standard deviation as in Figure~\ref{fig:mse_std_speeds}, we see that most directional pixels have little standard deviation. This is indicated by the gray line going down almost monotonically showing the counts for every speed std bin. Also notice that we do not show the full y range on the left.

If we relate this to MSE for speed of the submissions in the core competition, we see that speed MSE per directional pixel in the bin is going up almost monotonically and faster than in a linear fashion Figure~\ref{fig:mse_std_speeds}. The curves of the different participants are pretty in parallel, so it looks as if they struggle at the same places. The different participants are represented by different colors.

If we multiply the two curves of Figure~\ref{fig:mse_std_speeds} top, we get the dashed curves of Figure~\ref{fig:mse_std_speeds} bottom, which shows the sum of the MSEs of all directional pixels in the bin. 
This allows us to see which speed std areas contribute most to the total speed MSE. 
The monotonic growing solid curve are the cumulative sums of the dashed ones, containing the same information as the dashed ones. A peak in the dashed curves is reflected by steepness in the solid curve. 
The final value top right is the sum over all $(495, 436, 4)$ virtual speed detectors of the city, summing the per-pixel and per-heading MSE over all 100 test slots. If we divide the final cumulated number by the the number of such directional pixels, we get the speed\_mse of 148.

We find two critical ranges of speed std where most of the final MSE is accumulated: around 35--55 and  around 105--125.
In the lower critical range (35--55), average MSE per speed std bin more than, doubles
and increases monotonically, while the number of pixels per bin is slightly decreasing, which cumulates heavily; so despite the medium average MSE level, this translates into a peak in the per-bin cumulated MSE bin curves. 
The higher band (105--125) covers very high average MSE, but at a low number of pixels, which also translates into a peak.

\subsubsection{Mapping out the two critical ranges}
We now visualize the locations of these two critical speed std ranges.

\begin{figure}[htbp]
  \centering
  \includegraphics[width=0.4\textwidth]{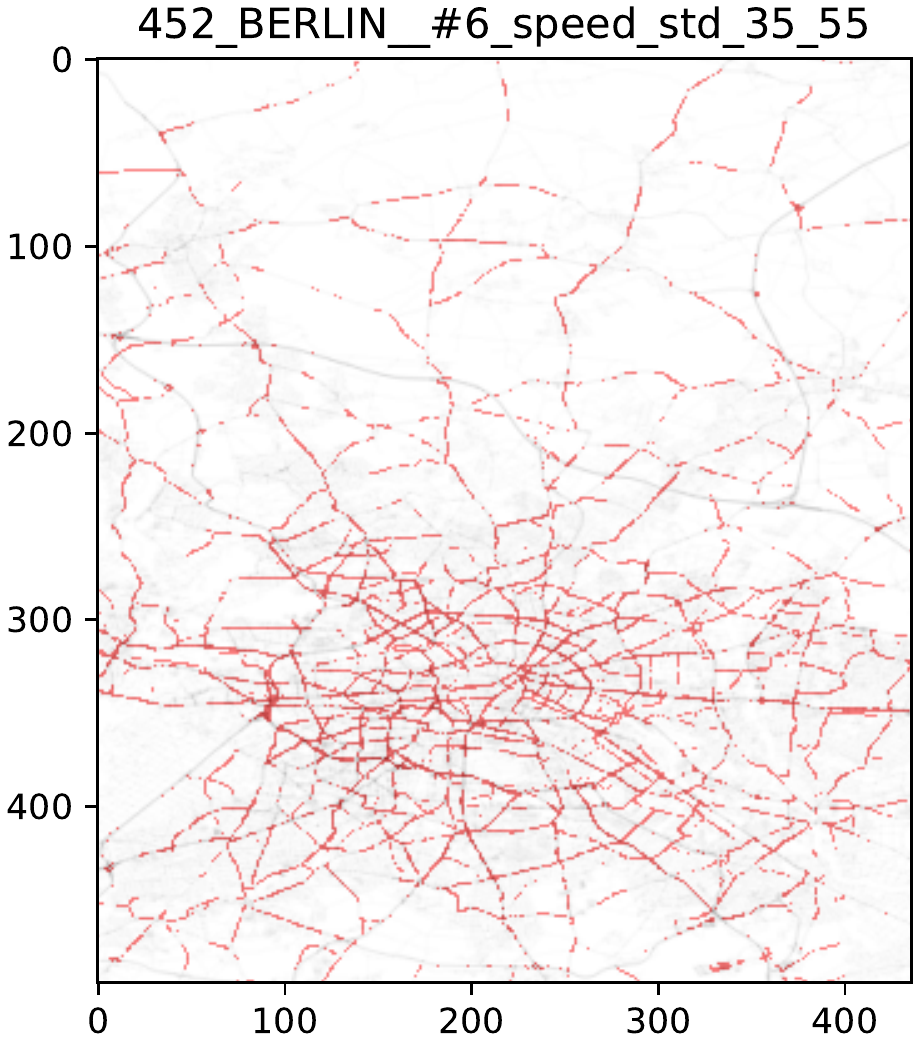}
  \includegraphics[width=0.4\textwidth]{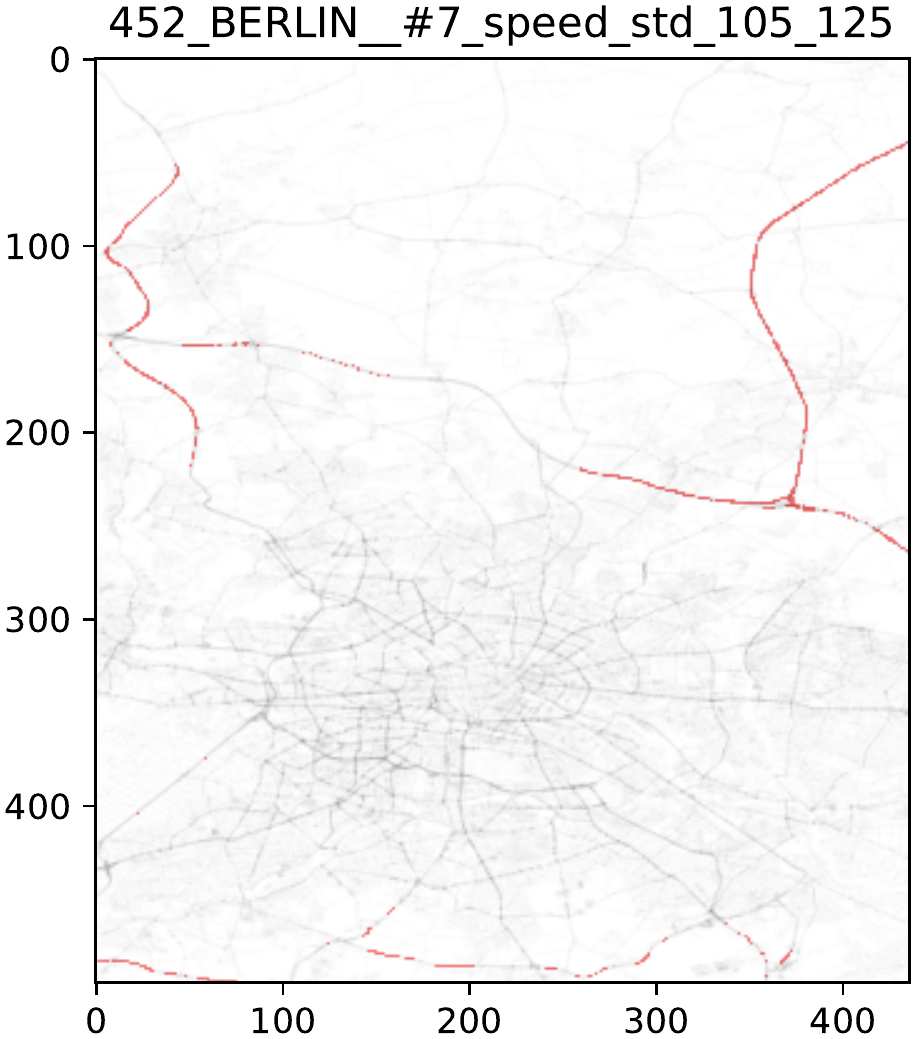}
  \vspace*{-4mm}
  \caption{Locations (red) of the 35--55 (left) and 105--125 (right) speed std bands}
  \label{fig:upper_lower_band}
  \vspace*{-2mm}
\end{figure}

Figure~\ref{fig:upper_lower_band} shows the locations of the 35--55 speed std band on the left side. This covers the main arteria as well as parts of long-range country roads where commuting traffic starts early in the morning. So these may be areas that are naturally hard to predict. This stems on the one hand from medium ``free flow speeds'' and high speed variance or no data, and on the other hand from relatively high ``free flow speeds'' and a lot of no-data in off-peak times.

The 105--125 speed std band on the right side in Figure~\ref{fig:upper_lower_band} shows areas with usually high speeds and some no-data during the night.

\subsubsection{Revisiting 3 Berlin locations}

\begin{figure}[htbp]
  \centering
  \includegraphics[width=1.0\linewidth]{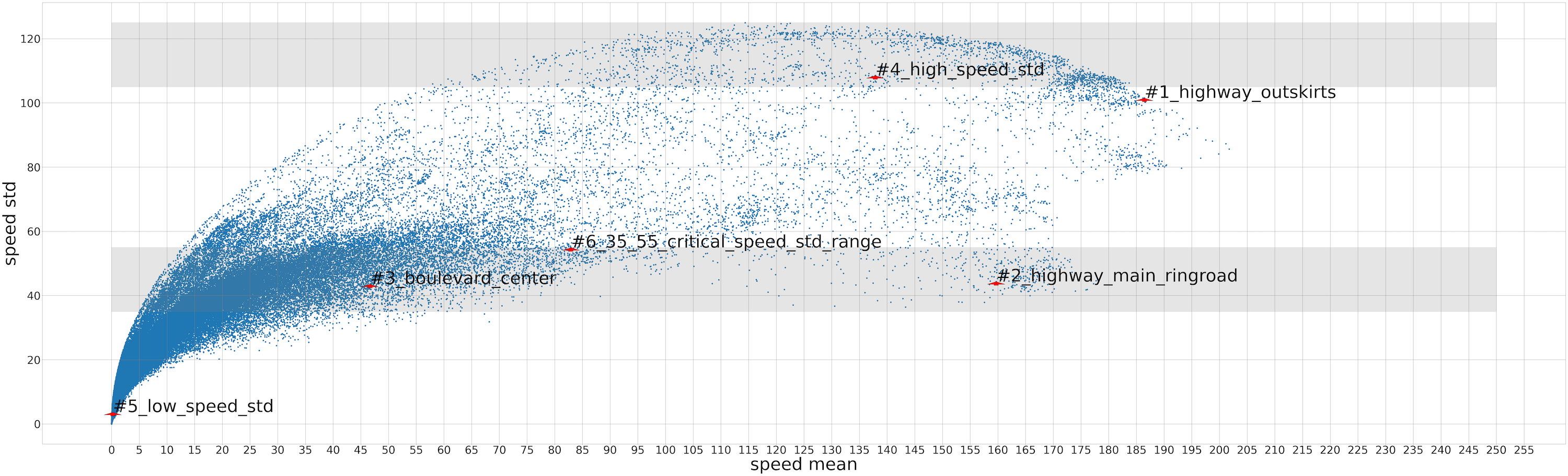}
  \vspace*{-6mm}
  \caption{Critical ranges and sample situations in Berlin.}
  \label{fig:critical_bands}
  \vspace*{-2mm}
\end{figure}

\begin{figure}[htbp]
  \centering
  \includegraphics[width=1.1\linewidth]{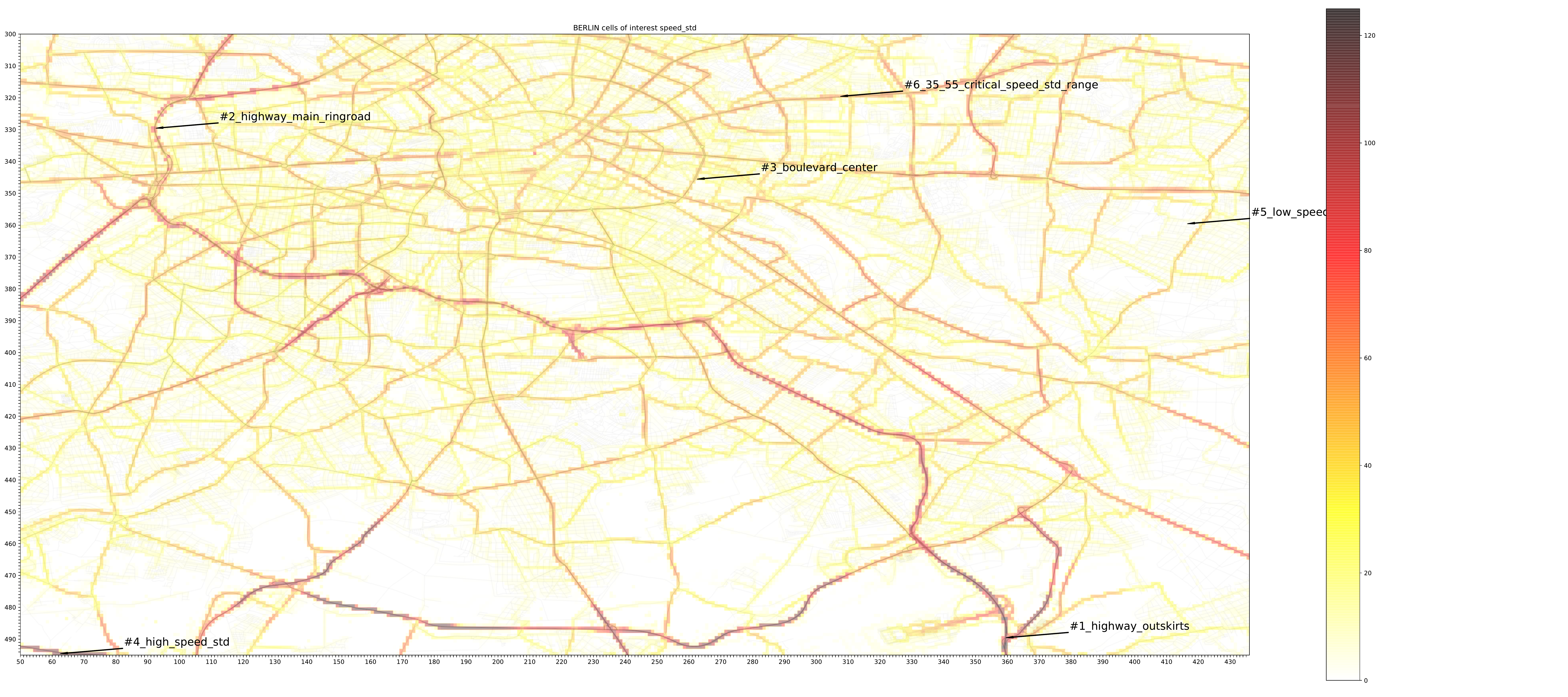}
  \vspace*{-8mm}
  \caption{Speed std heatmap for Berlin showing the max of all 4 headings per pixel, showing the 3 Berlin locations.}
  \label{fig:speed_std_heatmap}
  \vspace*{-2mm}
\end{figure}

\begin{figure}[htb]
  \centering
  \subfigure[\#1 highway outskirts (489, 359, NE). vol: $6.78\pm{}6.68$, speed=$186.46\pm{}101.55$]{
\includegraphics[width=0.9\linewidth]{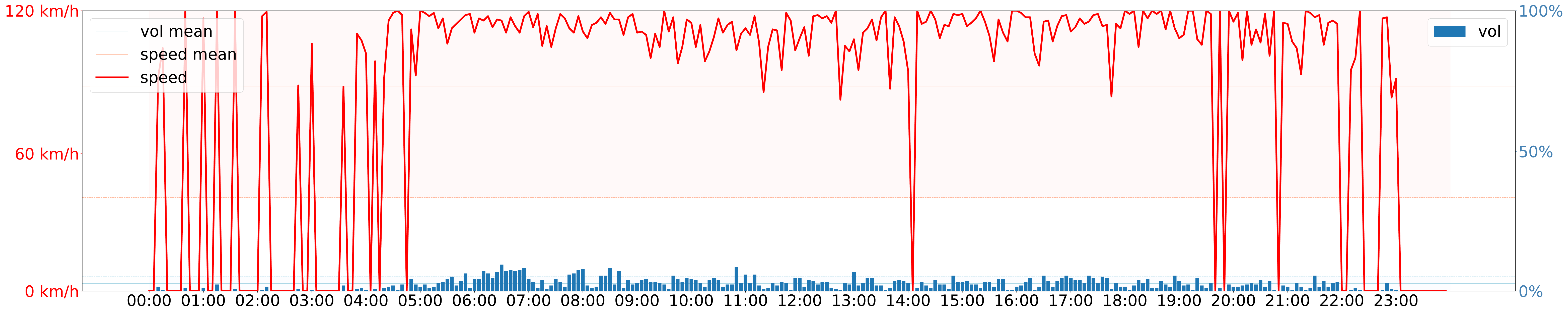}
  }
  \subfigure[\#2 highway main ringr. (329, 92, NE). Vol: $16.80\pm{}15.112$, speed=$ 159.69\pm{}44.37$]{\includegraphics[width=0.9\linewidth]{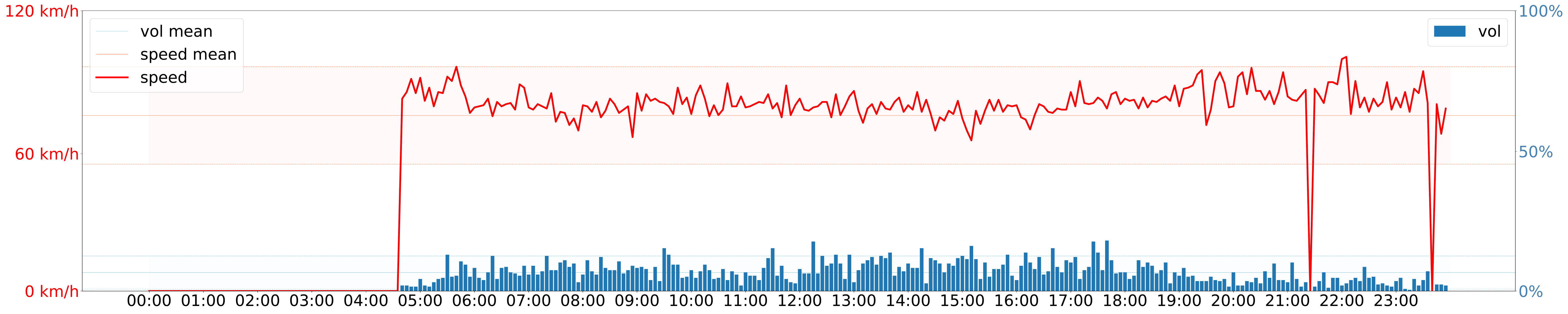}}
    \subfigure[\#3 boulevard center (345, 262, NE). Vol: $2.57\pm{}43.15$, speed=$ 46.69\pm{}43.56$]{\includegraphics[width=0.9\linewidth]{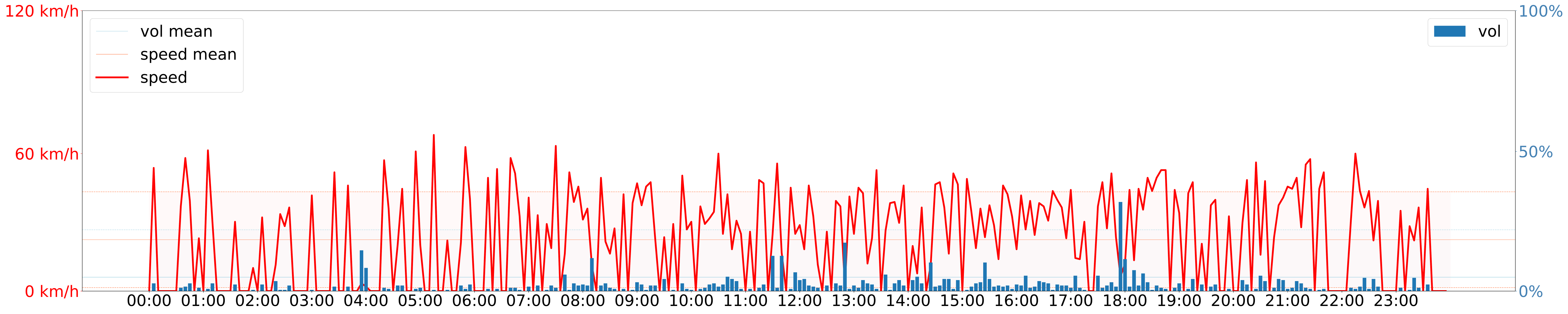}}
  \subfigure[\#4 (494, 62, SE) in 105--125 speed std r. Vol: $2.78\pm{}3.48$, speed=$ 137.87\pm{}108.55$]{\includegraphics[width=0.9\linewidth]{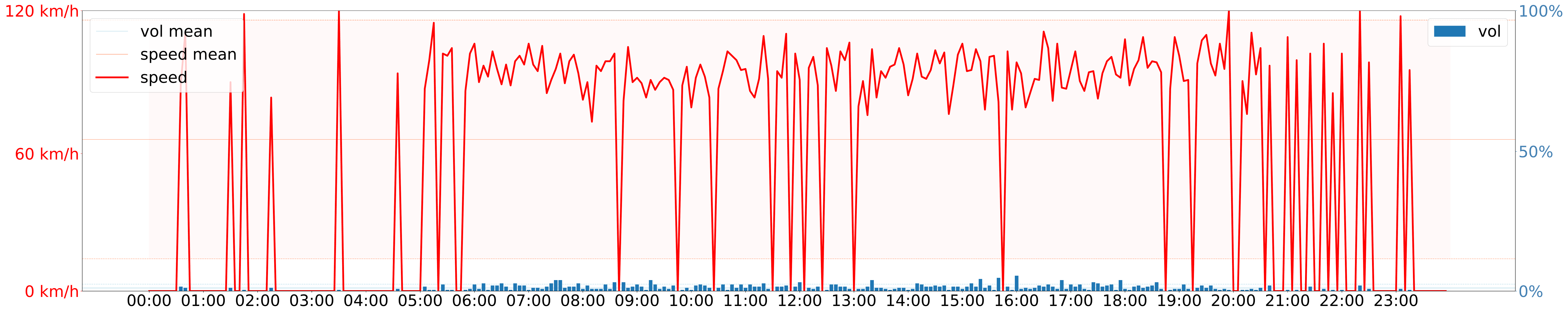}}
  \subfigure[\#5 (359, 416, SW) with low std. Vol: $0.04\pm{}0.73$, speed=$ 0.23\pm{}3.69$]{\includegraphics[width=0.9\linewidth]{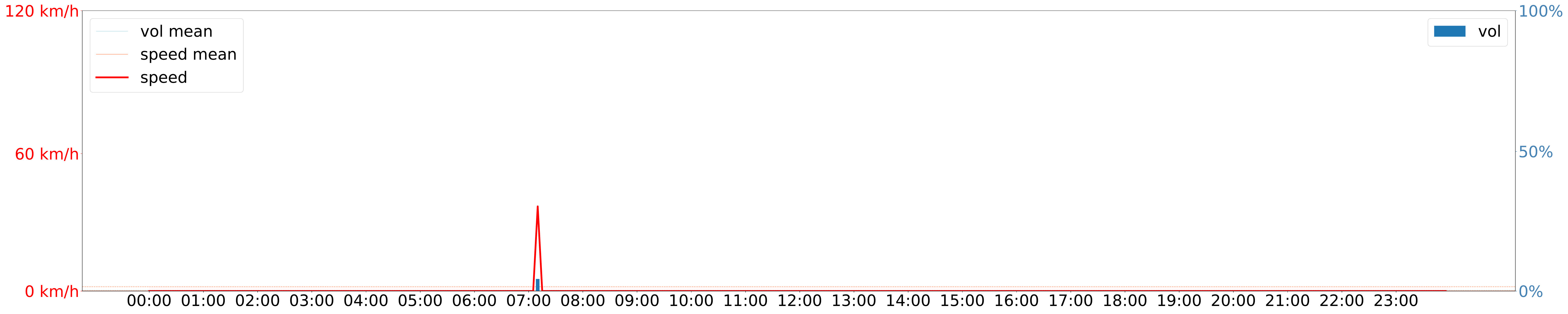}}
  \subfigure[\#6 (319, 307, NE) in 35--55 speed std r. Vol: $5.67\pm{}7.35$, speed=$ 82.94\pm{}54.94$]{\includegraphics[width=0.9\linewidth]{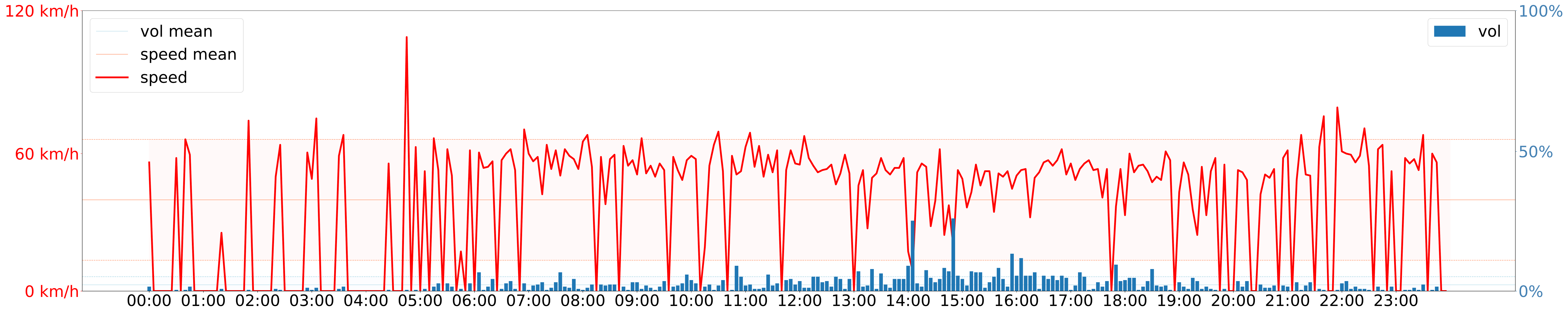}}
  
  \vspace*{-4mm}
  \caption{Data from one day in the target domain of the core competition for the 3 Berlin locations and 3 sampled for low speed std, and the two critical speed std ranges. Speed curve in red and volume bars in blue, along with mean lines and std hulls, mean and std from all test slots of the core competition (input and ground truth).} 
  \label{fig:daylines_mse_std}
  \vspace*{-2mm}
\end{figure}

If we take each cell and heading as a speed detector, we have $495\cdot436\cdot4$ virtual speed detectors. We can now plot the mean against the std speed for data collected from the test slots of the core competition, see Figure~\ref{fig:critical_bands}. 
Each dot corresponds to one cell in the grid and one heading. 
Since a lot of cells never encounter any GPS probes, most of those points are close to the origin of the mean-std coordinates system.
We choose 6 Berlin locations in Figure~\ref{fig:speed_std_heatmap}.
In Figure~\ref{fig:daylines_mse_std}, we plot volumes and speeds for one sample day for each of them and discuss these illustration with respect to the two critical ranges.
\begin{description}
    \item[\#1] is a highway in the outskirts has high speed mean and high speed std. This reflects the high speeds during day time and the data sparsity during night time. \#1 is close to the 105--125 critical range; it has many no-data points and the average speed is even higher, so  every no-data is penalized a lot if the prediction is too high.
    \item[\#2] is a main ringroad highway, has high speed mean and relatively high speed std. Although, there is more data in the night, the sporadic speed drops make standard deviation still pretty high. Pretty high free flow speed with a few or no-data points is hard to guess, as every no-data is penalized considerably if the prediction is too high. The zero-volumes during the first hours of the day might be a production artifact, but the location has sparse traffic during the night in all other days we checked.
    \item[\#3] is a boulevard in the center of Berlin, which shows that moderate free flow speed and frequent no-data can make the prediction task hard, too. 
    \item[\#4] is an example from 105--125 speed std critical range, which translates visually into a very spiky speed curve over day. Notice that volumes are low and the spikes mainly come from the no data points. The variance in the speed measurements masked on non-zero volume would be much lower.
    \item[\#5] is an example of very low speed std. It shows that there are many directional pixels not or extremely sparsely covered with traffic data, either due to the fleet bias (not covering full traffic) or because there is no traffic in those regions at all.
    \item[\#6] is similar to \#3, but with slightly higher volumes, lower speed levels and a clearer difference between day and night traffic, which is plausible for a location on the main arteria but not on a highway.

\end{description}


\clearpage
\subsection{Relating vol variance to vol MSE}
\begin{figure}[htbp]
    \centering
    \includegraphics[width=0.95\textwidth]{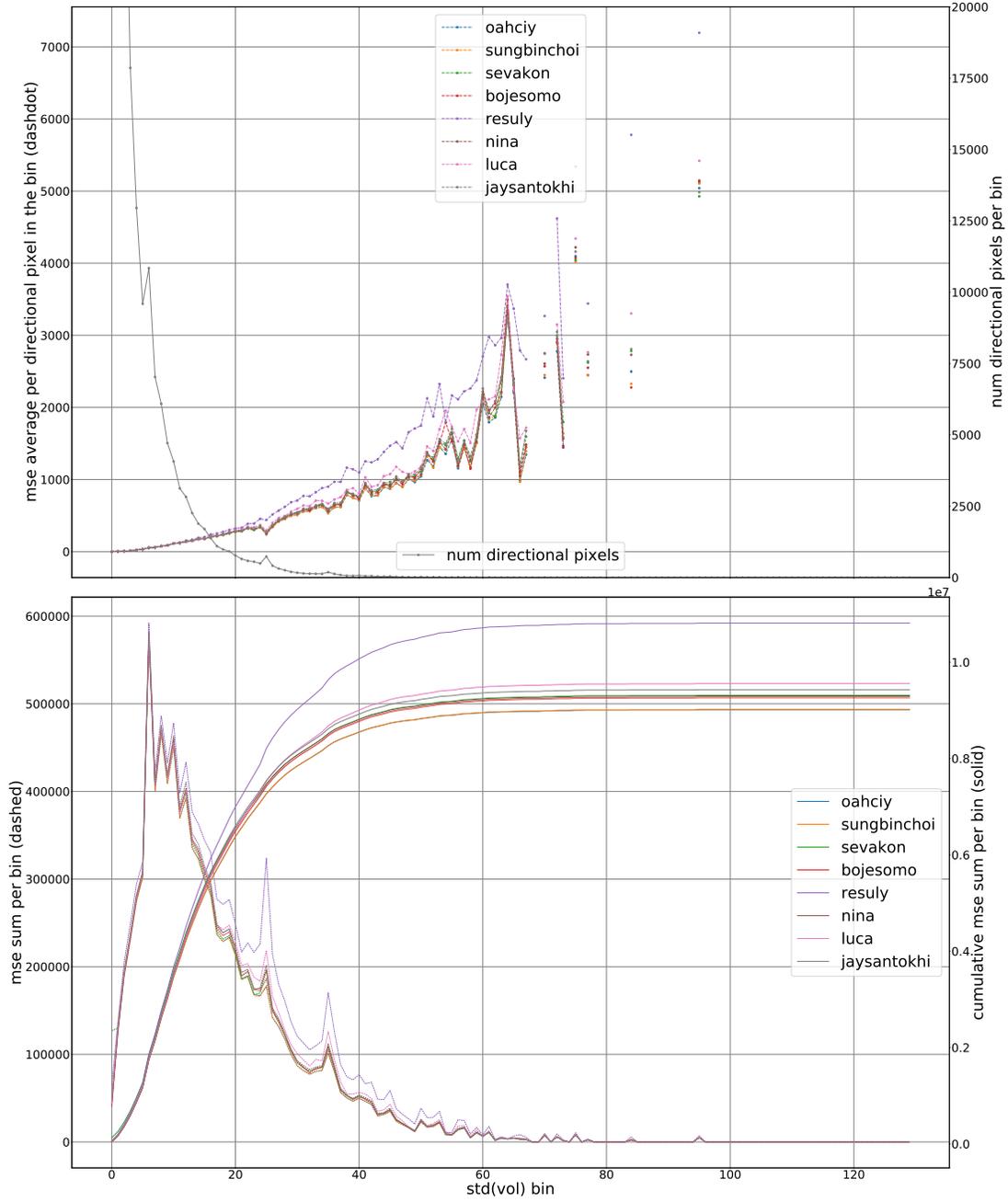}
    \vspace*{-2mm}
    \caption{Relating MSE to std for volumes: distribution of std among oriented pixels and average MSE (top); summed MSE and cumulated summed MSE (bottom). The shaded gray areas highlights the two critical speed std ranges.}
    \label{fig:mse_std_vols}
    \vspace*{-2mm}
\end{figure}

The corresponding plot for volumes is Figure~\ref{fig:mse_std_vols}.

\subsection{Limitations of the analysis and future work}

As remarked in Section~\ref{sec:results_performance_difference}, MSE does not optimize each directed location independently, so the interpretation here has to be taken \textit{cum grano salis}. 
Despite this limitation, we think it is important to relate 

\begin{itemize}
    \item the binning of the directional pixels is based on std in the test slots (input hour and ground truth predictions), it should be checked whether this binning is robust when taking not only the sampled data for the test slots;
    \item conduct the analysis for other cities as well;
    \item in the analysis, the mean or percentiles in the data should be taken into account to put the interpretation of related the traffic patterns onto a more solid basis;
    \item  in the same spirit, the analysis should be carried for day and night time separately and compared to the full day.
    \item In addition to looking at raw MSE values, it might also be interesting to calculate the per-pixel $R^2$ values for different models. 
\end{itemize}

\clearpage
\section{Outlier Special Prize}\label{appendix:SpecialPrize}
Here, we give more details on the Outlier Special Prize introduced in Section~\ref{sec:results_outliers}.

\subsection{Outlier Special Prize}
Quality of traffic prediction heavily relies on the performance in anomalous situations. We aim to have a first look at how models from the t4c21 perform in those situations, both quantitatively as well as qualitatively. 
Even  without  sparsity  due  to  vehicle fleets  nor  daytime  nor regional sparsity, outliers are rare in the data and hence MSE  does  not  give  a  lot  of  weight  to  them.   Also,  typically there are many plausible future scenarios in an outlier situation with a fat tail of scenarios. 
We invited all Summit/Symposium Participants of \t4c 2021 to re-run their models on a new test set for the \t4c 2021 Outlier Special Prize.
Every participant was allowed to submit two predictions:
one with the "plain vanilla" model, as used for the original
submission, 
and one with further training applied. The score was not disclosed to the participants and re-submission was not allowed.
The participants were asked to make a prediction on the full without disclosing the outlier location nor the heading.
As in the core and extended competitions, only time of day and day of week were disclosed.
Also, we did not disclose whether the new slots would before or after the Covid shift.

\subsection{Outlier Heuristics}

We tackled outliers in a very pragmatic way. We were not interested in a general definition of outliers, but very much in finding some examples and plot what the models predict in these situations.
The outcome of a few trial-and-error iterations was a heuristic that finds outlier situations in a single directional cell.

We  compute  quantiles  on  the  data  for  each  pixel  and  each  channel  separately and use them for a lower and an upper threshold criterion, respectively. 
Also, in order to focus on situations were a continuous flow is hold down by some jam situation, we focus on 8am to 8pm. In order to exclude situations too short to be sure whether they are false positives, we search for situations were the above criteria hold for at least two consecutive time bins. And finally, we take also the 2h mean speed and mean volume into consideration.
We choose outlier situations from 9 Tuesdays in September and October 2019 and two cities from the core competition (Berlin and Istanbul, 100 tests for each city).  This should give us situations without temporal shift with respect to the training data, neither due to Covid nor due to summer holiday.   

In summary, we search and select pixels (just one speed and volume channel) using the following conjunctive filter criteria:
\begin{enumerate}
    \item volume and speed quantiles (volume above 90\% quantile, speed below 5\% quantile)and volume threshold (volume above 5)
    \item time between 8 A.M. 8P.M.
    \item outlier duration (at least two consecutive time bins)
    \item  outlier mean volume and speed (outlier mean volume above 1.5 times 2h mean, outlier mean speed above 0.7 times 2h speed mean)
\end{enumerate}
These criteria might be highly redundant and we did not check for a minimal set.
These  criteria  are  supposed  to  give  us  situations  where  volume  is  high  and  speed  is  low in contrast to normal situation as reflected by full-day quantiles and 2h means and which are robust (two consecutive positives and excluding sparse situations during night time).
Formally, an outlier is a quintuple (row, column, heading, start time of day, duration).

For each test, we keep a mask that evaluates MSE on one pixel and two channels, the volume and speed channels of the outlier heading.

\subsection{Test Slot Distribution}

The procedure just described was applied on 9 Tuesdays in September/October 2019 before Covid temporal shift.
Due  to  data  scarcity,  the  above  criteria  were  not  applicable  to  Chicago  and  Melbourne from the core competition,  so we excluded them as we were interested in the qualitative analysis of the situations.  
We randomly sample 100 outliers situations per city from 559 (Berlin) and 1266 (Istanbul). The other two cities of the core competitions turn out not to match the pre-conditions of the heuristics and are not included in the Outlier Special Prize.
We generate tests such that the first outlier time bin is the last bin of the 1 hour test input window.

\begin{figure}[htbp]
  \centering
  \subfigure[Berlin]{\includegraphics[width=0.45\linewidth]{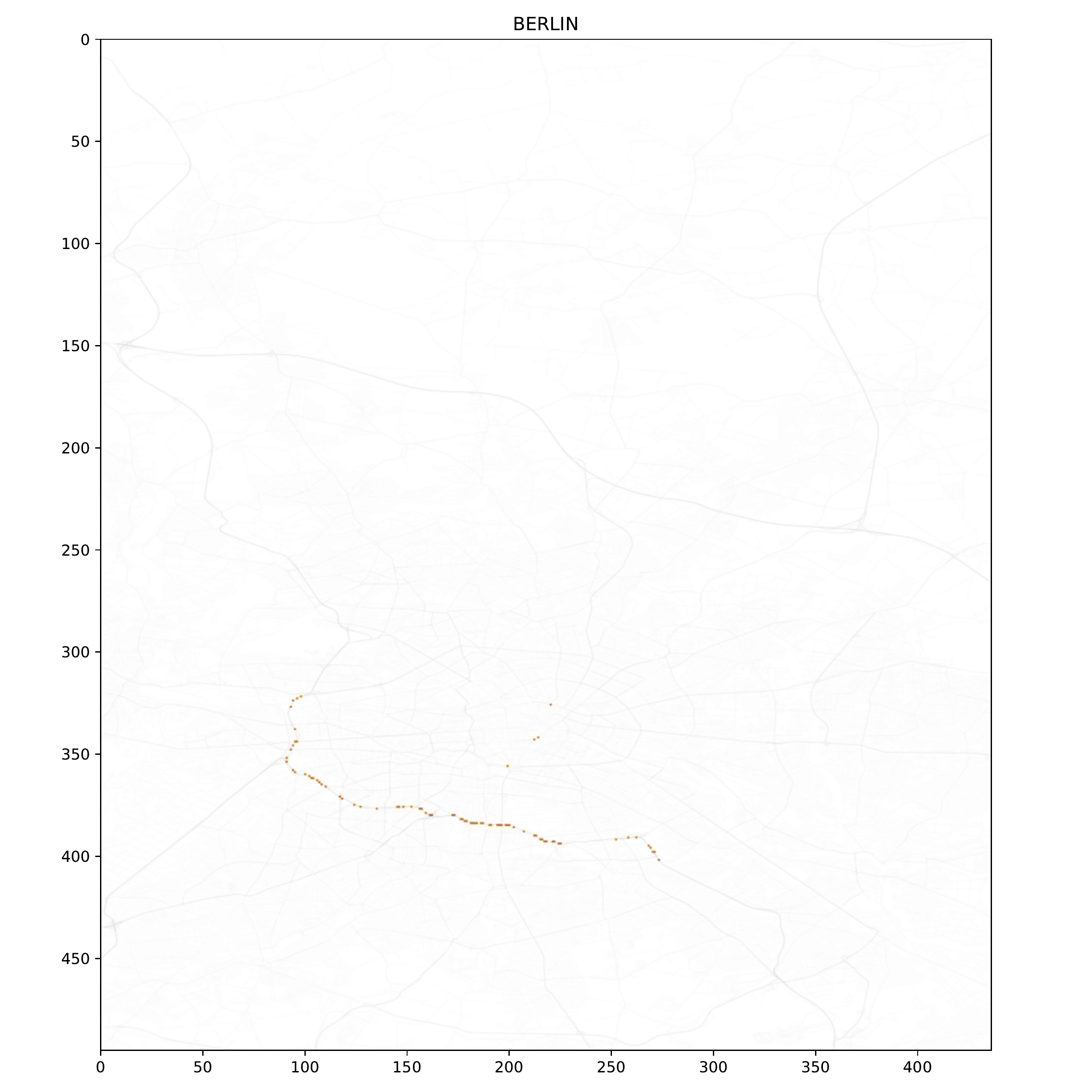}}
  \subfigure[Istanbul]{\includegraphics[width=0.45\linewidth]{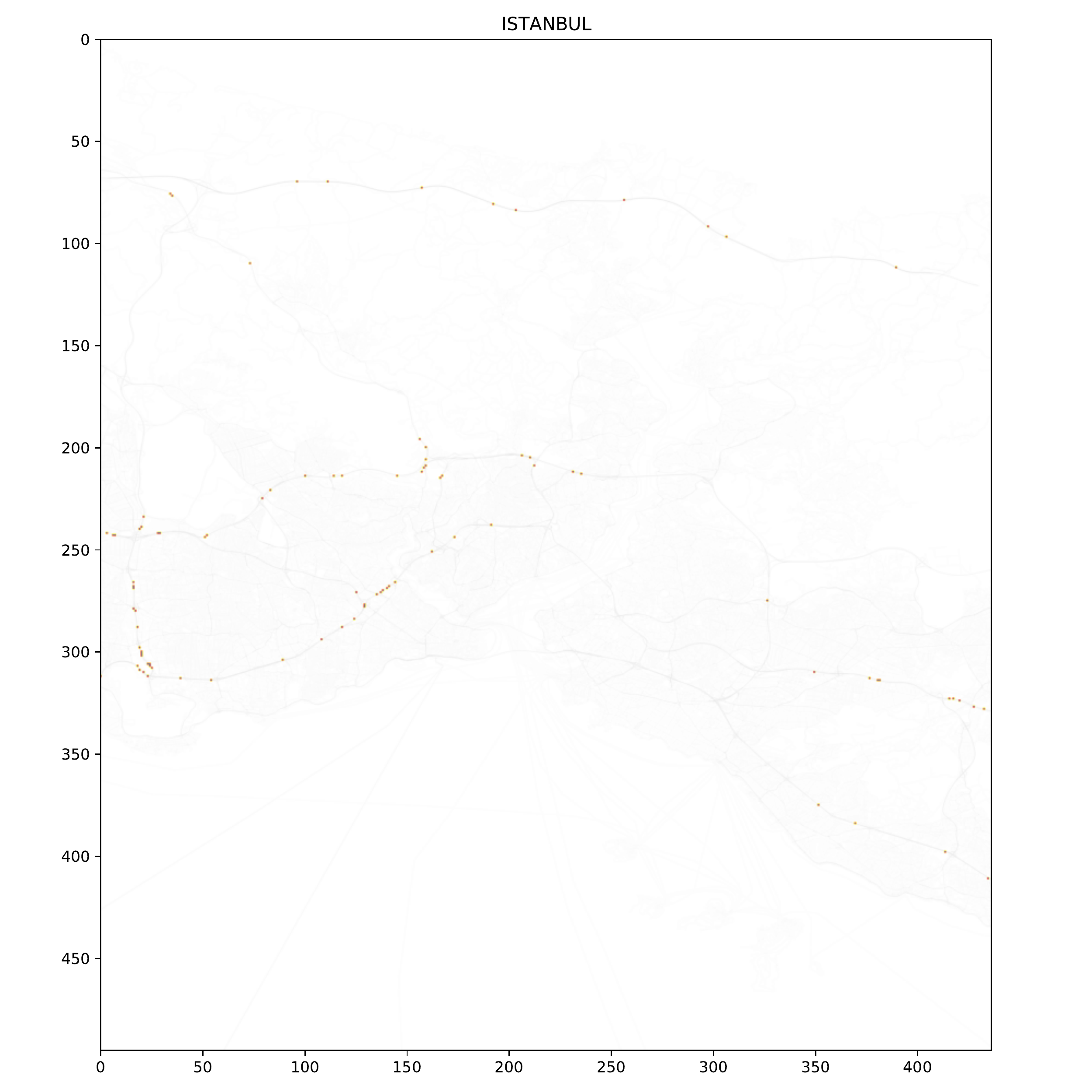}}
  \vspace*{-2mm}
  \caption{Spatial Distribution of Special Prize Tests.}
  \label{fig:specialprizespatial}
\end{figure}

\begin{figure}[ht]
  \centering
  \subfigure[Berlin]{\includegraphics[width=0.9\linewidth]{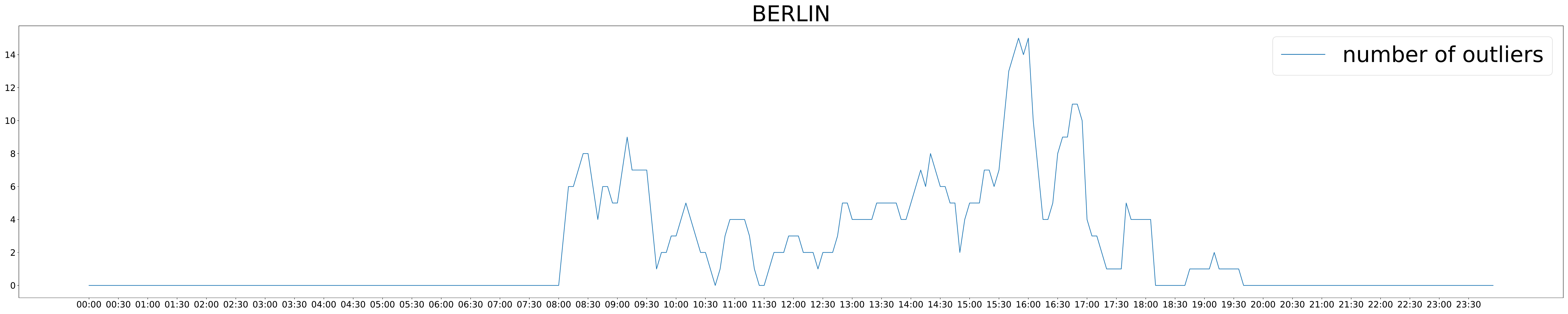}}
  \subfigure[Istanbul]{\includegraphics[width=0.9\textwidth]{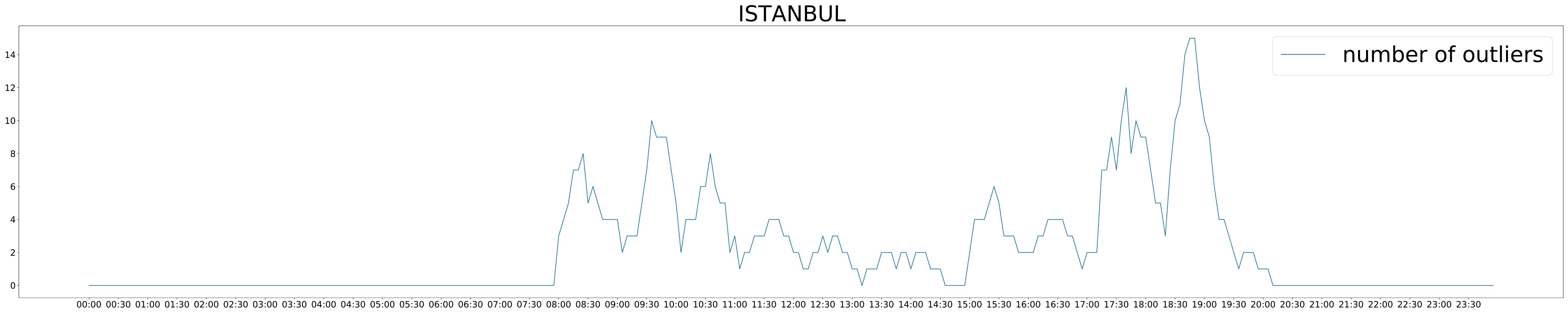}}
  \vspace*{-2mm}
  \caption{Temporal Distribution of Special Prize Tests.}
  \label{fig:specialprizetimeofday}
\end{figure}

The spatial and temporal distribution of the slots are shown in
Figures~\ref{fig:specialprizespatial}--\ref{fig:specialprizetimeofday}, respectively. As expected, the spatial distribution is concentrated along main arteries with continuous flow, and the temporal distribution reflects the morning and afternoon peak hours.

\subsection{Outlier Special Prize Submissions and Leaderboard}

We invited all Summit/Symposium Participants of \t4c 2021 to re-run their models on a new test set for the \t4c 2021 Outlier Special Prize, and from 8 summit participants invited, 7 participated with a total of 11 submissions. 4 participants submitted the solution from re-running their core competition best model, but also submitted a second solution:
\begin{description}
    \item[oahciy\_v1] is obtained by exactly the same models that as for the
core competition \cite{lu2021learning}.
    \item[oahciy\_v2] is obtained by further fine-tuning the models for 1 epoch on
BERLIN and ISTANBUL data only.
]
    \item[ai4ex\_36]  Epoch=36: This is exactly the submissions for the competition (plain vanilla) \cite{bojesomo2021hierarchical}.
    \item[ai4ex\_43]  Epoch=43: further training the model.
    \item[nina\_orig] the predictions
of the original model (Unet++ patch-based prediction with 100x100 patch and
stride s=10 as explained in \cite{wiedemann2021traffic}). 
    \item[nina\_special]  trained the
original model further on the two cities separately. In detail, they loaded
the model weights, then trained for 500 further epochs on Berlin, and
predicted the values for the new Berlin test set. Same for
Istanbul (starting again from the weights of the original submission).
    \item[sungbin\_1] is from best run models on core task \cite{choi2021applying}.
    \item[sungbin\_2] is from best run models on extended task with different training methods which had not used in the core task (data augmentation method: input image flipping).
\end{description}   
The other 3 submissions are jaysantokhi \cite{santokhi2021dual}, GraphUNet\_luca \cite{hermes2021Graphbased} and Bo \cite{wang2021traffic4cast}.

\begin{figure}[ht]
  \centering
  \includegraphics[height=0.8\textwidth,angle=-90]{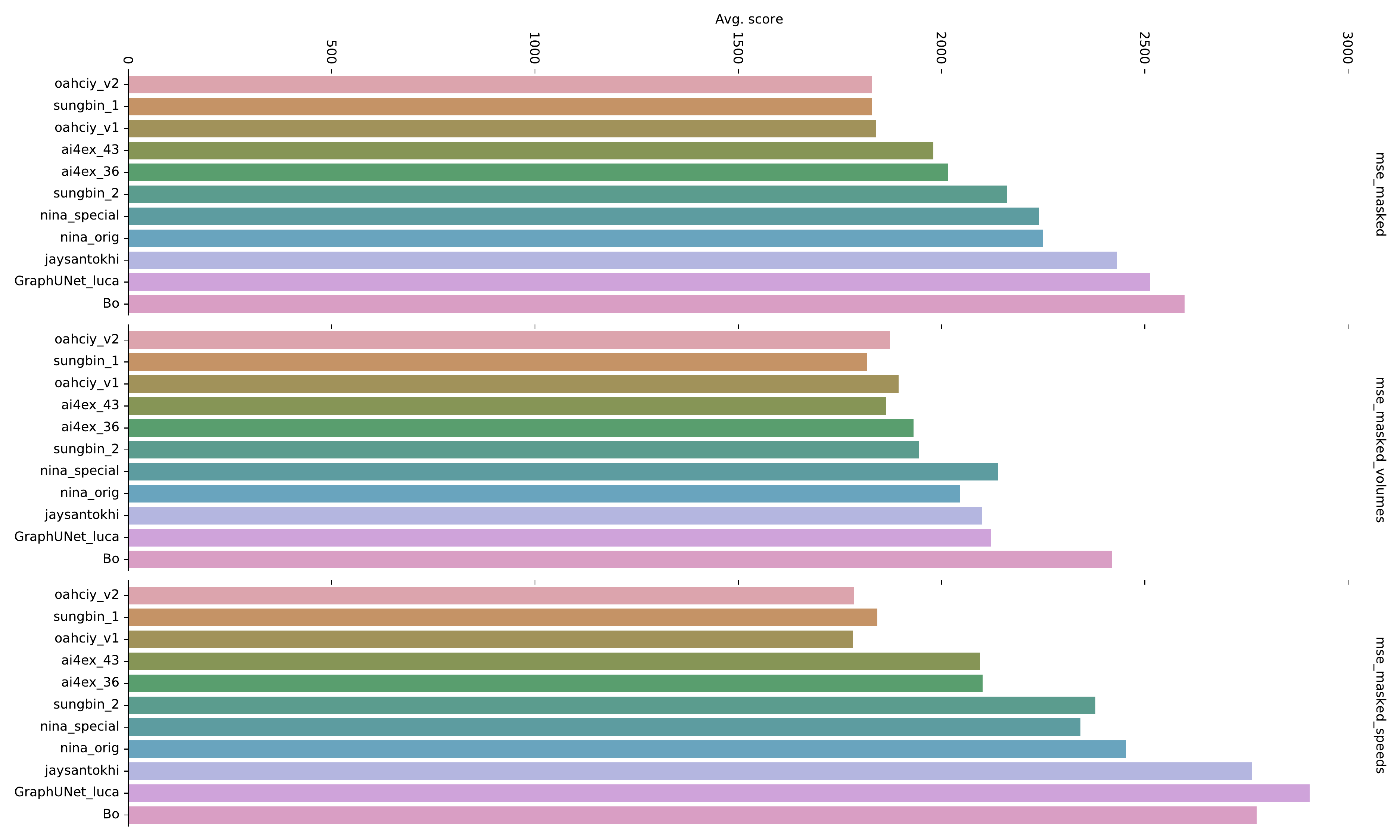}
  \vspace*{-4mm}
  \caption{Leaderboard special prize (anomalies) based on masked MSE highlighting one cell and one heading.}
  \label{fig:leaderboard_specialprize}
  \vspace*{-2mm}
\end{figure}

\begin{figure}[ht]
  \centering
  \includegraphics[height=0.8\textwidth,angle=-90]{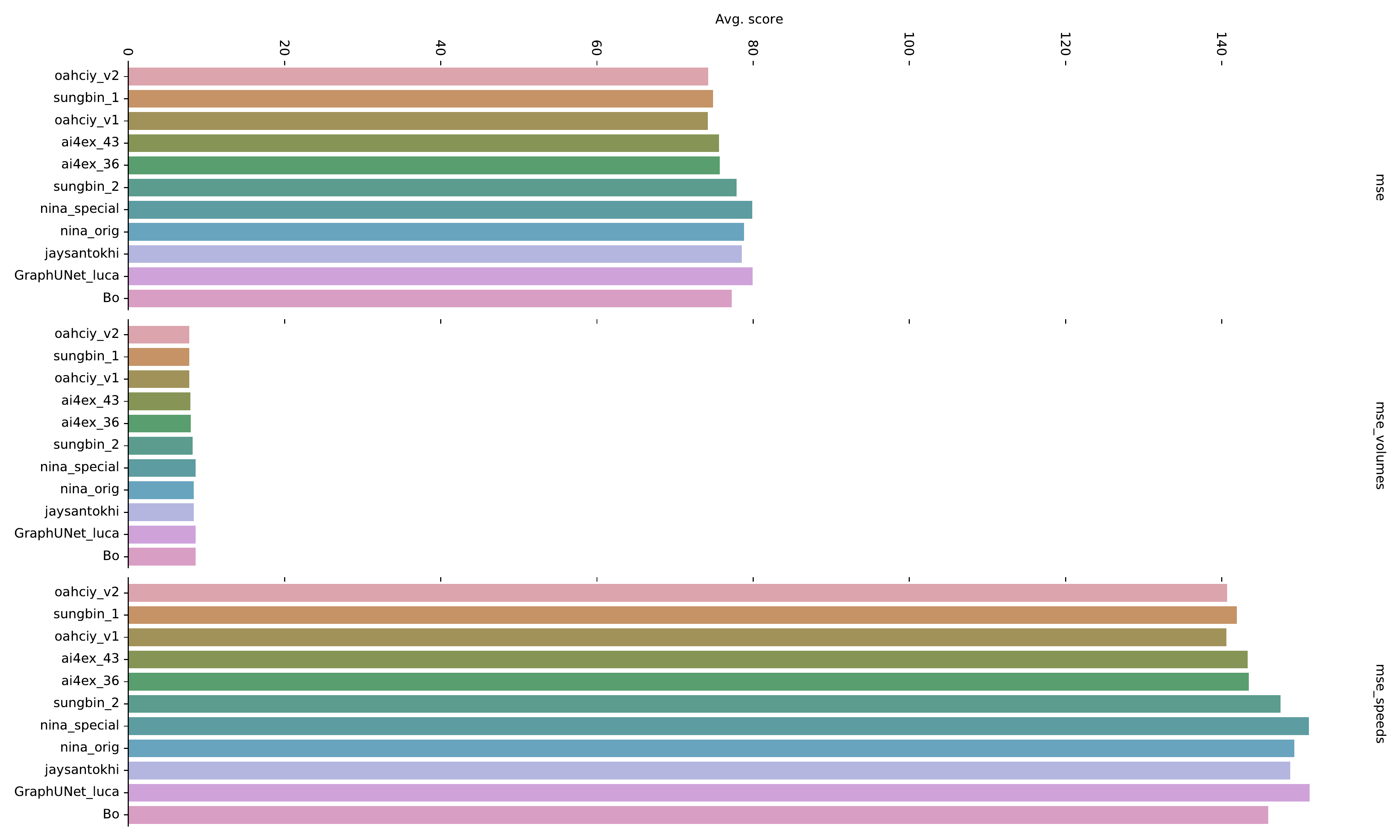}
  \vspace*{-4mm}
  \caption{Unmasked MSE in special prize (anomalies) based on masked MSE.}
  \label{fig:leaderboard_specialprize_full_mse}
  \vspace*{-2mm}
\end{figure}

For each test, we keep a mask that evaluates MSE on one pixel and two channels, the volume and speed channels of the outlier heading, see Figure~\ref{fig:leaderboard_specialprize}.
Hence, the quantitative evaluation takes place on much less data than the full city (as in Figure~\ref{fig:leaderboard_specialprize_full_mse}):
We observe that 
oahciy\_v2 \citep{lu2021learning}, sungbin\_1 \citep{choi2021applying} and oahciy\_v1 \citep{lu2021learning} are very close on masked MSE.
Surprisingly, sungbin\_2 \citep{choi2021applying} based on the winning extended challenge is much poorer than sungbin\_1, so it seems that sungbin\_2 is too ``conservative'' in this setting where the best guess in most situations is predicting a normalization of the situation. 
The solutions ai4ex\_43/aiex\_36 \citep{bojesomo2021hierarchical}, nina\_special/nina\_orig \citep{wiedemann2021traffic}, jaysantokhi \citep{santokhi2021dual}, GraphUNet\_luca \citep{hermes2021Graphbased} and Bo \citep{wang2021traffic4cast} are clearly beaten in the Special Prize challenge as well.

As a sanity check, Figure~\ref{fig:leaderboard_specialprize_full_mse} shows the unmasked MSE, which again shows similar level of MSE overall and for volume and speed separately as in the core competition. We see here the same top-3 submissions as in the core competition (oahciy\_v1 marginally better than oahciy\_v2, Sungbin\_1, ai4ex\_36, the third prize winner did not participate in the Special Prize). Sungbin\_2 trained for unseen cities is clearly inferior to Sungbin\_1 in both evaluations.

\subsection{Special Prize Qualitative Analysis}\label{appendix:SpecialPrizeQual}

Out of the 200 tests (100 for Berlin and Istanbul each), we choose 4 sample situations (see the outlier locations in Figure~\ref{fig:outlier_pixel_locations} and describe the anomalies qualitatively:

\begin{figure}[ht]
  \centering
  \subfigure[Berlin]{\includegraphics[width=0.8\linewidth]{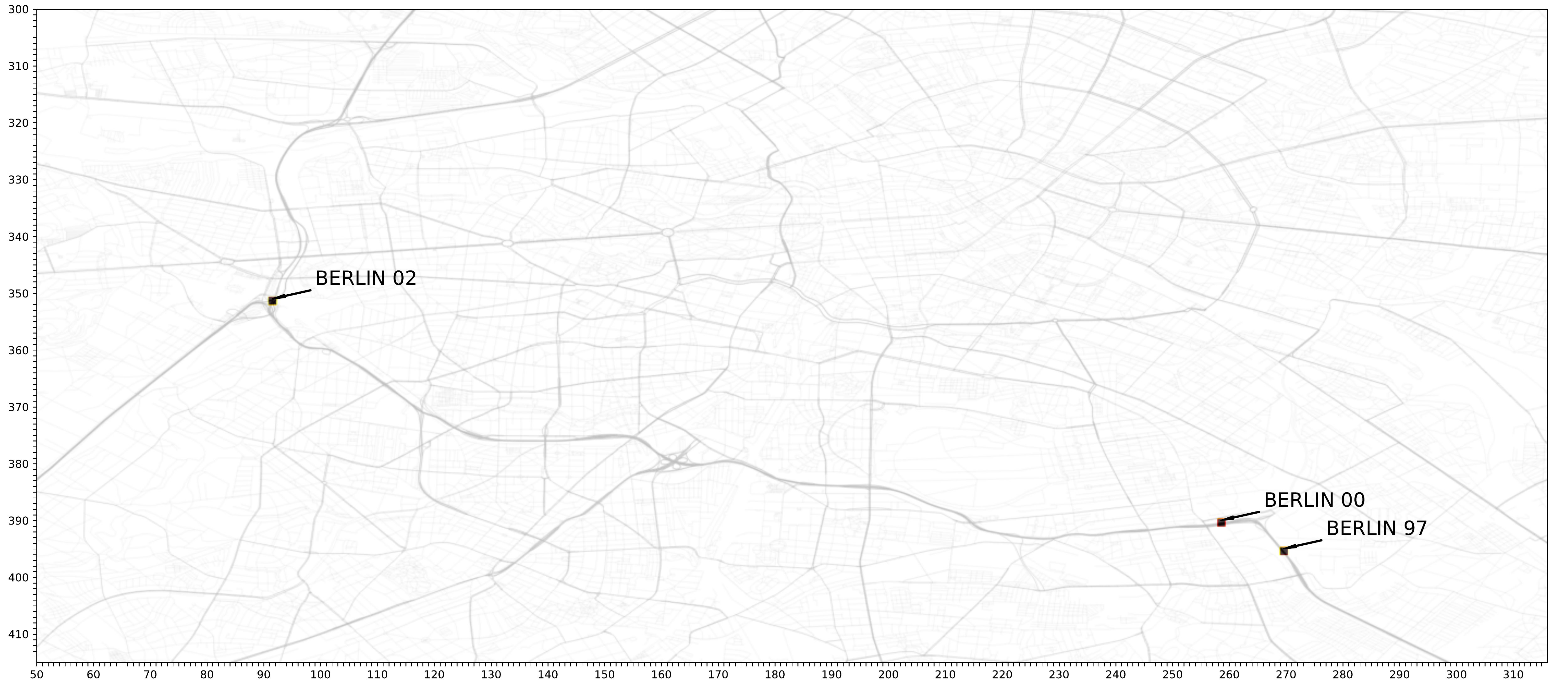}}
  \vspace*{-2mm}
  \subfigure[Istanbul]{\includegraphics[width=0.8\linewidth]{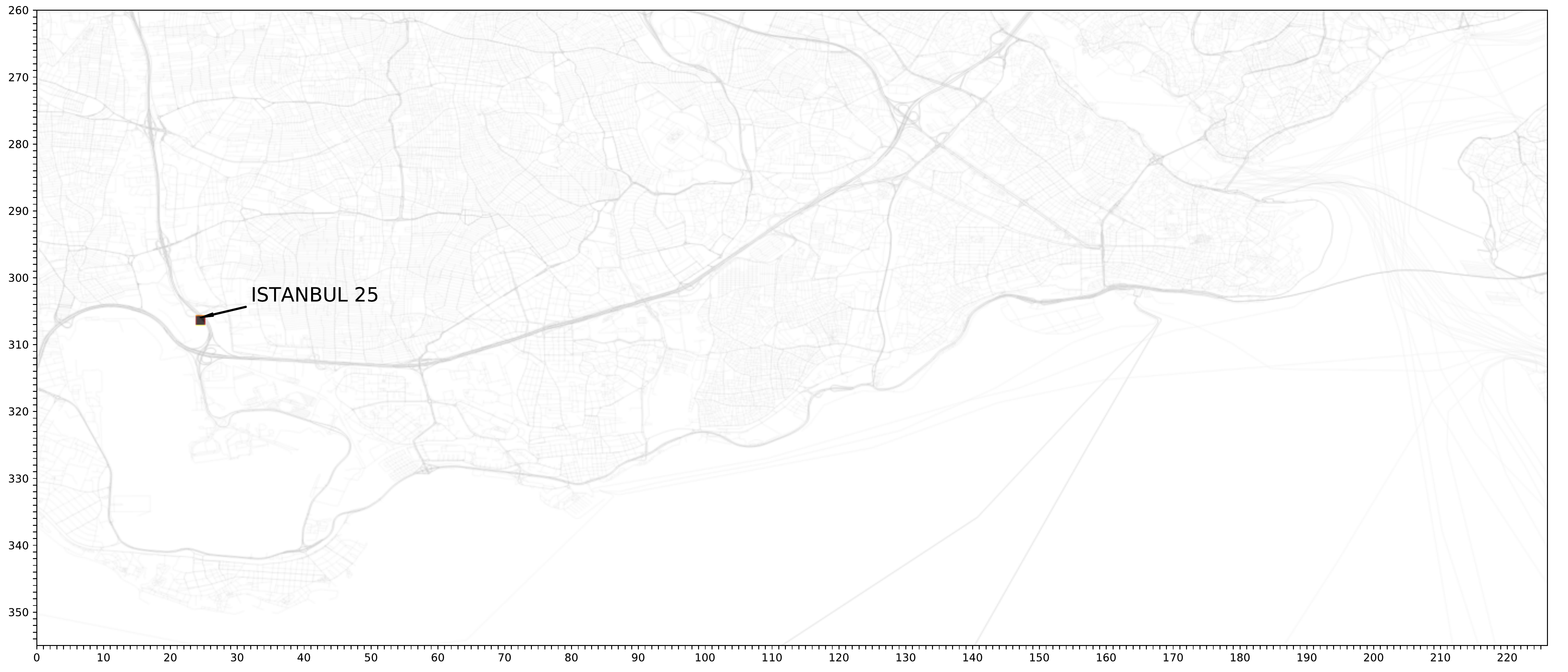}}
  \vspace*{-2mm}
  \caption{Sample situations}
  \label{fig:outlier_pixel_locations}
  \vspace*{-2mm}
\end{figure}

\begin{figure}[ht]
  \includegraphics[width=0.95\linewidth]{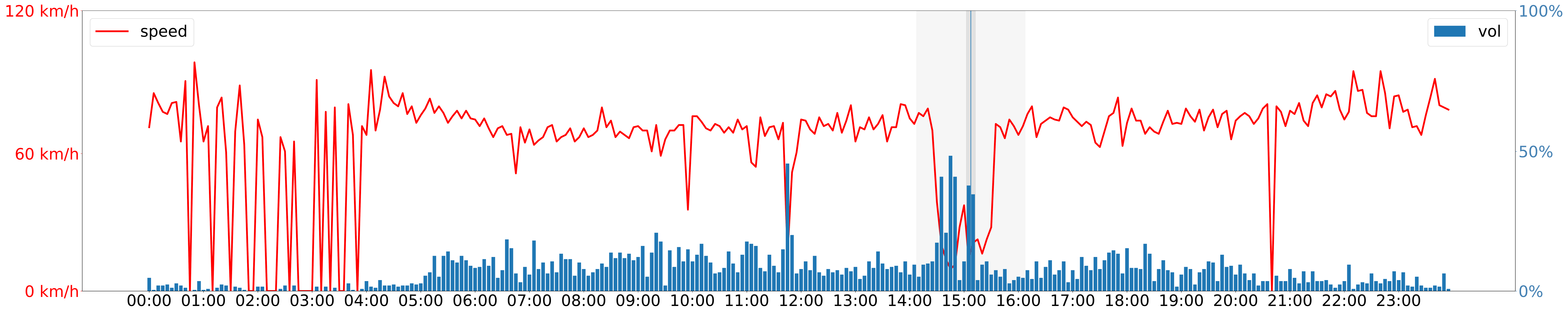}
  \\
  \includegraphics[width=0.95\linewidth]{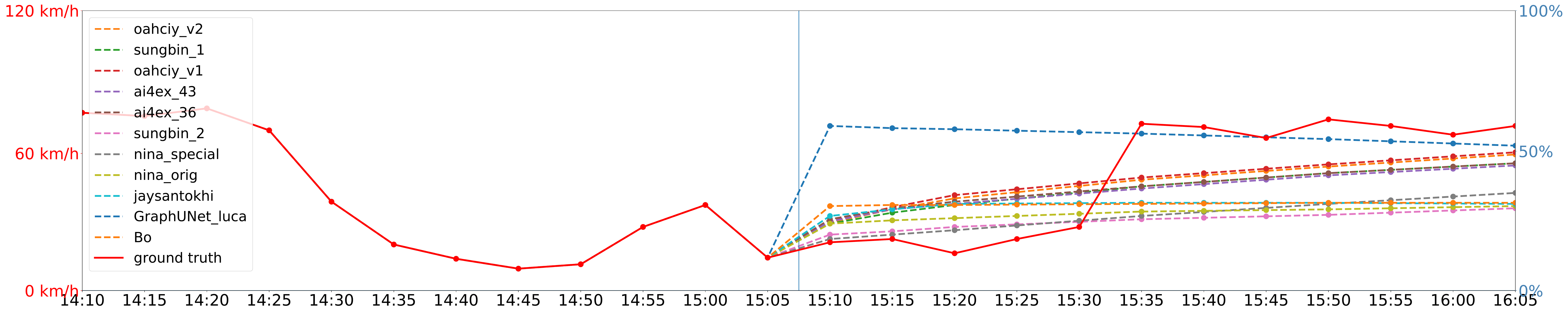}
  \\
  \includegraphics[width=0.95\linewidth]{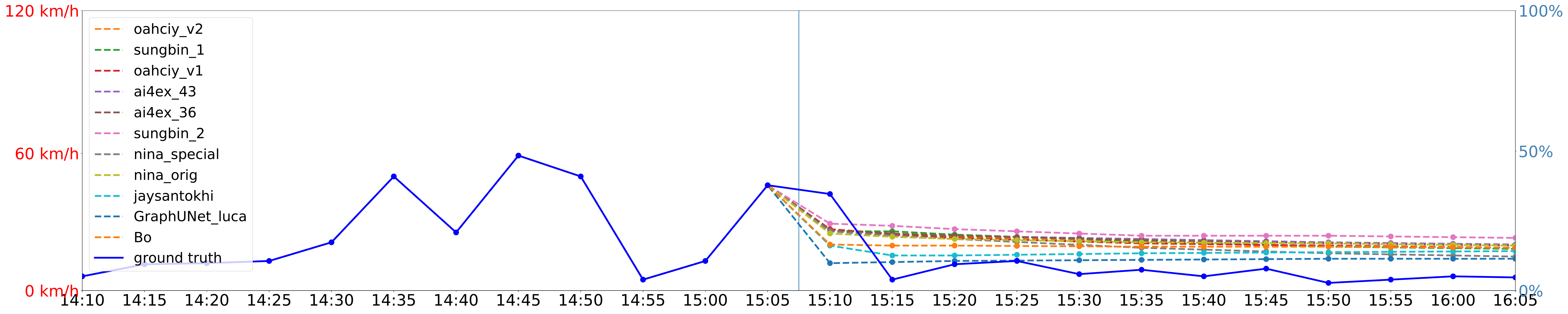}
  \vspace*{-4mm}
    \caption{Predictions Berlin 00. Top: Daytime curve of ground truth data for 288 bins from midnight to midnight with speeds (red curve) and volumes blue bars. Outlier in darker gray area with input and prediction hour in light gray separated by a blue vertical line.
      Middle, bottom: The dashed lines show the speed and volume predictions of the different submissions compared to ground truth (solid lines); the vertical blue line separates input from output in time.}
  \label{fig:specialprizeBerlin00}
  \vspace*{-4mm}
\end{figure}

\begin{figure}[ht]
  \includegraphics[width=0.95\linewidth]{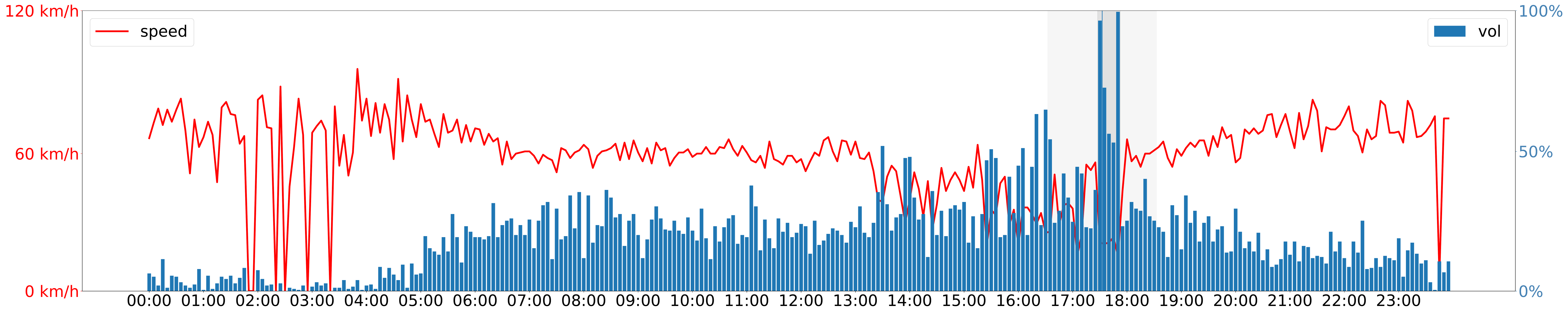}
  \\
  \includegraphics[width=0.95\linewidth]{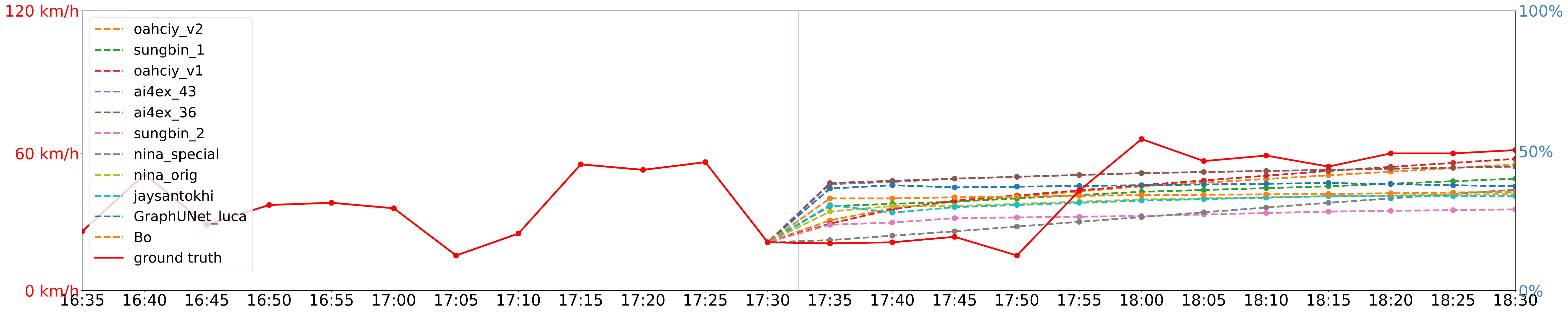}
  \\
  \includegraphics[width=0.95\linewidth]{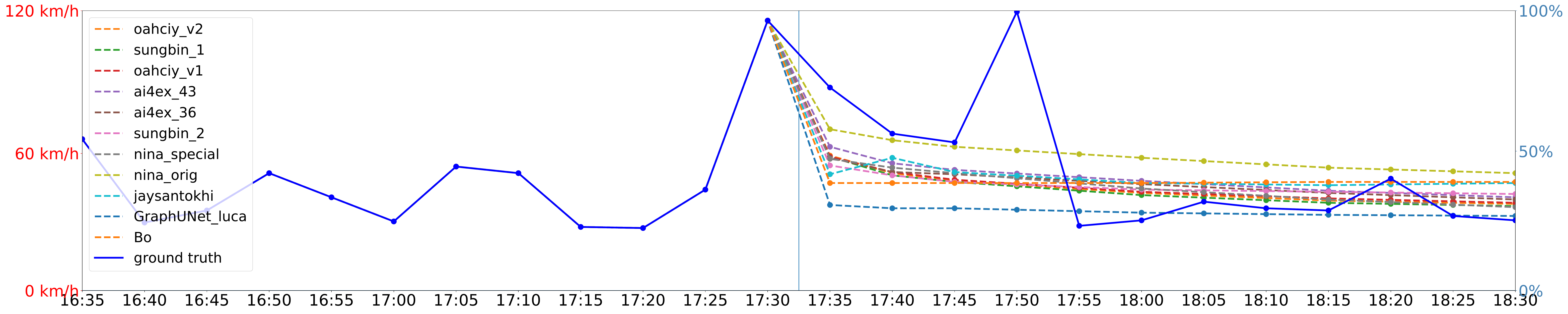}
  \vspace*{-4mm}
    \caption{Predictions Berlin 02}
  \label{fig:specialprizeBerlin02}
  \vspace*{-4mm}
\end{figure}

\begin{figure}[ht]
  \includegraphics[width=0.95\linewidth]{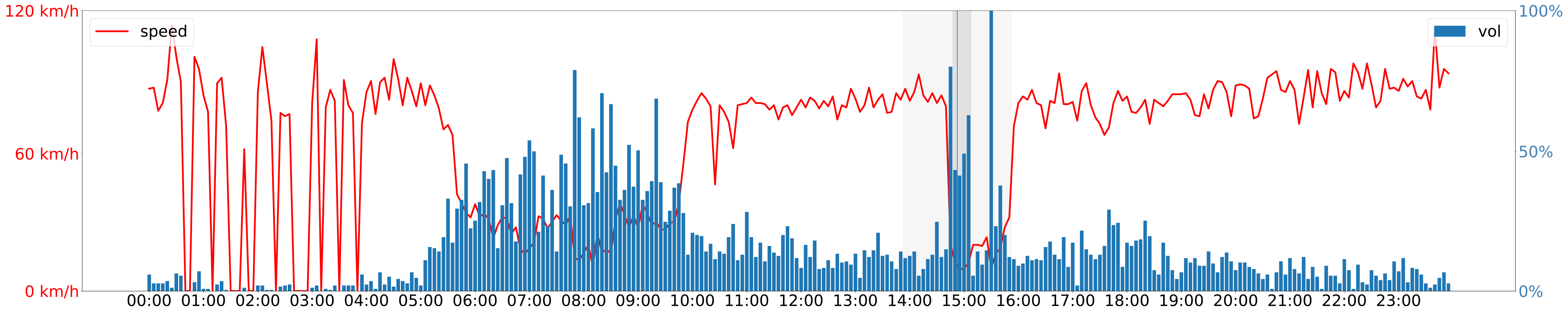}
  \\
  \includegraphics[width=0.95\linewidth]{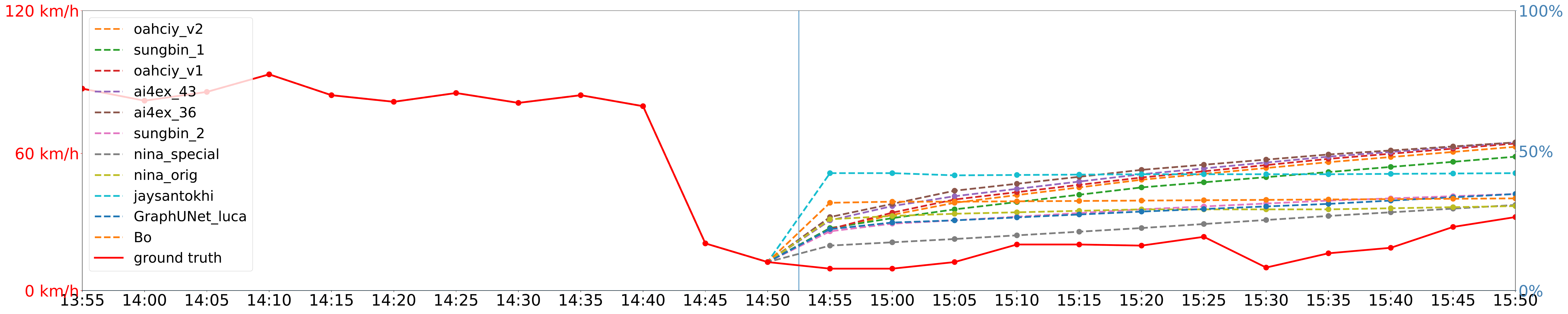}
  \\
  \includegraphics[width=0.95\linewidth]{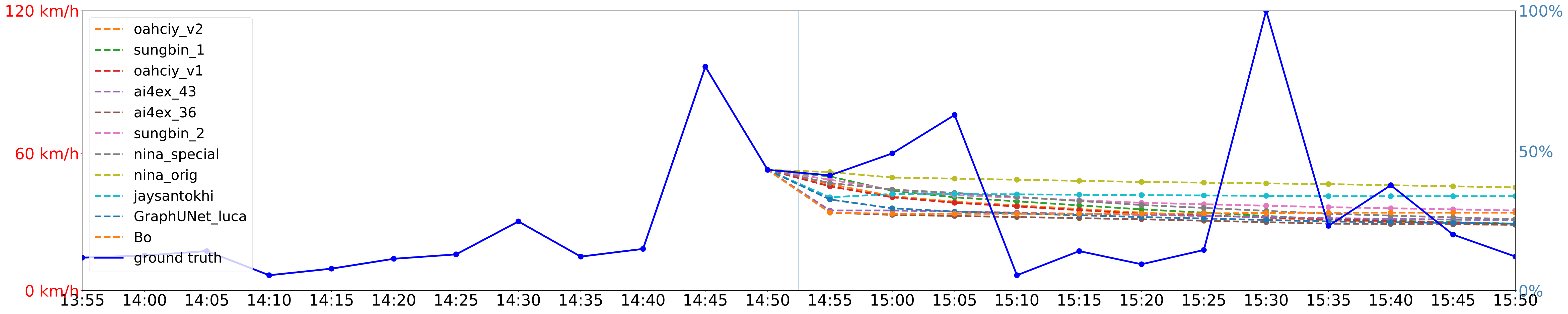}
  \vspace*{-4mm}
    \caption{Predictions Berlin 97}
  \label{fig:specialprizeBerlin97}
\end{figure}

\begin{figure}[ht]
  \includegraphics[width=0.95\linewidth]{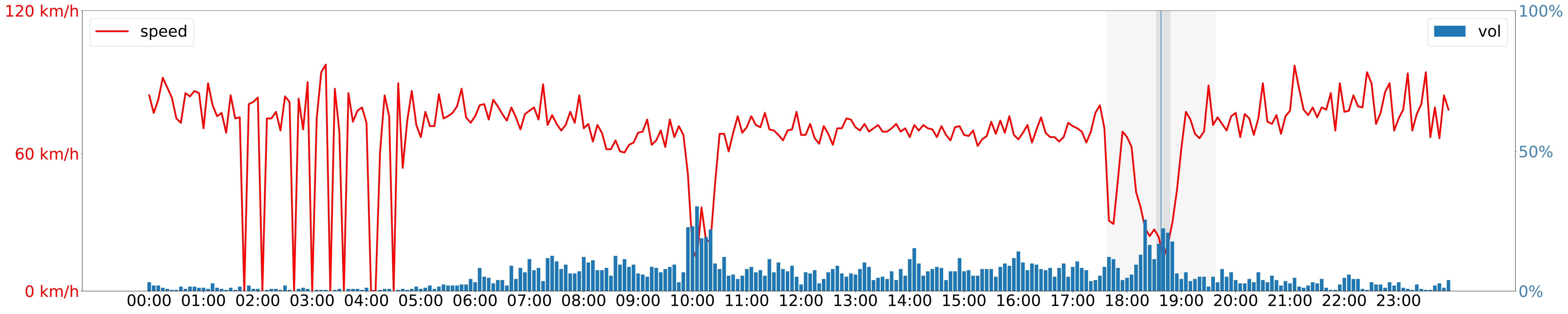}
  \\
  \includegraphics[width=0.95\linewidth]{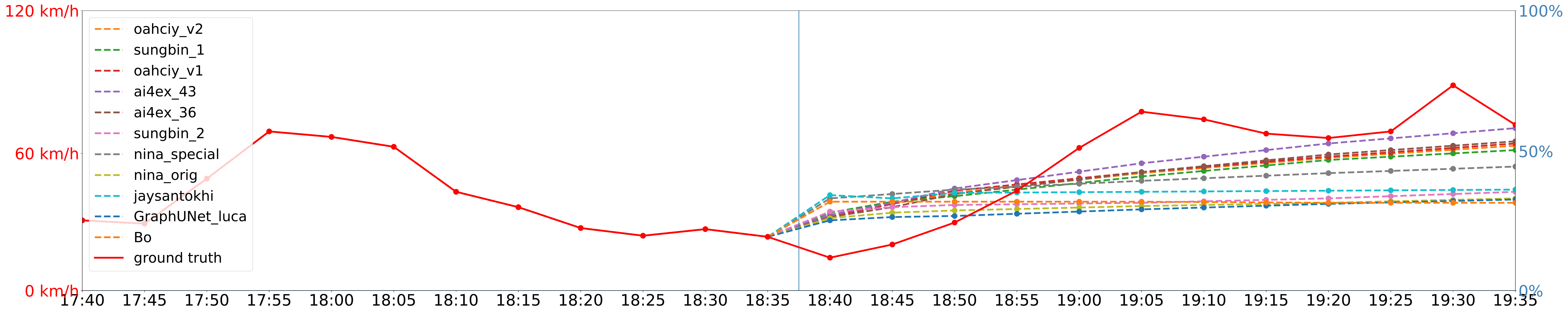}
  \\
  \includegraphics[width=0.95\linewidth]{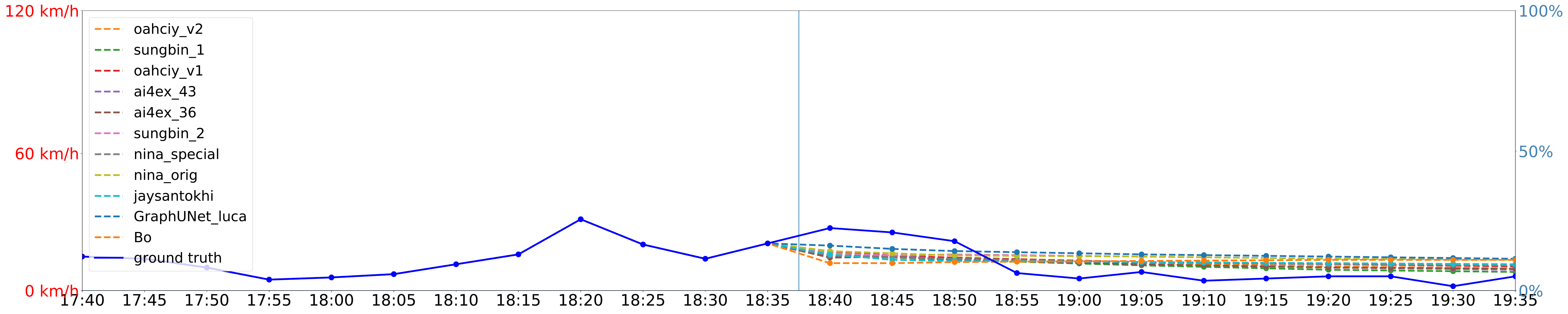}
  \vspace*{-4mm}
    \caption{Predictions Istanbul 25}
  \label{fig:specialprizeIstanbul25}
\end{figure}

\begin{description}
    \item[BERLIN 00] Figure~\ref{fig:specialprizeBerlin00} : the anomaly started 40 minutes before, stop and go during input hour, normalizing in prediction hour after 10 minutes.     We see speed drops during the night and in the evening due to zero volume. The outlier is in the afternoon peak. Prominently, speed has gone down and volume has gone up. The prediction seems to go beyond the mean speed and volume in the input, approximating normalization to ``free flow speed''.
    The highlighted gray area seems to be only at the second half of the jam. However, there was already a partial resolution of the jam, hence our outlier detection heuristic detected two consecutive outlier ane we see only the second sampled here. The winner prediction (oahciy\_v2 \cite{lu2021learning}) suggests a steady normalization of volume and speed.
    \item[BERLIN 02] Figure~\ref{fig:specialprizeBerlin02} unsteady flow during input hour, anomaly started 5 minutes before: peak  just before prediction start, normalizing over 30 minutes
    \item[BERLIN 97] Figure~\ref{fig:specialprizeBerlin97}: anomaly 15 minutes before, the jam does not fully resolve during the prediction horizon and we suspect multiple minor go and stops during the prediction horizon (there are two volume peeks in the prediction horizon with low speeds at the same time, which of course could be due measurement error), with speed going up only slowly at the end of the prediction horizon. In this case, the model again predicts a smooth normalization, but far too soon. If we look at the full day ground truth, however, we see that the jam resolved shortly after the prediction horizon, so in some sense the model anticipated that with one smoothed idealised guess, reflecting multiple scenarios only in a statistical sense.
    \item[ISTANBUL 25] Figure~\ref{fig:specialprizeIstanbul25}: slowdown started 30 minutes before,  going back to normal over 30 minutes
\end{description}
Without a quantitative verification, Figures~\ref{fig:specialprizeBerlin00}--\ref{fig:specialprizeIstanbul25} seem to suggest there are  3 typical behaviours:
\begin{description}
    \item[monotonic normalization] The first type is a monotonic normalization to close to free flow speed as in the two examples shown. This is the normal red level we see before the anomaly starts on the left. And the dashed red line gradually creeps back to that level over the one hour prediction horizon left to the vertical now line. The same happens for volume (blue) which goes down when the jam resolves. However again, the prediction does not fully recover so the model probably learns from the training data to expect to expect not full resolution of the jam over the prediction horizon. 
    They seem to provide a ``smoothing'' of the anomaly going back to normal. Clearly overestimating in Berlin 97, where the situation does not normalize during the prediction horizon.
    \item[monotonic towards mean of input] The second class does predictions which are monotonic towards mean of input. The mean of the input is shown by the faint horizontal red line. So it looks as if in these cases  only consider the local effects in the input. 
    \item[static] The third class does a jump and stay prediction. Some models jump to the mean speed of the input as in the example shown on the right-hand side. Other models of the third class jump even to something close to free flow speed. 
\end{description}

All models seem to predict pretty high volumes from what we would expect as the non-jammed normal volume from the input hours in the left half of the plots.

In summary, the best models predict smoothed version of a jam resolution (Berlin 00), underestimating speed and overestimating density; models are fooled in case jam does not resolve (Berlin 97), MSE makes prediction blurred towards the mean in the data, never predicting a rare scenario such as jam resolving more or less quickly than in expectation; MSE for volume and speed is at the same level.

We do not provide a thorough and systematic exploration of this classification, but we think it still illustrates some shortcomings of the current task formulation discussed in the main text.

\clearpage
\section{Leaderboards Core and Extended Competitions}\label{appendix:leaderboard}
In this appendix, we highlight some aspects of the leaderboard to highlight some features of MSE evaluation in the \t4c setting. The code used to generate them and more plots can be found in \cite{traffic4cast2021-github}.

\begin{figure}[ht]
  \centering
  \includegraphics[width=0.9\linewidth]{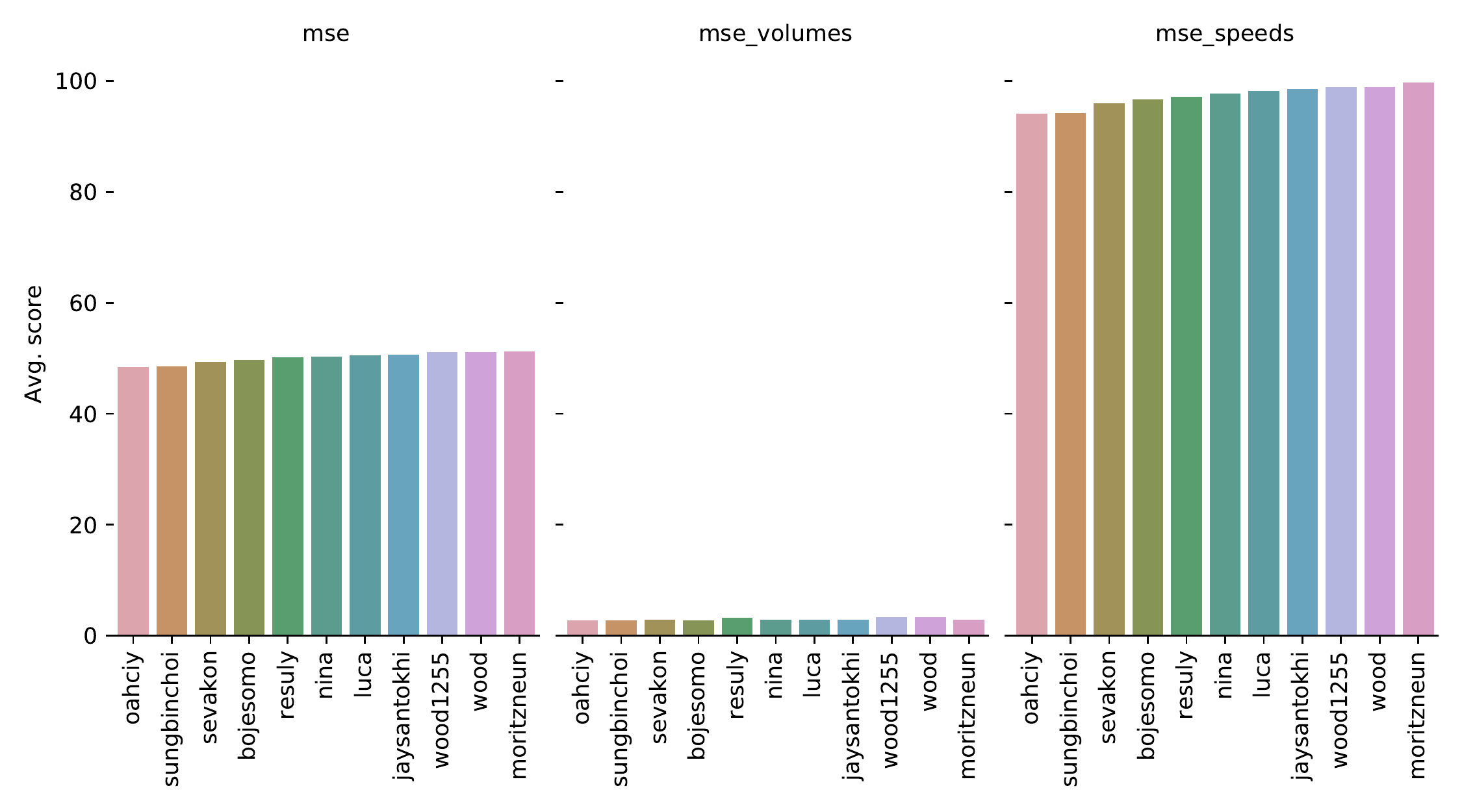}
  \includegraphics[width=0.9\linewidth]{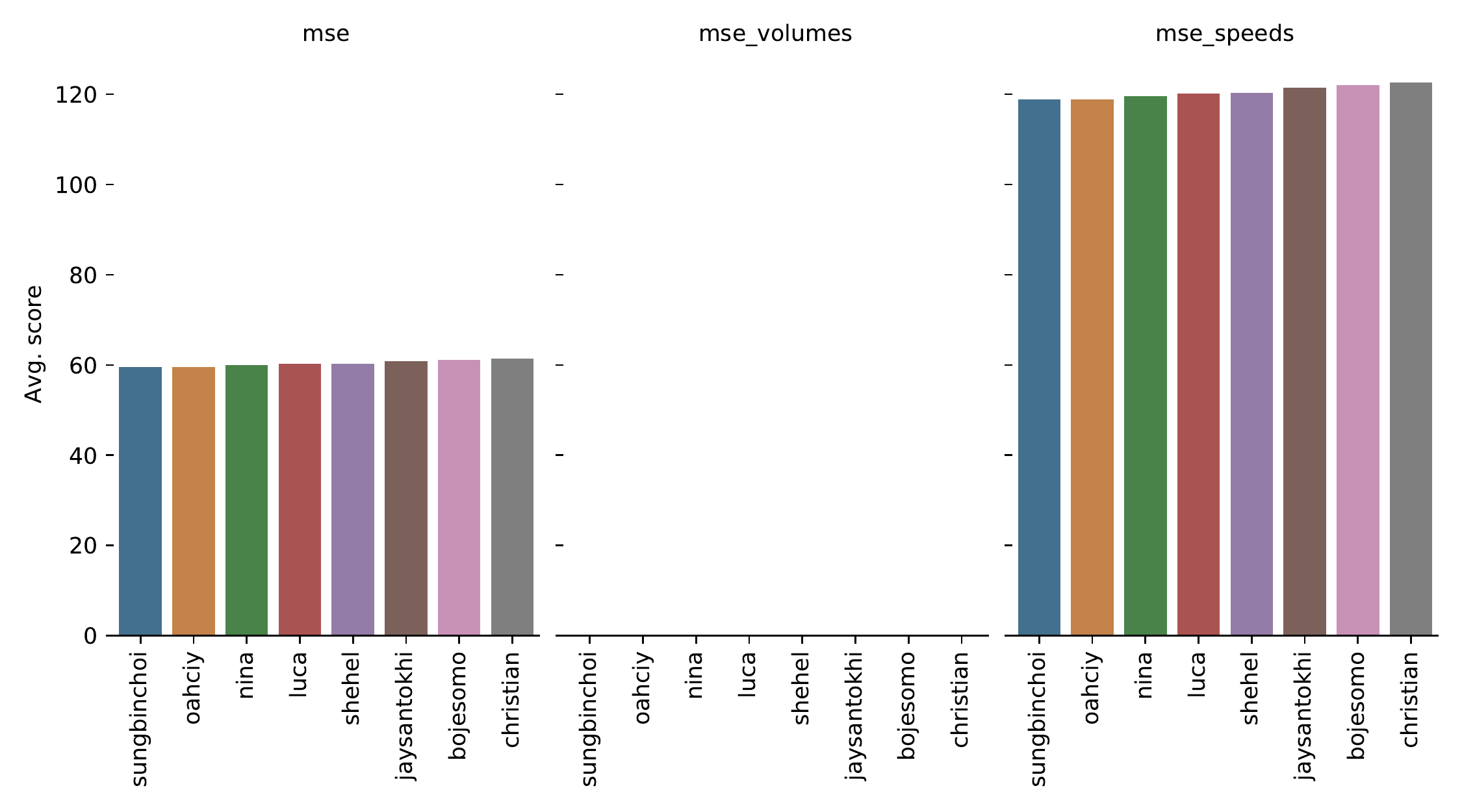}
  \vspace*{-4mm}
  \caption{MSE Volume vs. speed bias: MSE over all channels and MSE on volumes and speeds separately for core competition (temporal shift, top) and extended competition (spatio-temporal shift, bottom). MSE is the average of volumes and speed MSE by definition.}
  \label{fig:leaderboard_speeds_volumes}
  \vspace*{-2mm}
\end{figure}

Figure~\ref{fig:leaderboard_speeds_volumes} shows the dominance of speed channels in both competitions.

\part*{Supplementary Material}
\addcontentsline{toc}{part}{\protect\textbf{Supplementary Material}}

\section{Design of \t4c 2021}\label{sec:desgin_t4c2021}
In this Section, we motivate  the design of \t4c 2021, referring to past \t4c competitions and to the scientific literature.

\subsection{Background and Motivation}
Since 2019, our \t4c competition series at NeurIPS has contributed both methodological and practical insights, advancing AI-driven traffic forecasting and research on the general applicability of the resulting methods to predicting other spatial processes.
Although such traffic forecasts are thought to form the basis for building and managing our cities to provide efficient and sustainable mobility \cite{bucher2019location,jonietz2018identifying,lee2018stochastic}, this form of traffic prediction is still largely considered to be an unsolved problem~\cite{guo2019attention}.

In our first edition of \t4c competition at NeurIPS 2019, we encouraged contestants to predict traffic flow volumes, velocities, and dominant flow directions 15 minutes into the future on a unique, large, real world data set~\cite{Kreil_Traffic4cast_2019}. 
A key innovation of our \t4c competitions was the chosen representation of traffic: we aggregated the data from individual sensor measurements in space and time bins. The values of the spatial representation of each time bin could be interpreted and visualized as a `movie' frame, thus effectively recasting the traffic prediction task as a video frame prediction task~(see \cite{Kreil_Traffic4cast_2019,Kopp_Traffic4cast_2020}). These Traffic Map Movies, for the first time, provided a high resolution privacy preserving comprehensive view of urban traffic.  
The design of the $2021$ competition was built on the success and insights already achieved so far (\textit{cf.} Section~\ref{sec:competition_history}). In a nutshell,
\begin{enumerate}
    \item Our chosen representation was highly effective and -- as independent work has shown for precipitation prediction~\cite{agrawal2019machine} -- should be considered a promising new technique for tackling complex geo-spatial processes.
    \item Our $2020$ competition asked participants to predict traffic in the second half of $2019$ from given data in the first half of $2019$. Thus temporal transfer learning across seasons was possible, although a performance boost could be achieved when also using the validation set data provided for training \cite{qi2020traffic4cast}. 
    \item The work of \citet{martin2020graph} following their $2019$ prize winning solution indicated that some degree of transfer of learned traffic patterns to almost unseen cities was possible with their GNN approach. 
\end{enumerate}

\subsection{Competition Tasks and their Academic and Real-World Applications}\label{sec:tasks_and_applications}
Going beyond the challenges at NeurIPS $2019$ and $2020$, the $2021$ edition explored models that adapt to domain shift both in space and time, focussing on the question of model robustness and generalizability across time and space, even if this involves a clear domain shift -- as in moving from one city to an entirely different city, or in moving from pre-COVID times to times after COVID hit the world. The change in mobility even in partial lock-downs is well-documented \cite{google_covid-19_mobility_report,axhausen_covid_2021,strava_sportjahr_2020}, and we know from colleagues in city governments that classical traffic models struggled with the drastic changes, a general well observed and explored phenomena in machine learning~\cite{bendavid2010, kouw2018, kouw2019,webb2018,gama2014,widmer1996}. The question arises how much better data driven models can perform.
We compiled an order of magnitude more data that covers 10 cities across 2019 and 2020, giving us the flexibility to explore this question on a unique, diverse real-world data set. 
We did complement these dynamic data by static information on road geometry.

With that we provided two challenges to participants:
\begin{itemize}
    \item In the {\bf core challenge}, participants were tasked to handle temporal domain shift (an active field of machine learning research \cite{bendavid2010, kouw2018, kouw2019,webb2018,gama2014,widmer1996}) in traffic due to COVID-19. In addition to the full data for four cities described, participants received pre-COVID data for four further cities, \textit{plus} one hundred 1\,h slots from $2020$ after COVID struck. The challenge then was to predict the dynamic traffic states $5,10,15,30,45$ and $60$ minutes into the future after each of the one hundred time slots for each of the additional $4$ cities.

    \item In an {\bf extended challenge}, participants were asked to again predict dynamic traffic states for two further cities, for which provided only static road geometry. Like for the first challenge, traffic needed to be predicted $5,10,15,30,45$ and $60$ minutes into the future following 100 given 1\,h time slots. Yet there was no further traffic data provided for these cities. Moreover, for each city, $50$ of these $100$ 1\,h time slots were from the period before COVID, and 50 from the period after COVID, without revealing which!
\end{itemize}

Our \textbf{core challenge} dealt with the real world problem of developing a new model or adapting an existing model to be robust to temporal domain shifts. Our competition allowed us to study this directly by providing data from pre-COVID times \textit{vs} in-COVID times for the same locations for training. The few-shot transfer learning task then was to successfully apply these models to a new city, with full data for the pre-COVID period but just a few input samples from the in-COVID period. Our \textbf{extended challenge} increased the complexity of the goal by providing no large-scale training data for the new test cities at all. Models thus needed to handle both spatial and temporal changes in the data guided by only a few example measurements from the test set. Solutions to both problems are of direct importance to real world traffic prediction and for effective city planning. 

We note that a solution to the `extended challenge' could also be applied to the `core challenge', thus truly making it more universal. Both challenges are few-shot learning challenges, an actively developing area of AI~\cite{bendavid2010, fei2006one,lu20,guo19}, and could be tackled in a multitude of ways given the data provided.
The common underlying scientific question is how to build robust models that can predict how a complex spatial process evolves over time, so that the models can swiftly adjust to domain shifts~\cite{bendavid2010, kouw2018, kouw2019,webb2018,gama2014,widmer1996} both in space and time after seeing only a few examples from the new regime.

Moreover, the ability to quickly adapt existing models to temporal domain shifts is a key concern in real world applications more generally, many of which are critical to the functioning of our society. Solutions of our \t4c competition so far seemed to generalise well to other geo-spatial processes, when aggregated similarly in space and time (such as seen for weather \cite{agrawal2019machine, herruzo_high_resolution_2021, Gruca_cdceo_21}, \textit{cf.} \url{https://weather4cast.ai/} at CIKM and IEEE Big Data 2021). Approaches developed in our competition can thus have a similarly profound impact there. Moreover, the emergence of novel approaches to key questions on model stability, adaptability, and transferability will in itself constitute a critical advance towards tackling some of humanity's largest problems.
The clear connection of the \t4c challenges to real world scenarios, the large-scale nature of the underlying data used, and their direct appeal addressing classic AI research questions were aiming at highly topical fields of few-shot and transfer learning, meta-learning more generally, graph based modelling, deep learning, and video frame prediction.

In the traffic domain, this has direct substantial implications for city planning, as first order effects of road closure or building new roads could be tackled from a data-driven perspective.
Being able to infer the traffic states of hitherto unseen cities from few observations would open the door to offering accurate traffic predictions at much lower computational cost, in more places, ultimately being able to work with more diverse, non-recurrent, data sources. This is of particular importance to cities in developing nations.
Thus, even only considering the impact on traffic, our competition will have a large impact on how we design our future way of living (societal), on our ability of how we can plan our mobility needs (economical), and our ability to provide these key services to everyone at minimal cost (humanitarian).

\subsection{Additional Competition History and Insights}\label{sec:competition_history}
Video prediction is a highly active field with promising distinct approaches~\cite{han2019video,kwon2019predicting,lee2018stochastic,oprea2020review,srivastava2015unsupervised,walker2016uncertain,xue2016visual} which we hoped could be harvested by the traffic prediction community as well.

Our 2019 competition yielded the following insights~\cite{Kreil_Traffic4cast_2019}:
\begin{itemize}
    \item  Re-phrasing traffic forecasting as a video prediction problem turned out to be of merit and -- as independent work has shown on precipitation prediction~\cite{agrawal2019machine} -- should be considered a promising new technique in tackling complex geo-spatial processes. Capturing the complex spatio-temporal dependencies of such processes is known to be a hard problem, usually referred to as `spatial is special'~\cite{anselin1989special}, and our 2019 competition contributed evidence that neural network techniques in this simple video frame prediction setting yield promising results.
    \item  Anecdotal evidence stemming from the attempt by some contestants indicated that trying to add additional prior static or dynamic location knowledge about the spatial bins did not seem to improve results significantly~\cite{martin2019traffic4cast,HerruzoPMLR2020}, suggesting that such information was already encoded in the complex traffic data itself. Moreover, no successful strategies in our competition used recurring traffic patterns based on time of day or day of week, which would have been possible in the competition design, providing further anecdotal evidence that traffic is `quasi-markovian', \ie{} mostly dependent on the immediate past only.
    \item Most successful solutions indicated~\cite{choi2019traffic,martin2019traffic4cast,yu2019crevnet} that our discretized `majority heading' channel was the least informative and was also hardest to predict.
\end{itemize}

Given the unexpected success of phrasing our traffic forecasting problem as a movie prediction task, we decided to maintain our simple spatial and temporal aggregation approach for the \t4c competition at NeurIPS 2020 and examine further some of the findings while challenging our assumptions in the design of the first competition:
\begin{itemize}
    \item We decided to double down on the implicit question thrown up by \citet{martin2019traffic4cast} and \citet{bucher2019location} of whether additional static and dynamic data potentially relevant to the geo-spatial process of traffic could improve predictions. This question was heavily explored by our contestants. There is anecdotal evidence \cite{Kopp_Traffic4cast_2020} that especially the provided incidents information was of little help. 
    \item The literature presents empirical evidence that the performance of numerous traffic prediction models would typically decrease significantly past a 15min horizon, and a majority of studies of `short-term prediction' thus focus on such time horizons \cite{Ermagun2018,dunne2011regime,ermagun2018spatiotemporal,Lana2018}. In line with common practice, we therefore restricted our 2019 competition to that horizon. After these initial experiences, in our 2020 competition we  extended the horizon  to 60min into the future and the results point to a strong affirmative answer as similar and improved architectures from the 2019 competition (\eg{} \cite{choi2020utilizing,wu2020tlab}) showed little deterioration over the longer time horizon.
\end{itemize}

Moreover, \cite{qi2020traffic4cast} reports that merging the results of a U-net architecture \cite{ronneberger2015u} with that of a Graph Ensemble Network (GEN) (extending the GNN in \cite{martin2020graph}) leads to a performance improvement, indicating that their GEN can capture different dynamic aspects to their U-net approach. They argue that in contrast to pure image-based approaches, formulating the prediction problem on a graph allows the neural network to learn properties given by the underlying street network, facilitating the transfer of a learned network to predict the traffic status at almost unknown locations. This last insight, first formulated in \cite{martin2020graph}, lies at the heart of our proposed `extended challenge' for this year's competition, essentially asking participants to use transfer learning or few-shot learning techniques to use knowledge of existing traffic dynamics in order to predict traffic states at unknown locations.


\section{Competition Data}\label{sec:data}

\subsection{Data Provisioning}
We provide a unique data set derived from industrial scale trajectories of raw GPS position fixes (consisting of a latitude, a longitude, a time stamp, as well as the vehicle speed and driving direction recorded at that time). The data are aggregated and made available by HERE Technologies and originate from a large fleet of probe vehicles which recorded their movements in multiple culturally and socially diverse metropolitan areas around the world. The time span of the data covers both the years 2019 and 2020 providing an opportunity to observe the effects of the COVID pandemic in different cities. We provide data for 10 entire cities in different configurations to empower both the temporal and spatial transfer learning tasks. Over $10^{12}$ GPS points are used to generate the data sets for the 10 cities covering a time span of 2 years.

\begin{figure}[htbp]
  \centering
  \includegraphics[width=0.85\linewidth]{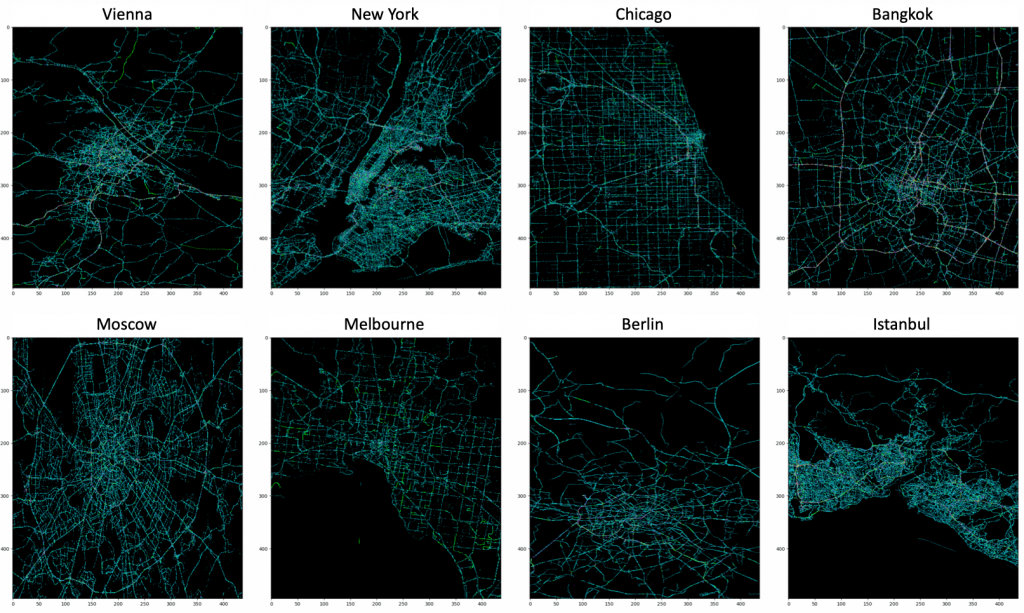}
  \vspace*{-4mm}
  \caption{The figure below shows Traffic Movie renderings from 8 different cities highlighting the diversity of locations covered with differences in road network structure, fleet layout and traffic behavior and other cultural effects. Each snapshot covers $\sim50\,\textrm{km}\times50\,\textrm{km}$ of urban area, with our data thus providing full comprehensive coverage of complex cities. Pixel brightness indicates traffic volume for an area ($\sim100\,\textrm{m}\times100\,\textrm{m}$) at the time of the snapshot and a 5 minute time bin. We show the sum over multiple channels and multiple time slots. Note, like in previous years, we rotated the data for Moscow by 90 degrees in order to fit the entire city with the enclosing ring road into the traffic movie shape. This has no influence on the prediction
  } 
  \label{fig:data_the_cities_8}
\end{figure}

The encoding and aggregation scheme of the shared data is an evolution of the data used in the two previous competitions \cite{traffic4cast2019-github,traffic4cast2020-github}. In terms of coverage and volume, the amount of data is growing and provides a unique dataset to allow conclusive results. Similarly to the previous years, the data was made freely available for download from HERE Technologies. As also mentioned in the task description Section~\ref{sec:tasks_and_applications}, we are holding back 
some time intervals for some cities (core challenge, temporal transfer), as well as 
most of the data of two further cities (extended challenge, spatio-temporal transfer).

\begin{figure}[htbp]
  \centering
  \includegraphics[angle=90,height=0.95\textheight]{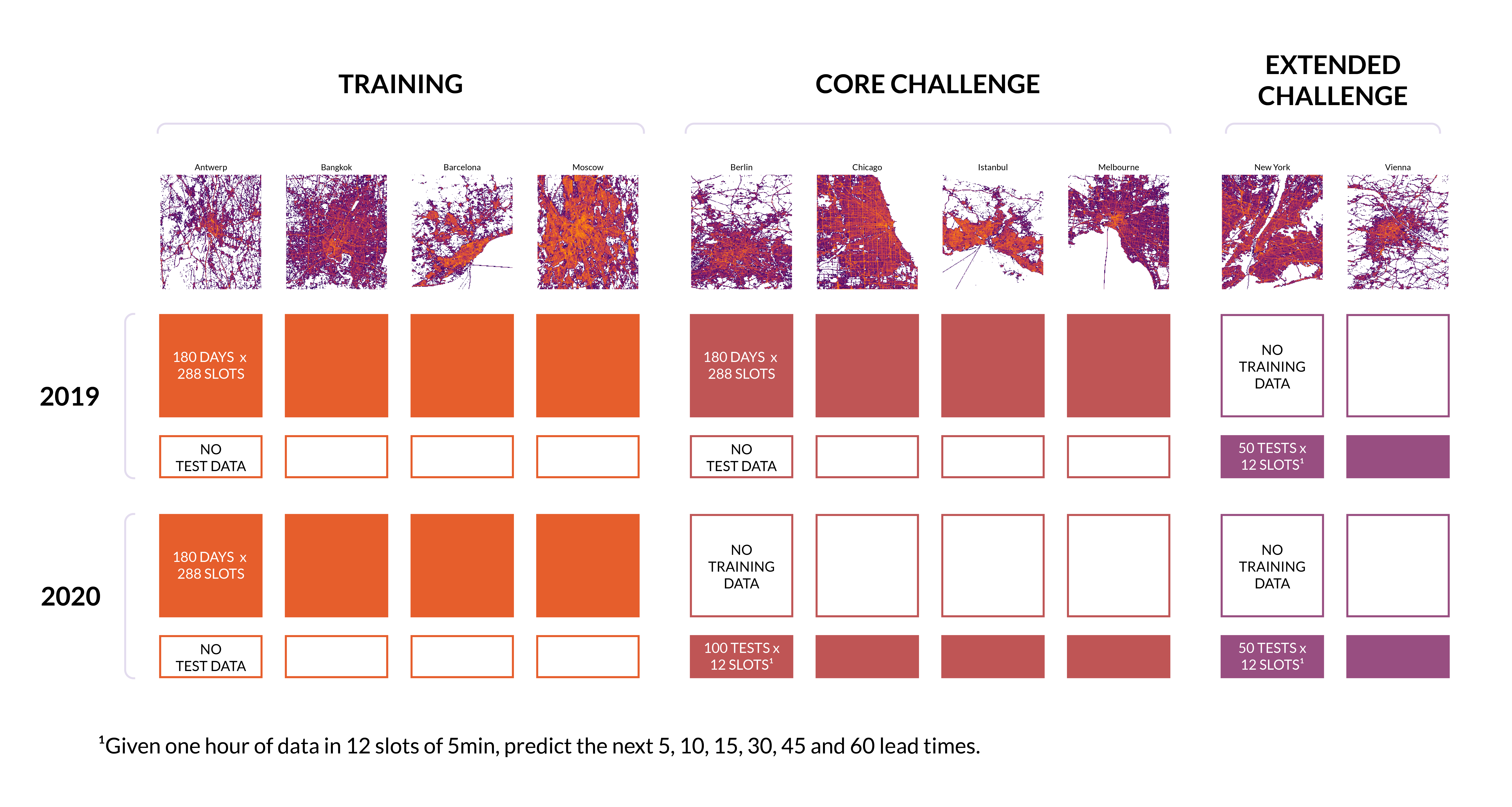}
  \caption{Data Provisioning: Training and test data provisioning for 10 cities for \t4c 2021. 
  For all cities, a static road graph was provided as well.}
  \label{fig:data_provisioning}
\end{figure}

Table~\ref{tab:data_overview} gives an overview about how data is used for training and testing of the core and extended challenge:
\begin{table}[]
    \centering
    \small
\begin{tabular}{|l|c|c|c|c|c|}
\hline
cities $\downarrow$/data$\rightarrow$ & static  & \multicolumn{4}{c|}{dynamic} \\
& & \multicolumn{2}{c|}{training}  & \multicolumn{2}{c|}{few-shots/test}\\
 &  & pre-covid  & in-covid & pre-covid & in-covid\\
 &  & (2019)  & (2020) & (2019) & (2020)\\
\hline\hline
\begin{minipage}{5cm}$4$ (known cities: Moscow, Barcelona, Antwerp, Bangkok)\end{minipage}& \checkmark & 180 days (24h)& 180 days (24h)& -- & --  \\
\hline
\begin{minipage}{5cm}$4$ (temporal: Berlin, Istanbul, Melbourne, Chicago)\end{minipage} & \checkmark & 180 days (24h) & -- & -- & 100 (1h) \\
\hline
\begin{minipage}{5cm}$2$ (spatio-temporal: Vienna and New York) \end{minipage}& \checkmark & -- & -- & 50  (1h)  & 50 (1h) \\
\hline
\end{tabular}
    \caption{Data Provisioning: Training and test data provisioning for 10 cities for \t4c 2021. For all cities, a static road graph was provided as well.}
    \label{tab:data_overview}
\end{table}
A visual representation of the provisioned data can be found in Figure~\ref{fig:data_provisioning}.

\subsubsection{Dynamic channels}

\begin{figure}[ht]
  \centering
  \includegraphics[width=0.6\linewidth]{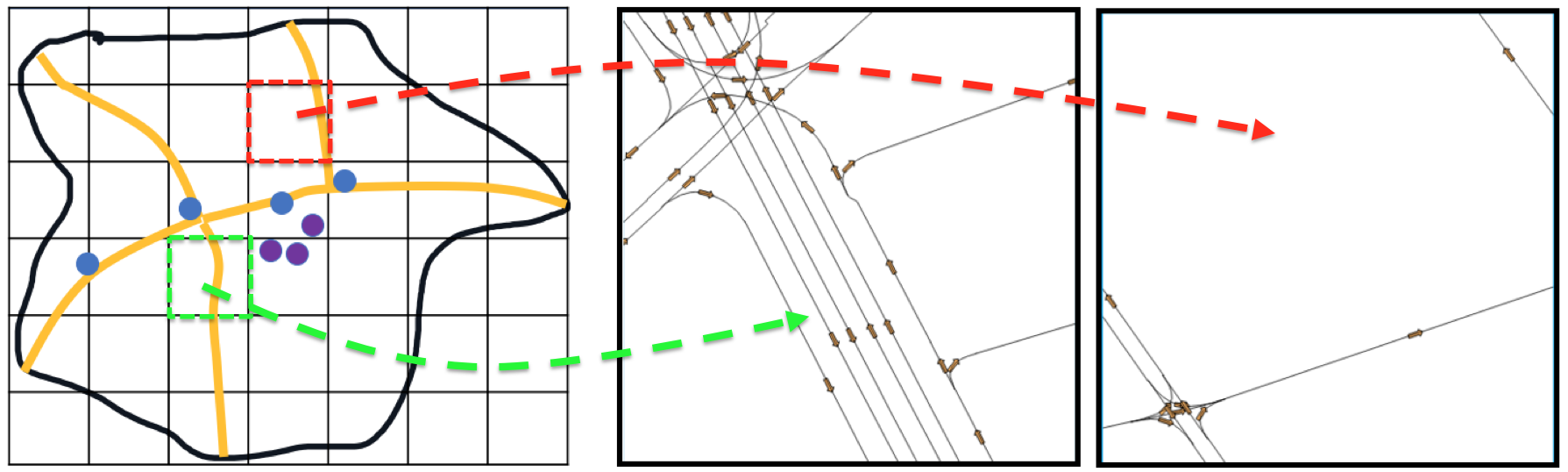}
  \vspace*{-4mm}
  \caption{Spatial tessellation of the road network. The left box shows a gridded topology (the yellow lines represent roads, dots symbolize example GPS recording locations, the black contour line outlines the city border). The two boxes on the right show a magnification of two exemplary grid cells to illustrate the contained road network complexity. The two cells are surrounded by green and red dashed square boxes and are related to the two big boxes on the right as indicated by the green and red dashed arrows.}
  \label{fig:data_probe_movies_100m}
\end{figure}

The GPS measurements are aggregated across spatial cells of $\sim100\,\textrm{m}\times100\,\textrm{m}$ and over a time-window of $5$ minutes. Figure~\ref{fig:data_probe_movies_100m} shows a simplified spatial tessellation (left) as well as the actual resolution of a road network illustrating the contents of an example $100 \times 100$ cell in a dense city area. The output of this aggregation can be encoded (described below) as an `$8$ channel movie' with each 5 minute bin representing a time frame and with the densities and other aggregate features of the cells being mapped to generalized pixel values in different channels. Hence, the output is represented as a generalized movie with 288 frames for a single day of one city.

\begin{figure}[ht]
  \centering
  \includegraphics[width=1.0\linewidth]{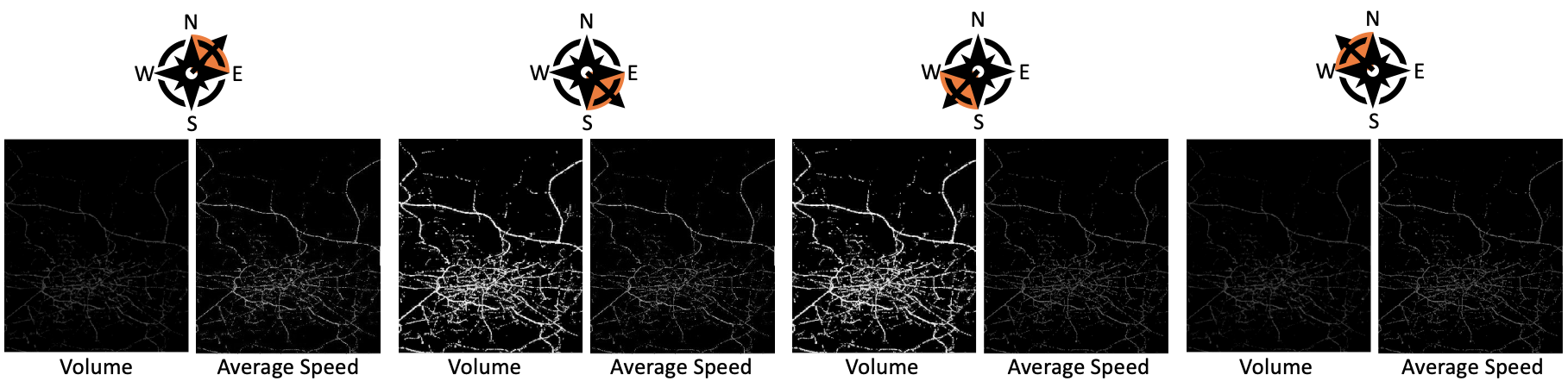}
  \vspace*{-4mm}
  \caption{The 8 probe channels, 2 for each heading quadrant.}
  \label{fig:data_eight_probe_channels}
\end{figure}

In the \textit{Traffic4cast} at NeurIPS 2020 competition, we extended the GPS feature aggregation from the initial $3$-channel (volume, mean speed and main heading) to an $8$-channel encoding (see Figure~\ref{fig:data_eight_probe_channels}), where two features are calculated for each heading direction quadrant of North--East (heading 0\textdegree--90\textdegree), South--East (heading 90\textdegree--180\textdegree), South--West (heading 180\textdegree--270\textdegree) and North--West (heading 270\textdegree--0\textdegree):
\begin{itemize}
  \item Volume: The number of probe points recorded from the collection of HERE sources capped both above and below and normalized and discretized to an integer number between $0$ and $255$. 
  \item Mean speed: The average speed from the collected probe points. The values are capped at a maximum level and then discretized to $\{1,2,\ldots, 255\}$, by linearly scaling the capping speed to $255$ and rounding the resulting values to the nearest integer. If no probes were collected, the value is $0$.
\end{itemize}
This $8$-channel representation of dynamic data is backwards compatible and allows for an easy adoption and potential re-use of solutions of our previous competitions.

Unlike in our \t4c 2020, neither static POI (point-of-interest) density maps nor the dynamic incident channel (which proved unhelpful to participants, see~\cite{Kopp_Traffic4cast_2020}) were provided.
The summary of the channel changes can be seen in Table~\ref{tab:channels_history}.
\begin{table}[htb]
\begin{center}
\small
\begin{tabular}{|l|c|c|c|c|}
\hline
description & \underline{d}ynamic / \underline{s}tatic & nb. of channels & 2020 & 2021 \\
\hline
\hline
volume per heading & d & 4 & \checkmark & \checkmark\\
 mean speed per heading & d & 4 & \checkmark & \checkmark\\
incident level & d & 1  & \checkmark &$\times$\\ \hline
POIs & s & 5 & \checkmark & $\times$\\
junction cardinality and junction count & s & 2  & \checkmark & $\times$ \\
road map & s & 9 & $\times$ & \checkmark \\
\hline
\end{tabular}
\end{center}
\caption{Channels used in the 2020 and 2021 competitions.}
\label{tab:channels_history}
\end{table}

\subsubsection{Static channels}
We provided additional static channels in order to encode contextual information about a cell.
As in most cities road geometry features at this aggregation level are reasonably static over time, for these channels only a spatial resolution is required. 

Instead of the junction cardinality and junction count channel of the previous competition \cite{Kopp_Traffic4cast_2020}, we provided a road map channel which encodes the road geometry  connections to the neighbouring cells and their general relative throughput in the present competition. In particular, with regards to the spatial transfer learning component of the 2021 competition, it appeared vital to provide such non-local connectivity information to support learning of general, transferable rules, that only rely on standardized, city-independent road features.

\begin{figure}[ht]
  \centering
  \includegraphics[width=0.9\linewidth]{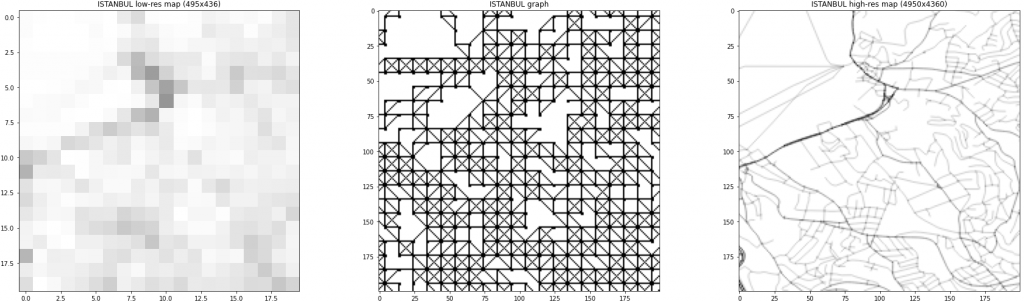}
  \vspace*{-4mm}
  \caption{Low-res grayscale (left), connectivity (middle) and high-res gray-scale image(right) for a part of Istanbul.}
  \label{fig:Istanbul_static}
\end{figure}

The data provided is illustrated in Figure~\ref{fig:Istanbul_static}:
\begin{itemize}
    \item a gray-scale representation of the city map in the same resolution as the dynamic data (Figure~\ref{fig:Istanbul_static} left).
    \item a binary encoding of whether the cell is connected to its neighbor cell/pixel to the N, NE, …, NW. This connectivity matrix (see a visualization (Figure~\ref{fig:Istanbul_static} middle)) has the same resolution as the dynamic data (495,436) but has been derived using a 
    \item finer grained representation as input (Figure~\ref{fig:Istanbul_static} right).
\end{itemize}
The city maps are meant to be compatible to the way how the probe movies are generated. The general aggregation pattern of the data is as simple as using the WGS84 degree coordinates and binning them to 0.001 intervals. Hence, every pixel represents an area of 0.001 by 0.001 degrees. In the text we often refer to this as an $\sim100\,\textrm{m}\times100\,\textrm{m}$
 area as it is more illustrative, but of course this depends on the latitude (\eg{} in Istanbul
 $\sim111\,\textrm{m}\times85\,\textrm{m}$, in Berlin $\sim111\,\textrm{m}\times70\,\textrm{m}$ or in Bangkok $\sim111\,\textrm{m}\times108\,\textrm{m}$). This can lead to the impression of seeing the data stretched/compressed compared to a typical map representation.
The variances of real cell sizes between cities are acceptable as the general aggregation scheme is coarse and also does not directly mitigate \eg{} the error introduced by different GPS populations in different cities having different general accuracies. Within a city, the cells have comparable relative areas and stable neighbourhood relationships.

\begin{figure}[ht]
  \centering
  \includegraphics[width=0.9\linewidth]{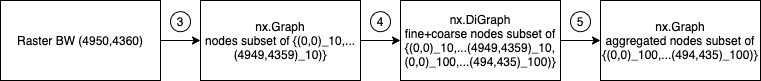}
  \vspace*{-4mm}
  \caption{Steps 3--5 of static road map generation showing input and output data structures.}
  \label{fig:static_steps2}
\end{figure}

We will now give more details about the chosen encoding and the creation of the provided connectivity layers. We refer to the 5 steps in Figure~\ref{fig:static_steps1} and \ref{fig:static_steps2} and to the code published in \cite{traffic4cast2021-github}:
\begin{figure}[ht]
  \centering
  \includegraphics[width=0.8\linewidth]{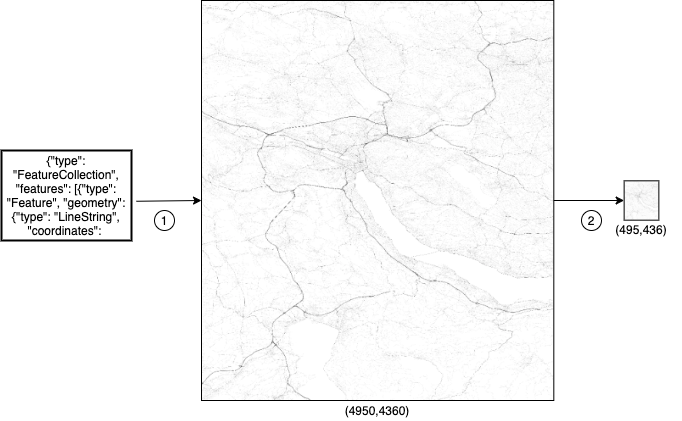}
  \vspace*{-4mm}
  \caption{First two steps of static road map generation showing input and output data structures.}
  \label{fig:static_steps1}
\end{figure}

\begin{enumerate}
    \item The source for all layers is a rasterization of the street network to a picture with 4950×4360 pixels for each city. This raster has been generated using road data from HERE Technologies (see step 1 below). The result of this rasterization is provided as the high-res image. 
    \item We generate a low-resolution raster by downsampling the image to 495$\times$436. This is the first channel in the connectivity file and just gives a very rough density value for the underlying road network.
    \item We then use the high res raster from step 1 to build a “pixel graph” at high-res, introducing  edges between non-white pixels (\ie{} not taking into account gray scales, only ``black and white'').  
    \item After building the raw pixel graph, we introduce edges between the high-res nodes (upper right of Figure~\ref{fig:connectivity_example_extract}) 
\begin{figure}[ht]
  \centering
  \includegraphics[width=0.6\linewidth]{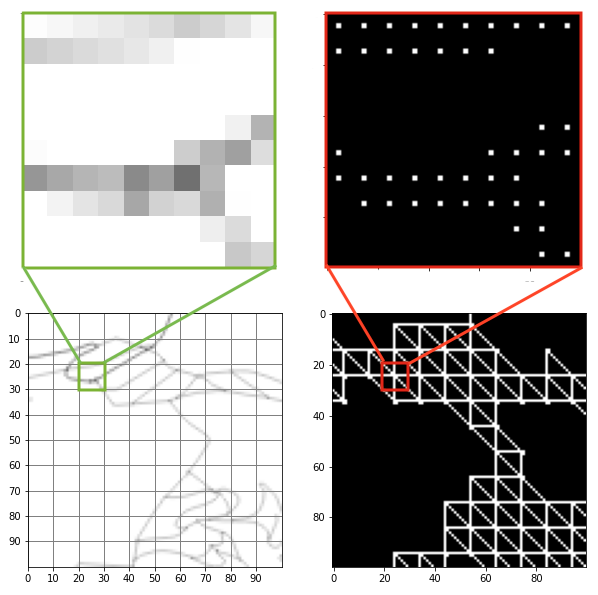}
  \vspace*{-4mm}
  \caption{Step 4 of connectivity derivation. Left: gray-scale, right: graph, top: high-res, bottom: low-res. The upper right (red border) shows an extract of the raw pixel graph for a $\sim100\,\textrm{m}\times100\,\textrm{m}$ area. Every dot is a node and every node is connected with all its direct neighbors (edges not visualized). In the upper left (green border), the magnified contents of the high res map raster image are shown. In the lower left corner, the larger context and road network in a ~$1km^2$ area with the target node resolution 495$\times$436. The green and red lines show the locations of these two extracts in the low-res representations.}
  \label{fig:connectivity_example_extract}
\end{figure}
    and their corresponding low-res nodes (lower right of Figure~\ref{fig:connectivity_example_extract}), resulting in a graph with both types of nodes.
    \item Additional detour connections at the corners of a pixel are introduced if there is a path of length $\le 7$. The detours are needed for properly representing the connectivity across the corners in the Moore neighborhood. This is illustrated in Figure~\ref{fig:connectivity_example_detours}.
    \begin{figure}[ht]
     \centering
     \includegraphics[width=0.6\linewidth]{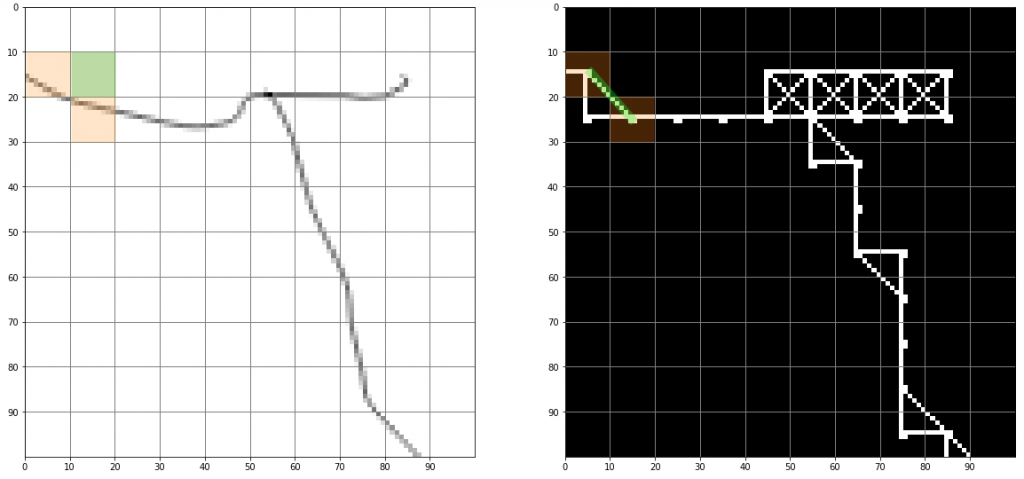}
     \vspace*{-4mm}
     \caption{Adding detours in step 5 of the connectivity derivation. Orange cells are not directly connected in the road graph but depending on the frequency a GPS point very well could move “directly” between them. Hence, there is a connection via the green cell added.}
      \label{fig:connectivity_example_detours}
    \end{figure}
\end{enumerate}
Finally, the plots of Figure~\ref{fig:Istanbul_static} show the contents of the resulting delivered static datasets for a slightly larger area. The coarse (low-res) map (right plot) has the same resolution as the dynamic data and can be used directly as a density channel. The high-res map (middle plot) serves as reference to understand the underlying road network and can also be used freely to derive other representations for solving the competition challenges. The 8 channel connectivity graph data (visualization in the left plot) represents the direct neighborhood relationships for every cell in the same resolution as the dynamic data. This representation is richer (see \eg{} the detours above) than the plain rendering of the low-res raster map and can be used more flexibly.

\subsubsection{Training Data Format}\label{sec:training_data_provisioning}
Data was made available in HDF5 format of 2 different types:

\begin{description}
    \item[static] The static part for a city consists of a $(9, 495, 436)$ tensor and  a $(4950,4360)$ tensor of higher-res gray-scale map where each pixel corresponds to $\sim10\,\textrm{m}\times10\,\textrm{m}$, and it is easy to map the pictures by factor 10. In fact, the lower-res map (first channel) is a down-sampled version of this higher-res map, and participants could generate their own static encoding at the prediction resolution. The first channel is a gray-scale representation of the city map in the same resolution as the dynamic data. The other 8 layers are a binary encoding of whether the cell is connected to it neighbor cell/pixel to the N, NE, ..., NW.
    
    \item[dynamic] The dynamic part of the training set consisting of 180 dynamic layer files (h5) each containing a $(288, 495, 436, 8)$ tensor. 
    The first two of the 8 channels encode the aggregated volume and average speed of all underlying probes whose heading is between 0 and 90 degrees (\ie{} NE), the next two the same aggregation of volume and speed for all probes heading SE, the following two for SW and NW, respectively.
    \\
    The test set for each city in both competitions asked to make 100 predictions spread over 180 days of the dynamic probe data portion, each predicting 8-channels and 6 time slots ($5,10,15,30,45$ and $60$ minutes into the future). Each single test set thus contains a $(12, 495, 436,8)$ tensor and participants are required to return a prediction consisting of a $(6, 495, 436, 8)$ tensor for this test set. In additions, for each test, a $(2)$ tensor is provided: 
    the first channel indicates the day of week (0 = Monday, ..., 6 = Sunday) and the second channel the time of day of the test slot (0, ...240) in local time.

 \end{description}
 All tensors are \texttt{uint8}.

With the focus of our proposed prediction tasks on the temporal and spatial transfer the data provided focuses on the dynamic channels from GPS data as well as the static channels encoding the properties and descriptive features of the road graph.

\subsubsection{Test data}\label{sec:test_data_provisioning}
\begin{figure}[htbp]
\centering
    \includegraphics[width=1.0\textwidth]{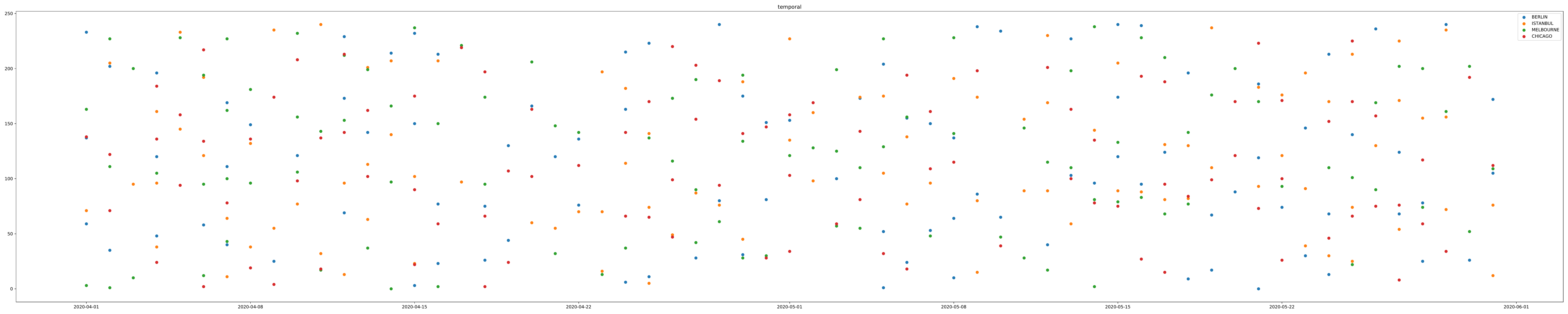}
    \includegraphics[width=1.0\textwidth]{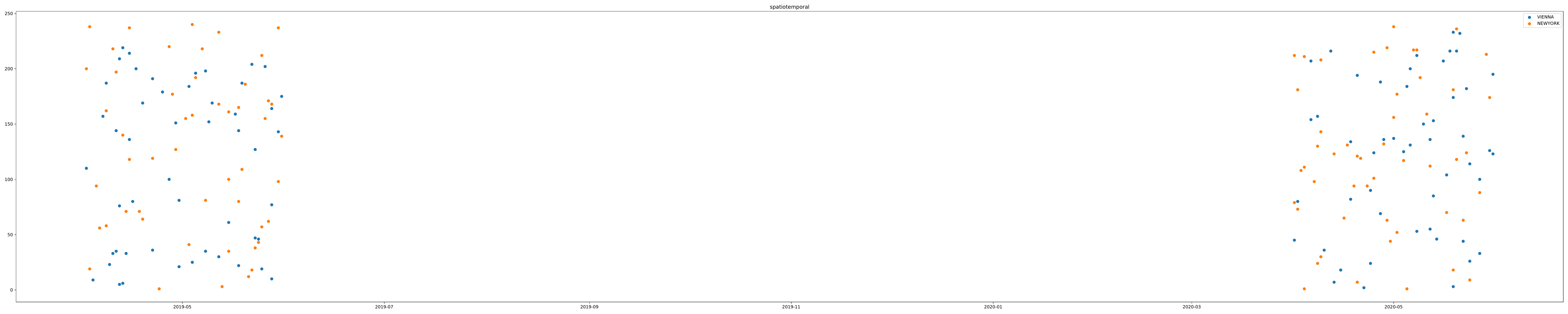}
\caption{Distribution of first slot of test input for temporal (top) and spatio-temporal challenge (bottom). The 100 slots per city for the temporal (core) challenge are sampled uniformly from April and May 2020, the 100 slots per city for the (extended) spatio-temporal challenge are sampled uniformly from April and May both 2019 and 2020. Test slots start at time 0,...,240 (8PM local time).}
\label{fig:scatter_slot_starts}
\end{figure}

For the temporal (core) challenge, tests were sampled uniformly in the target period (April/May 2020). For the spatio-temporal (extended) challenge, tests were sampled uniformly from April/May both 2019 and 2020. See the scatter plot of Figure~\ref{fig:scatter_slot_starts}.
The year of the test slot was not revealed to participants, only time of day and day of week, see Section~\ref{sec:training_data_provisioning}.

\begin{figure}[htbp]
\centering
    \includegraphics[width=0.97\textwidth]{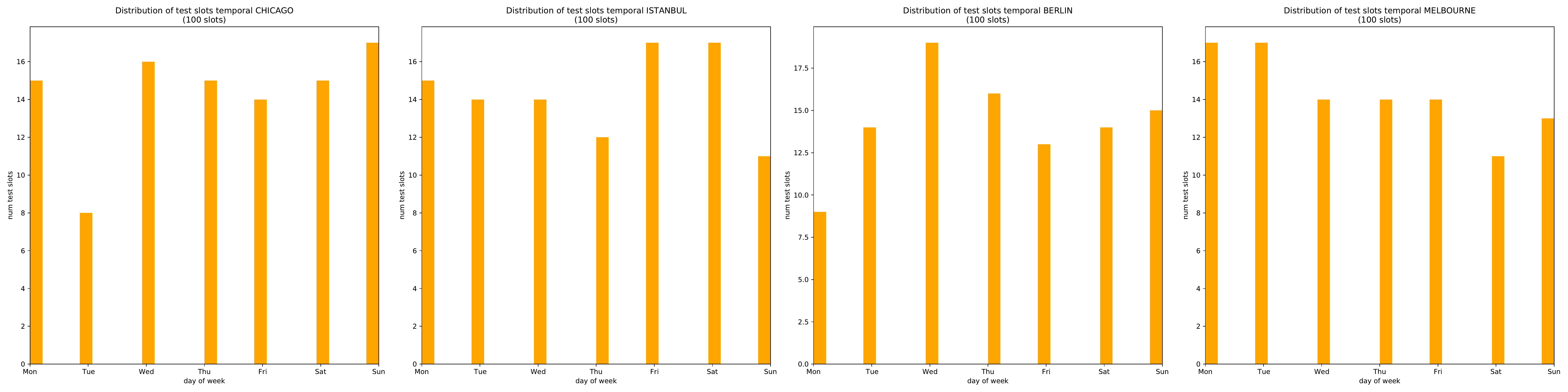}
    \includegraphics[width=0.48\textwidth]{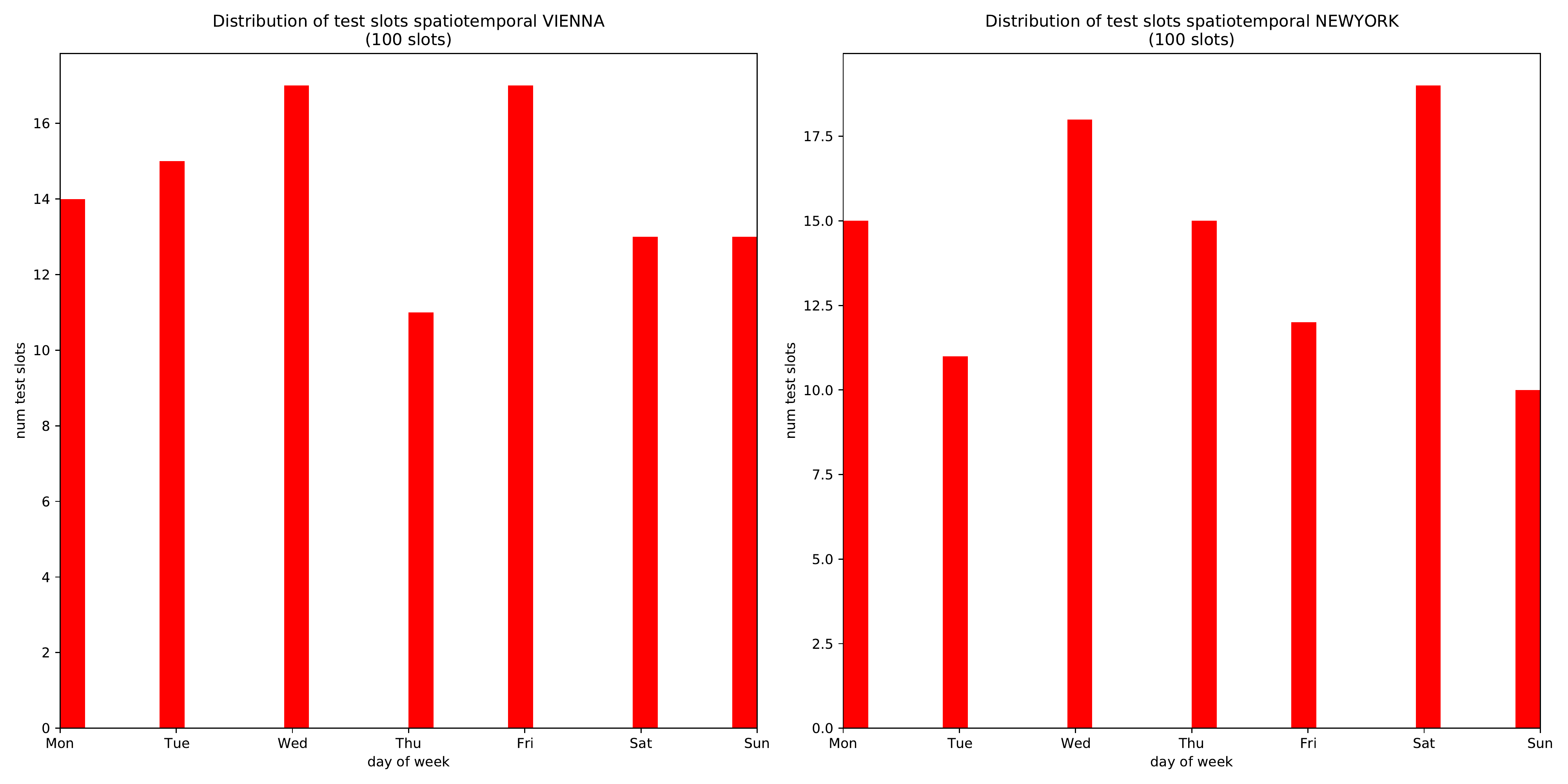}
    \vspace*{-4mm}
\caption{Per-city day of week distribution of test slots for temporal (core, yellow)
and spatio-temporal (extended, red) 
competitions.}
\label{fig:weekday_distribution_slots}
\end{figure}

\begin{figure}[htbp]
\centering
    \includegraphics[width=0.97\textwidth]{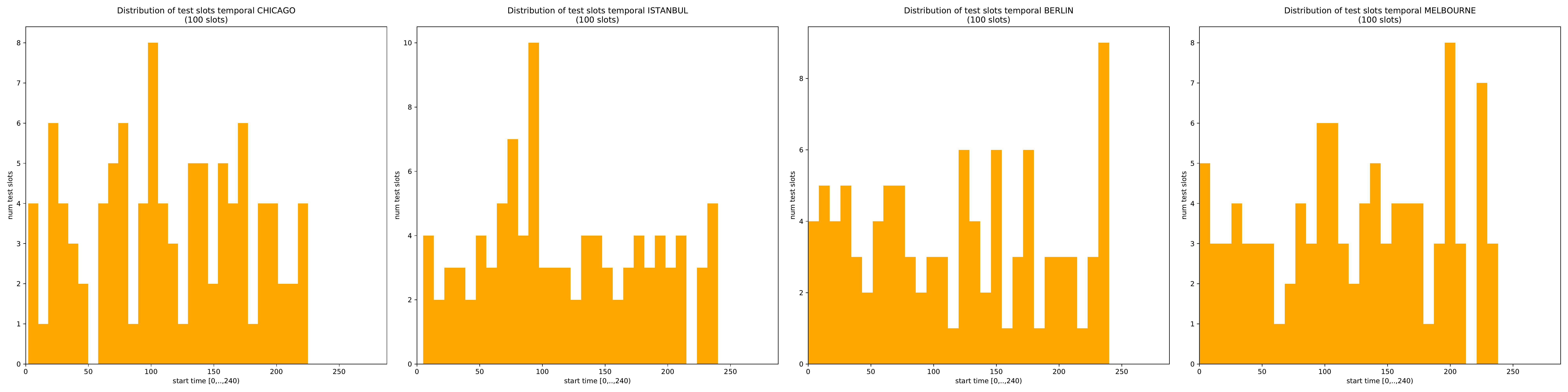}
    \includegraphics[width=0.48\textwidth]{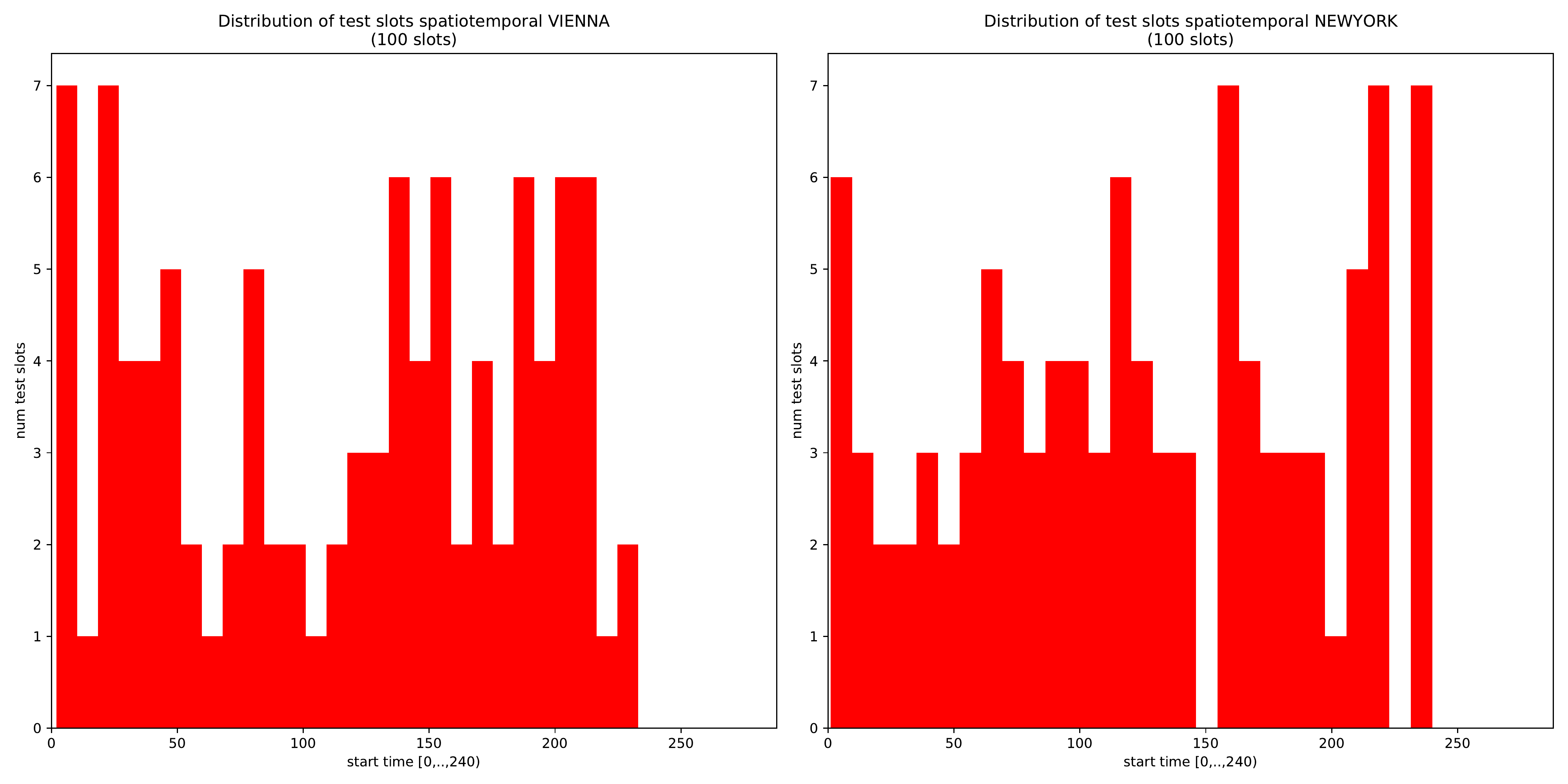}
    \vspace*{-4mm}
\caption{Per-city time of day distribution of test slots starting times for temporal (core, yellow) and spatio-temporal (extended, red) 
competitions.}
\label{fig:daytime_distribution_slots}
\end{figure}

Figure~\ref{fig:weekday_distribution_slots} shows the day of week distribution of the test slots, which are uniform for the core and extended challenges. Figure~\ref{fig:daytime_distribution_slots} again shows uniform sampling for the core and extended challenges.

\clearpage
\subsection{Temporal Shift from pre-COVID to in-COVID}\label{sec:temporal_shift}
\begin{figure}[htbp]
\centering
  \includegraphics[width=0.9\linewidth]{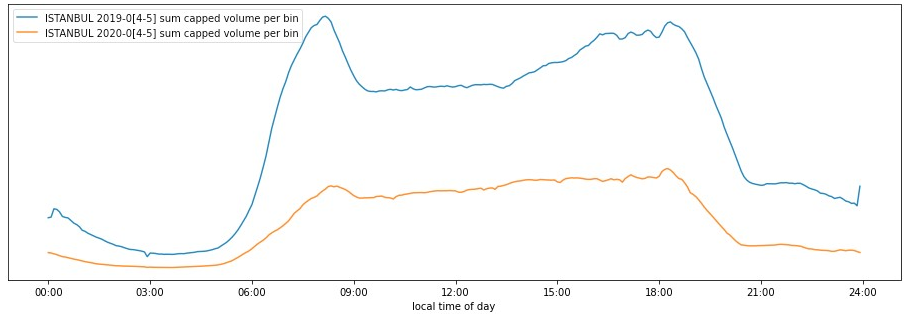}
  \caption{Istanbul traffic volumes per time of day pre-COVID vs in-COVID}
  \label{fig:covid_volume_shift_istanbul}
\end{figure}

\begin{figure}[htbp]
\centering
  \includegraphics[width=0.9\linewidth]{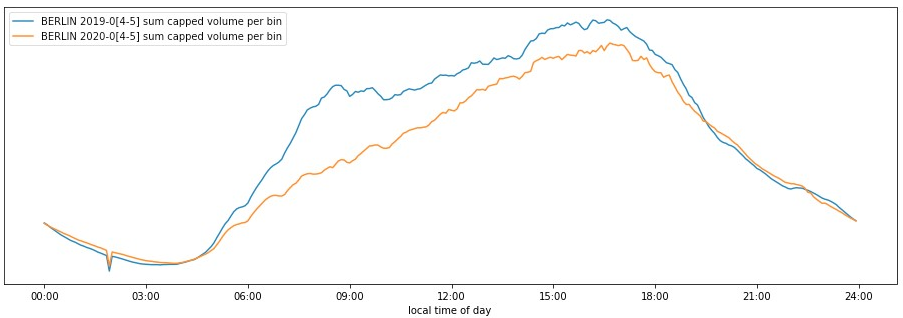}
  \caption{Berlin traffic volumes per time of day pre-COVID vs in-COVID}
  \label{fig:covid_volume_shift_berlin}
\end{figure}

\begin{figure}[htbp]
\centering
  \includegraphics[width=0.9\linewidth]{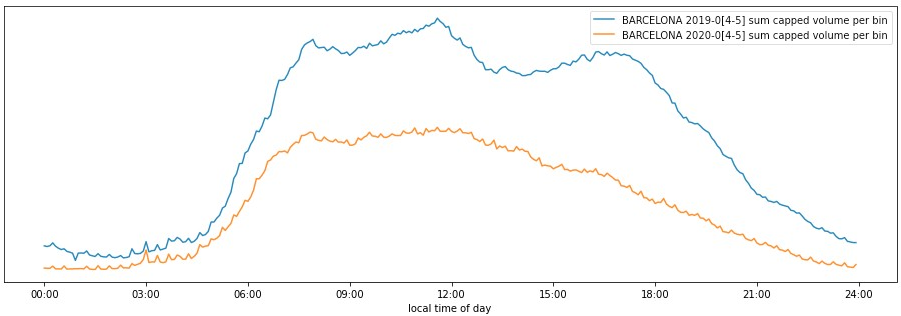}
  \caption{Barcelona traffic volumes per time of day pre-COVID vs in-COVID}
  \label{fig:covid_volume_shift_barcelona}
\end{figure}

\begin{figure}[htbp]
\centering
  \includegraphics[width=0.9\linewidth]{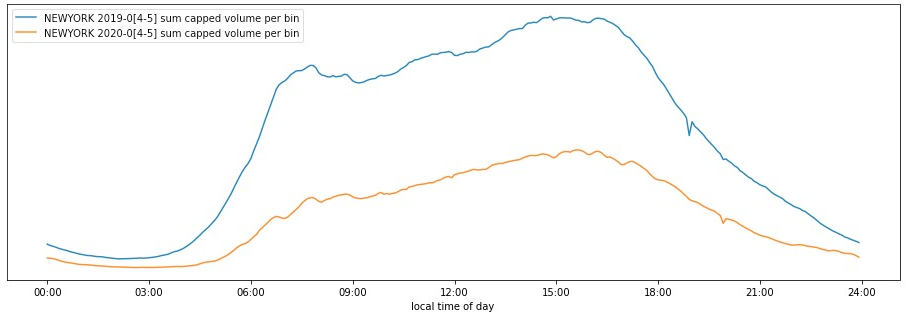}
  \caption{New York traffic volumes per time of day pre-COVID vs in-COVID}
  \label{fig:covid_volume_shift_newyork}
\end{figure}

This section gives an illustration of the temporal shift in the data which gave rise to the design of the core competition as a temporal domain-transfer task (see Section~\ref{sec:tasks_and_applications}).
In the plots below (Figures ~\ref{fig:covid_volume_shift_istanbul}--\ref{fig:covid_volume_shift_newyork}), we compare volumes for some example cities in the competition for the pre-Covid (April/May 2019) and the in-Covid (April/May 2020) regimes, respectively.

The plots show the sum of the volume for all four heading directions over all 288 5-minute bins of the day. We see a clear shift in the relative volumes for all cities. Note, volumes in each pixel are capped and normalized during the data preparation and, hence, the sums are not reflecting the absolute volume and, therefore, we do not label the y-axis in the plots below. Also, we normalized all data-sets to local time instead of UTC to have comparable curves.

Looking at Istanbul (Fig.~\ref{fig:covid_volume_shift_istanbul}), we can see a clear reduction in volume while the general pattern of traffic volume evolution during the day remains quite similar. The curfews in Turkey in 2020 were only limited to certain age groups and weekends.

The plot of Berlin (Fig.~\ref{fig:covid_volume_shift_berlin}) shows that the measures in Germany seem to have dampened the morning commute quite considerably while the afternoon peak still got to almost pre-pandemic level.

In contrast, in Barcelona (Fig.~\ref{fig:covid_volume_shift_barcelona}), it seems that the strict curfew in Spain had a clear effect on mobility during evening hours while the pattern of regular day mobility remained similar at reduced volume.

Finally, in New York (Fig.~\ref{fig:covid_volume_shift_newyork}), we see that the general pattern of traffic volumes was also preserved during April and May 2020, however at a significantly reduced volume.

\clearpage
\subsection{Exploring the Spatial Data Properties}
This section provides additional information on the inherent temporal and spatial sparsity properties of the provided in different cities and in different regions of the same city. The provided examples show how the differences in coverage as well as in update frequency of different pixels are visible in the data.

\subsubsection{City-Specific Biases}
\begin{figure}[htbp]
\centering
  \includegraphics[width=0.7\linewidth]{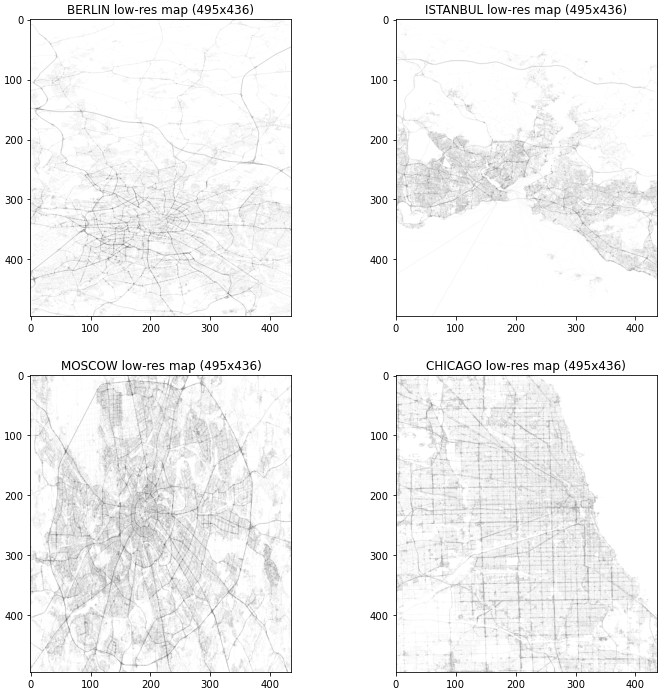}
  \caption{Simple rendering of the low-res road map in 4 competition cities}
  \label{fig:citybias_data4cities}
\end{figure}

From the simple rendering of the road network in Figure~\ref{fig:citybias_data4cities} already a quite significant difference in the density of the road network is visible. This is, of course, also related to the population density (Berlin ~3.8M, Istanbul ~15.5M, Moscow ~12.9M, Chicago ~2.9M) and other socio-economic parameters such as housing types.
In addition to the available road network, there are also other parameters and biases in the GPS data which was used to generate the competition probe data movies. The GPS recordings represent only a sub-sample of the total traffic in a city and they can have different frequencies and other temporal patterns dependent on the city and on the fleets which contributed their data. In addition to this “city bias”, there are also differences in the frequency and stability of the recordings dependent on the area of the city. This ``collection bias" is not systematic and can only be assessed empirically. Information on the fleets and recording details are not shared by the data provider. 

\subsubsection{Volume and Speed -- Coverage and Frequency}

\begin{figure}[htbp]
\centering
  \includegraphics[width=0.9\linewidth]{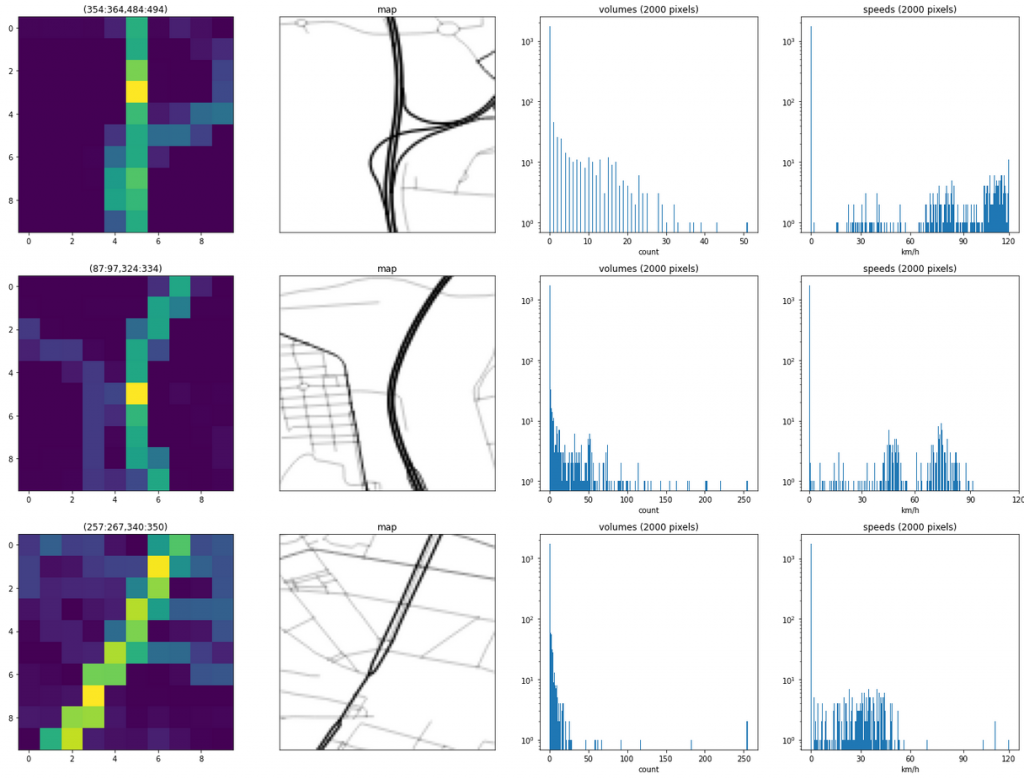}
  \caption{Three example areas in Berlin with volume heat map, road map, volume and speed histograms. The second column shows the map for illustration purposes. The third and forth columns show histograms of the recorded normalized volumes (4 channels) as well as the speeds (4 channels) during a 5 minute slot at 10am (collected for 5 consecutive days for better visibility) from a $10\times10$ area. The first column shows a log sum heatmap over all channels of the pixels in the selected area for the same data.}
  \label{fig:citybias_data3areahistograms}
\end{figure}

In order to illustrate the variance in terms of coverage and frequency in the data, we picked three example areas in Berlin (see Fig.~\ref{fig:citybias_data3areahistograms}). \#1 is a highway in the outskirts and \#2 a highway at the main ring-road. \#3 shows a boulevard in the center of Berlin with business as well as housing.

The general distributions of volumes and speeds in these areas is shown in  Figure~\ref{fig:citybias_data3areahistograms}. The volumes display the differences between a less frequently visited outskirt areas (\#1) and city center areas (\#2 and \#3). Regarding speeds, the highway in \#1 has higher speeds reported due to a higher speed limit and less traffic. The highway in \#2 has a lower speed limit and more traffic causing lower speeds during the day (see also below). The boulevard (\#3) has low speeds as expected.

\subsubsection{Volume and Speed -- Evolution Throughout a Day}

\begin{figure}[htbp]
\centering
  \includegraphics[width=0.8\linewidth]{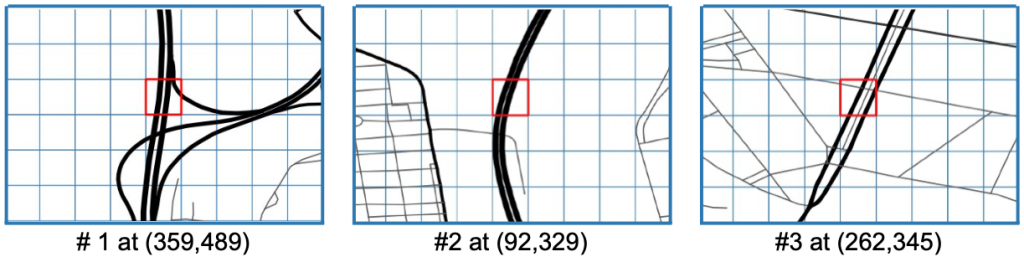}
  \caption{Three example areas in Berlin with chosen pixels (red box) for further analysis of NE heading.}
  \label{fig:citybias_data3areas}
\end{figure}

For further analysis we pick a single cell (see Fig.~\ref{fig:citybias_data3areahistograms} and \ref{fig:citybias_data3areas}) at the center of the areas from Figure~\ref{fig:citybias_data3areahistograms} for having a look at the volume and speed throughout an example day.

\begin{figure}[htbp]
\centering
  \includegraphics[width=0.9\linewidth]{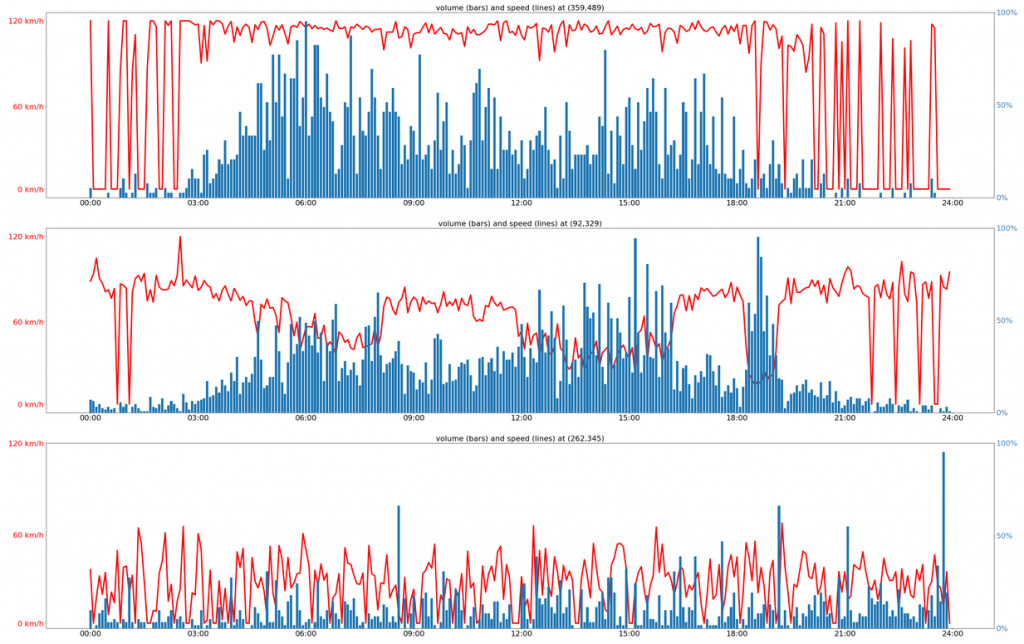}
  \caption{Volume (blue bars) and speed patterns (red line) for the selected pixels in NE heading on a Wednesday in 2019}
  \label{fig:citybias_data1pxvolspeedwed}
\end{figure}

Figure~\ref{fig:citybias_data1pxvolspeedwed} shows the speeds as red lines and the volumes as blue bars for the selected pixels on a Wednesday in 2019.
If no volume is reported (volume 0 or below privacy threshold), the speed is also 0. In the first plot for area \#1 we see the normal daily pattern with increasing volume during morning and evening rush hour but no congestion as we are in the outskirts. In the second plot for area \#2, we see the volume evolution during the day with the speeds indicating congestion and lower speeds during the day especially during morning and afternoon hours with a congestion between 6pm and 7pm. The third plot for area \#3 shows the city boulevard with very flaky data as speeds are generally low and volumes are spiky with \eg{} cars stopping.

\begin{figure}[htbp]
\centering
  \includegraphics[width=0.9\linewidth]{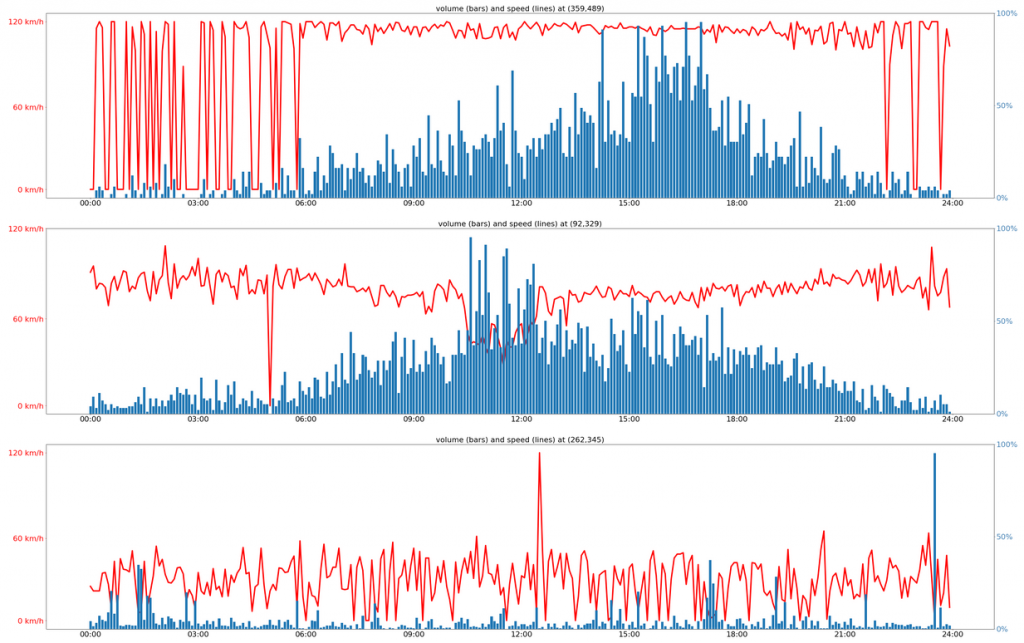}
  \caption{Volume and speed patterns for the selected pixels on a Sunday in 2019}
  \label{fig:citybias_data1pxvolspeedsun}
\end{figure}

In contrast, Figure~\ref{fig:citybias_data1pxvolspeedsun} show speeds and volumes for the same pixels, however for a Sunday in 2019. The reduced volume in early morning hours are clearly visible in all three plots, while the second plot for area \#2 also reports heavy traffic on the ring road in the late morning.

\subsubsection{Why does this matter?}

The examples shown in this section explore the sparsity of the input data. Both in terms of volume and speed the coverage as well as the update frequency do differ within a city and between cities. In order to properly predict the traffic volume, models need to be able to cope with this sparsity.

\clearpage
\section{MSE Details}\label{sec:mse}

\subsection{Spatially Masked MSE}
This section has a look at differences between a full and a masked MSE. In particular it is important to see whether the image-focused MSE also works for the traffic map prediction where \eg{} graph-based approaches would only predict in areas covered by the road graph. 

For the illustration we are using a prediction for Istanbul using the naive average baseline\footnote{\url{https://github.com/iarai/NeurIPS2021-traffic4cast/blob/master/baselines/naive_average.py}}, which takes the average over all input data for all 6 predictions. Istanbul is a “hard” example since, due to the waterways in the city, there are more non-road areas with traffic than in other cities. On the waterways vehicles are not following exactly the same path and hence a road-mask would not work, these variances are visible in the movie visualizations of the data.

\begin{figure}[htbp]
\centering
  \includegraphics[width=0.9\linewidth]{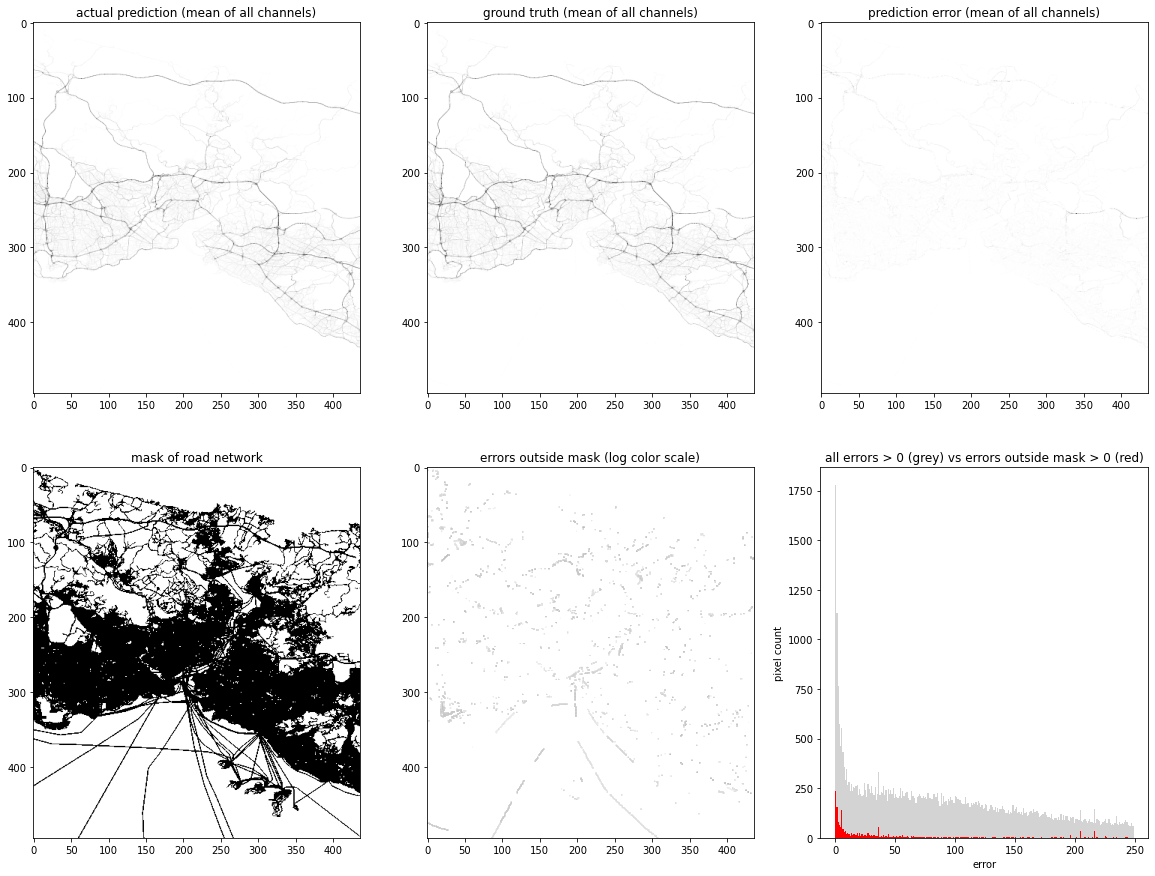}
  \caption{Illustrations of the masking in Istanbul. The first row shows the mean over all 8 channels for the actual prediction (left) and the ground truth (middle) for the full prediction horizon. The plot on the right-hand side shows the per-pixel mean squared prediction error (prediction – ground-truth) and one can already see that the visible errors are actually in areas where there are also important roads. The mask, shown in the bottom left plot, is based on the static road map which is provided as part of the competition data (1 for all pixels intersected by a road, 0 everywhere else). In the middle in the bottom row, only the prediction errors that would be masked away are shown. This output has been exaggerated logarithmically in order to visually see the small pixel errors from the ferry lines in the lower middle of the map. The bottom-right histogram shows the value distributions for the prediction error. The x-axis shows the magnitude of the error while the y-axis indicates the count of affected pixels. The red bars in the plot show the portion of the errors that the mask would take away.}
  \label{fig:blog_mse_masked_diffs_small}
\end{figure}

The value distribution in Figure~\ref{fig:blog_mse_masked_diffs_small} (bottom right) shows that the small number of affected pixels is quite homogeneously distributed. And, in the specific example of Istanbul, the difference of the masked MSE vs the normal one is approx. 0.4.  Table~\ref{tab:mse_masked_stats} shows the “graph only” difference for all cities in the core challenge and as expected, Istanbul is by far most affected and overall, the differences are even smaller.

\begin{table}[htb]
    \scriptsize\centering
    \footnotesize\centering
    \begin{tabular}{l|l|l|l|l|l|l|}
         city & full mse & both masked$^a$ & $\Delta$ both masked & graph only$^b$ & $\Delta$ graph only & test slots \\ \hline
         Istanbul & 68.041132 & 67.018380 & 1.022752 & 68.410726 & -0.369594 & 50 \\
         Berlin & 97.128263 & 96.426923 & 0.701340 & 97.270101 & -0.141838 & 50 \\
         Melbourne & 36.198404 & 36.133144 & 0.065290 & 36.195613 & 0.002791 & 50 \\
         Chicago & 45.284226 & 45.246546 & 0.037681 & 45.282553 & 0.001673 & 50 \\
         All & 61.663006 & 61.206241 & 0.456766 & 61.789748 & -0.126742 & 200 
    \end{tabular}
    \caption{MSE ``both masked'' and “graph only” difference for all cities in the core challenge. Data comes from 5 days of the training data and 10 time slots, predictions using the ``naive average baseline'' (taking the average of the input as output for all 6 time slots in the future for every pixel and channel). $^a$ mask applied to prediction and ground truth. $^b$ mask applied to prediction only.}
    \label{tab:mse_masked_stats}
\end{table}

The examples shown here can also be reproduced and explored in more detail in the notebook provided in the competition GitHub repo\footnote{\url{https://github.com/iarai/NeurIPS2021-traffic4cast/blob/master/metrics/metrics.ipynb}}.

\clearpage
\subsection{Exploring MSE}

\begin{figure}[htbp]
\centering
  \includegraphics[width=0.8\linewidth]{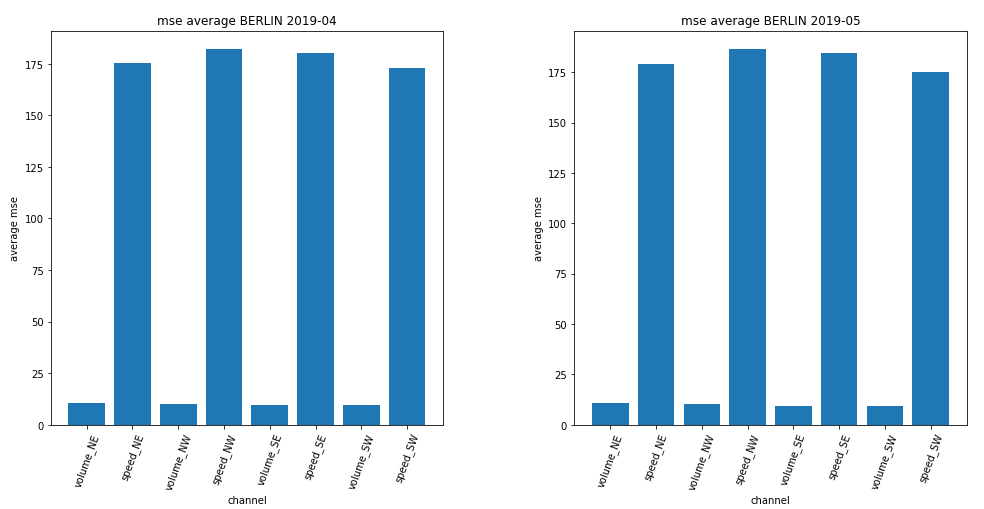}
  \vspace*{-6mm}
  \caption{Naive average baseline MSE per channel (speed and volume in 4 heading directions). Data from 3 sample days in Berlin from April 2019 (left) and May 2019 (right), 22 tests per day starting at every full day.}
  \label{fig:blog_mse_per_channel}
\end{figure}

This section uses the naive average baseline\footnote{\url{https://github.com/iarai/NeurIPS2021-traffic4cast/blob/master/baselines/naive_average.py}} to explore how the MSE behaves and where models might struggle. Figure~\ref{fig:blog_mse_per_channel} shows the general MSE levels of the different channels for the example from Berlin. It is clearly visible that the general level of the MSE for speed values is significantly higher than the volume MSE.

\begin{figure}[htbp]
\centering
  \includegraphics[width=0.8\linewidth]{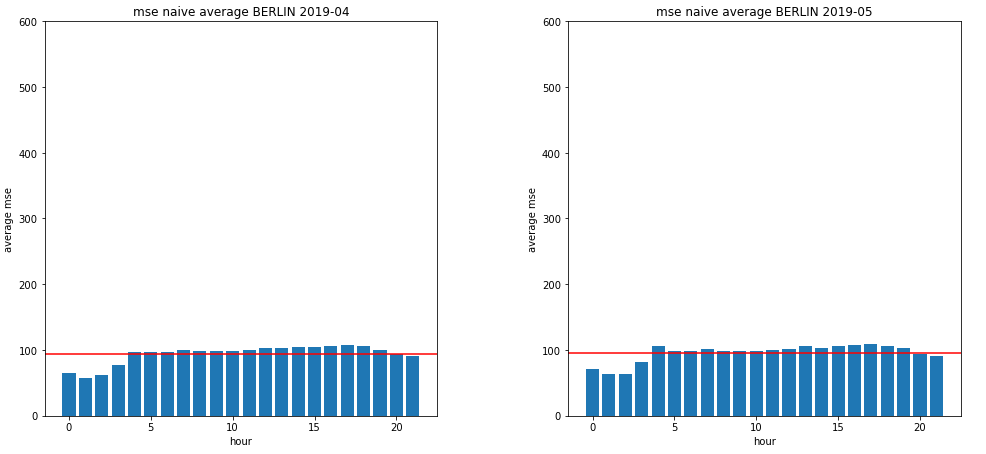}
  \vspace*{-6mm}
  \caption{Naive average baseline MSE throughout the day (hourly averages). Same data as in Fig.~\ref{fig:blog_mse_per_channel}.}
  \label{fig:blog_mse_over_day}
\end{figure}

The temporal dimension of the MSE is shown in Figure~\ref{fig:blog_mse_over_day}, we see that averaging over the input sequence follows the general peak pattern.

\begin{figure}[htbp]
\centering
  \includegraphics[width=0.8\linewidth]{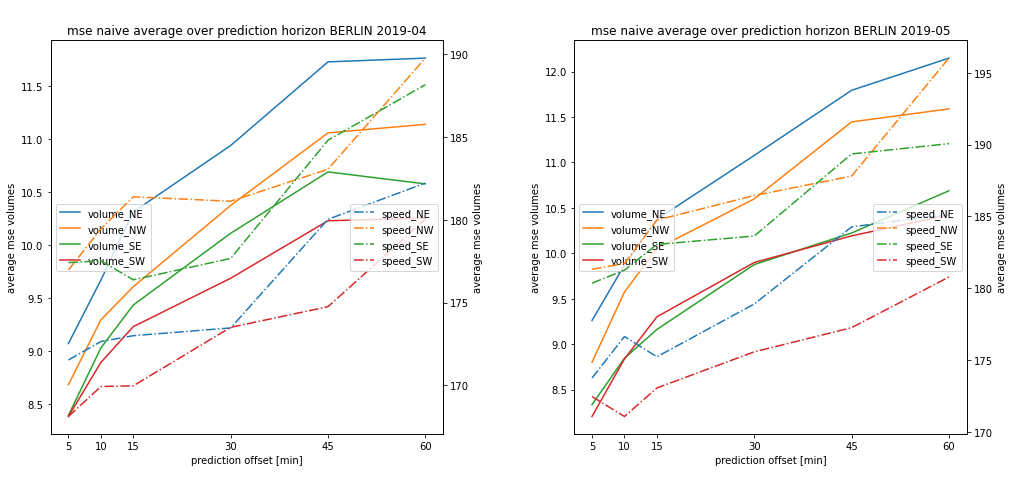}
  \vspace*{-6mm}
  \caption{Naive average baseline MSE over the prediction horizon (5, 10, 15, 30, 45 and 60 minutes) and per channel. Notice the different scales for volume (left) and speed (right). Same data as in Fig.~\ref{fig:blog_mse_per_channel}.}
  \label{fig:blog_mse_prediction_horizon}
\end{figure}

Figure~\ref{fig:blog_mse_prediction_horizon} shows the quality of the prediction over the prediction horizon. As expected, naive averaging is better in the next slots to predict, but the ascent is not as steep as one might have expected.

\begin{figure}[htbp]
\centering
  \includegraphics[width=0.9\linewidth]{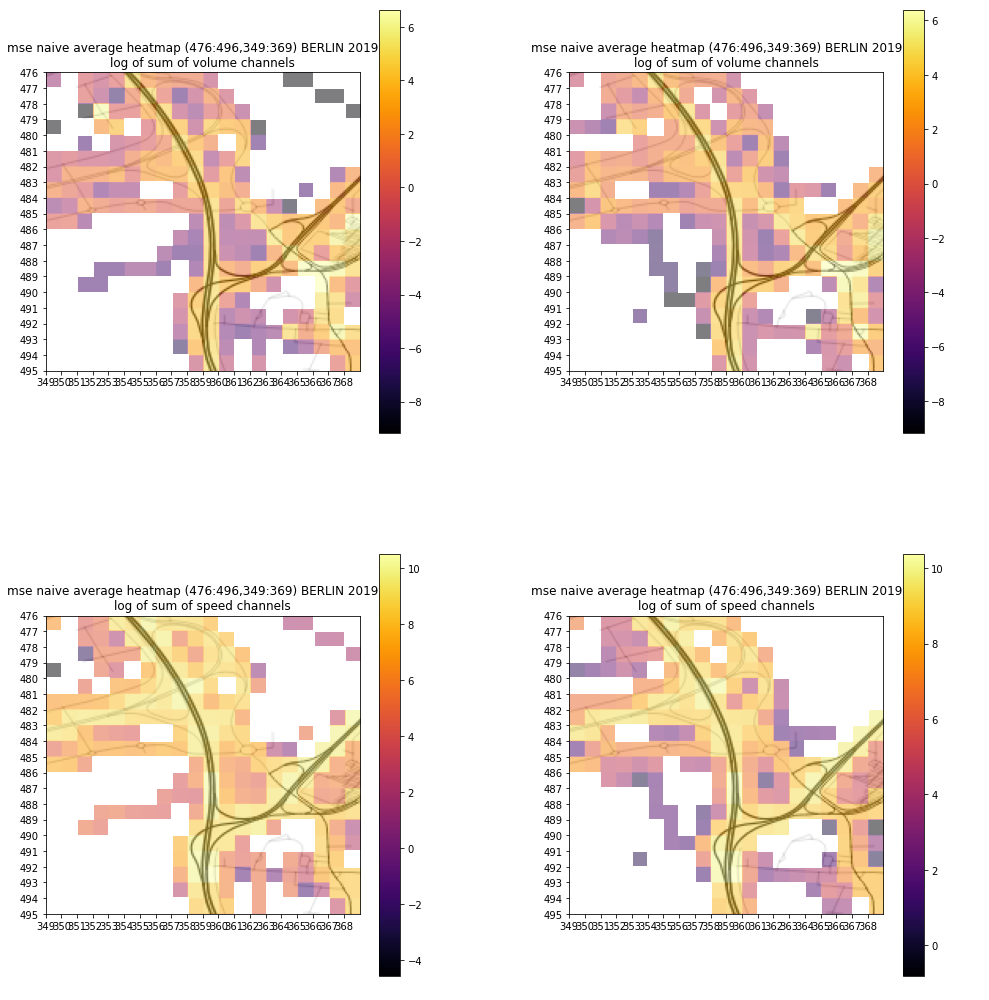}
  \vspace*{-6mm}
  \caption{Naive average baseline MSE heatmap (per pixel, volume top row and speed bottom row) for a $10\times10$ area in Berlin. Same data as in Fig.~\ref{fig:blog_mse_per_channel}.}
  \label{fig:blog_mse_mse_heatmp}
\end{figure}

Per-pixel loss is largest on most used roads, both in volume and speed in the examples in Figure~\ref{fig:blog_mse_mse_heatmp}.

\clearpage
\section{Baselines}\label{sec:baselines}
Here, we describe in more detail the two non-trivial baselines (see also Table~\ref{tab:synopsis_baselines}) provided.

For the core competition, we used a vanilla U-Net \cite{ronneberger2015u} and trained a separate model for each city for 4 epochs on the city's 2019 training data\footnote{The code for the U-Net baseline can be found at \url{https://github.com/iarai/NeurIPS2021-traffic4cast/blob/master/baselines/baselines_unet_separate.py}.}.

For the extended competition, a Graph ResNet was used following \cite{martin_graph-resnets_2020}.
In contrast to them, we do not the derive the graph from dynamic data, but derive it from the static data provided in the competition. We used a 194'000 training samples from BERLIN for one epoch. We enumerate the nodes of this graph; this enumeration corresponds to a subset of all the pixels (0,0),..,(494,435). This allows us to represent the graph's edges as pairs of integers. In addition, we pass the dynamic data indexed by the enumerated nodes. The dynamic data not covered by the graph is discarded. This data object is then used to train a graph resnet. 
In summary, the input to the graph res net is a \texttt{torch\_geometric} data object consisting of
\begin{verbatim}
    x: (num_nodes, 12 * 8) float tensor
    edge_index: (num_edges,2) long tensor
    y: (num_nodes, 6*8) float tensor
\end{verbatim}
This data object is then used to train a graph resnet. 
\begin{figure}[htbp]
  \centering
  \includegraphics[scale=0.6]{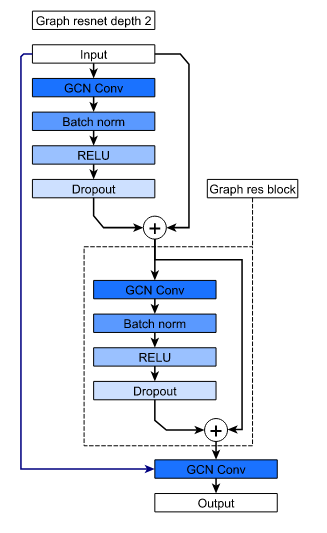}
  \caption{Graph res net architecture from \cite{martin_graph-resnets_2020}, reprinted with permission, based on \cite{kipf2017semisupervised}.}
  \label{fig:martin_gcn}
\end{figure}
Here, the input of the current layer is added to the output of the current layer before passing it on to the next graph residual block as shown in Figure~\ref{fig:martin_gcn}.
\clearpage
We deliberately did not explore fine-tuning and approaches feeding the output of the graph resnet to a U-net for instance. We did not want to bias participants too much in the hope our code can help participants get started more quickly\footnote{The code can be found under
\url{https://github.com/iarai/NeurIPS2021-traffic4cast/blob/master/baselines/graph_models.py}. The call was:
\texttt{python baselines/baselines\_cli.py --model\_str=gcn --limit=200000  --epochs=1 --batch\_size=5 --num\_workers=16 --train\_fraction=0.97   --val\_fraction=0.03 --file\_filter="**/*BERLIN*8ch.h5"}.}.

Table~\ref{tab:synopsis_baselines} gives an overview of the two baselines.
\begin{table}[htb]
    \scriptsize\centering
\begin{tabular}{|p{3.6cm}|p{1.7cm}|p{1.0cm}|p{1.2cm}|p{2.8cm}|p{1.6cm}|p{0.7cm}|}
\hline
Team, rank (core/ext), approach & road~graph, time-of-day, day-of-week$^a$ & models trained p. city$^b$ & \#{}models trained$^c$ & Training datasets$^d$ & $\sum\#{}$params core / ext $^e$ & mask$^f$\\ \hline \hline

\textbf{moritzneun (11/--)}\newline  separate U-Net & no & no& 4 / --& $\{C[1-4]\}$/-- & 31.1M/-- & --\\ \hline
\textbf{christian (--/8)}\newline Graph ResNet  & road graph & no& --/1 & --/$\{C1\}$  & --/0.2M & --\\ \hline
\end{tabular}
    \vspace*{-2mm}
    \caption{Synopsis for our baselines. $^a$ what supplemental information is used;
      $^b$ whether some of the trained models used in the inference are specifically trained on the target city;
      $^c$ 9/7 means 9 models in the core and 7 different models in the extended competition, whereas 1=1 means the same trained model was used in both competition;
      $^d$ T1=Moscow, T2=Barcelona, T3=Antwerp, T4=Bangkok, C1=Berlin, C2=Istanbul, C3=Melbourne, C4=Chicago, E1=Vienna, E2=New York, T*=all training cities, C*=all core cities, E*=all extended cities, \eg{} (9/7)$\times$\{T*,C*\} means 9 models trained on all training and core competition cities for the core competition and 7 from the same cities for the extended competition, and \{T*,C[1-4]\} expands to a model for each city trained on all training cities plus one of the core cities;
      $^e$ Sum of trainable weights of all the model checkpoints used in the inference;
      $^f$ Kind of mask used for post-processing..}
    \label{tab:synopsis_baselines}
    \vspace*{-2mm}
\end{table}

\end{document}